\documentclass[11pt,twoside]{article} 

\usepackage{amsmath,amsfonts,bm}

\def\eqref#1{equation~\ref{#1}}

\def\1{\bm{1}}

\DeclareMathAlphabet{\mathsfit}{\encodingdefault}{\sfdefault}{m}{sl}
\SetMathAlphabet{\mathsfit}{bold}{\encodingdefault}{\sfdefault}{bx}{n}

\DeclareMathOperator*{\argmin}{arg\,min}

\usepackage[utf8]{inputenc} 
\usepackage[T1]{fontenc}    
\usepackage{booktabs}       
\usepackage{amsfonts}       
\usepackage{nicefrac}       
\usepackage{microtype}      

\usepackage{subfigure}
\usepackage{epsf}
\usepackage{epsfig}
\usepackage{fancyhdr}
\usepackage{graphics}
\usepackage{graphicx}
\usepackage{psfrag}
\usepackage{fullpage}
\usepackage{pdfpages}

\usepackage{url}
\usepackage[colorlinks=True,linkcolor=magenta,citecolor=blue,urlcolor=blue,pagebackref=true,backref=true]
{hyperref}
\renewcommand*{\backref}[1]{\ifx#1\relax \else Page #1 \fi}
\renewcommand*{\backrefalt}[4]{%
    \ifcase #1 \footnotesize{(Not cited.)}%
    \or        \footnotesize{(Cited on page~#2.)}%
    \else      \footnotesize{(Cited on pages~#2.)}%
    \fi}
\usepackage{color}

\usepackage{amsthm}
\usepackage{amsmath}
\usepackage{amssymb,bbm}
\usepackage{caption}
\usepackage{algorithmic}
\usepackage{algorithm}

\newtheorem*{remark}{Remark}

\newtheorem{assumption}{Assumption}

\newtheorem{theorem}{Theorem}

\newtheorem{definition}{Definition}

\newcommand{\dislice}{\text{DSW}}
\newcommand{\digslice}{\text{DGSW}}

\newcommand{\conspace}{\mathbb{M}}

\newcommand{\sphere}{\mathbb{S}}

\def\argmin{\textnormal{arg} \min}

\newcommand{\norm}[1]{\left\lVert#1\right\rVert}

\newcommand{\widgraph}[2]{\includegraphics[keepaspectratio,width=#1]{#2}}

\begin{document}

\begin{center}

{\bf{\LARGE{Distributional Sliced-Wasserstein and Applications to Generative Modeling}}}
  
\vspace*{.2in}
{\large{
\begin{tabular}{cccc}
Khai Nguyen$^{\diamond}$ & Nhat Ho$^{\dagger}$ & Tung Pham$^{\diamond}$ & Hung Bui$^{\diamond}$
\end{tabular}
}}

\vspace*{.2in}

\begin{tabular}{c}
VinAI Research$^\diamond$, University of Texas, Austin$^{\dagger}$\\
\end{tabular}

\today

\vspace*{.2in}

\begin{abstract}
Sliced-Wasserstein distance (SW) and its variant, Max Sliced-Wasserstein distance (Max-SW), have been used widely in the recent years due to their fast computation and scalability even when the probability measures lie in a very high dimensional space. However, SW requires many unnecessary projection samples to approximate its value while Max-SW only uses the most  important projection, which ignores the information of other useful directions. In order to account for these weaknesses, we propose a novel distance, named \emph{Distributional Sliced-Wasserstein} distance (DSW), that finds an \emph{optimal} distribution over projections that can balance between exploring distinctive projecting directions and the informativeness of projections themselves. We show that the DSW is a generalization of Max-SW, and it can be computed efficiently by searching for the optimal push-forward measure over a set of probability measures over the unit sphere satisfying certain regularizing constraints that favor distinct directions. Finally, we conduct extensive experiments with large-scale datasets to demonstrate the favorable performances of the proposed distances over the previous sliced-based distances in generative modeling applications.
\end{abstract}
\end{center}
\section{Introduction}
Optimal transport (OT) is a classical problem in mathematics and operation research. Due to its appealing theoretical properties and flexibility in practical applications, it has recently become an important tool in the machine learning and statistics community; see for example,~\cite{courty2017joint, arjovsky2017wasserstein, tolstikhin2018wasserstein, gulrajani2017improved} and references therein.
The main usage of OT is to provide a distance named Wasserstein distance, to measure the discrepancy between two probability distributions. However,  that distance  suffers from expensive computational complexity, which is the main obstacle to using OT in practical applications.  

There have been two main approaches to overcome the high computational complexity problem: either approximate the value of OT or apply the OT adaptively to  specific situations. The first approach was initiated by Cuturi~\cite{cuturi2013sinkhorn} using an entropic regularizer to speed up the computation of the OT~\cite{sinkhorn1967diagonal, knight2008sinkhorn}. The entropic regularization approach has demonstrated its usefulness in several application domains~\cite{courty2014domain, genevay2018learning, bunne2019learning}. Along this direction, several works proposed efficient algorithms for solving the entropic OT~\cite{altschuler2017near, lin2019acceleration, lin2019efficient} as well as methods to stabilize these algorithms~\cite{chizat2018scaling,peyre2019computational, chizat2018scaling, schmitzer2019stabilized}. However, these algorithms have complexities of the order $\mathcal{O}(k^2)$, where $k$ is the number of atoms. It is expensive when we need to compute the OT repeatedly, especially in learning the data distribution. 

The second approach, known as "slicing", takes a rather different perspective. It leverages two key ideas: the OT closed-form expression for two distributions in one-dimensional space, and the transformation of a distribution into a set of  projected one-dimensional distributions by the Radon transform (RT)~\cite{helgason2010integral}. 
The popular proposal along this direction is Sliced-Wasserstein (SW) distance~\cite{bonneel2015sliced}, which samples the projecting directions uniformly over a unit sphere in the data ambient space and takes the expectation of the resulting one-dimensional OT distance. The SW distance hence requires a significantly lower computation cost than the original Wasserstein distance and is more scalable than the first approach. Due to its solid statistical guarantees and efficient computation, the SW distance has been successfully applied to a variety of practical tasks~\cite{deshpande2018generative, liutkus2019sliced, kolouri2018sliced, wu2019sliced, deshpande2019max} where it has been shown to have comparative performances to other distances and divergences between probability distributions. However, there is an inevitable bottleneck of computing the SW distance. Specifically, the expectation with respect to the uniform distribution of projections in SW is intractable to compute; therefore, the Monte Carlo method is employed to approximate it. Nevertheless, drawing from a uniform distribution of directions in high-dimension can result in an overwhelming number of irrelevant directions, especially when the actual data lies in a low-dimensional manifold. Hence, SW typically needs to have a large number of samples to yield an accurate estimation of the discrepancy. Alternatively, in the other extreme, Max Sliced-Wasserstein (Max-SW) distance~\cite{deshpande2019max}) uses only one important direction to distinguish the probability distributions. However, other potentially relevant directions are ignored in Max-SW. Therefore, Max-SW can miss some important differences between the two distributions in high dimension. We note that the linear projections in the Radon transform can be replaced by non-linear projections resulting in the generalized sliced-Wasserstein distance and its variants~\cite{beylkin1984inversion, kolouri2019generalized}. 

Apart from these main directions, there are also few proposals that try either to modify them or to combine the advantages of the above-mentioned approaches. In particular, Paty et al.~\cite{paty2019subspace} extended the idea of the max-sliced distance to the max-subspace distance by considering finding an optimal orthogonal subspace. However, this approach is computationally expensive, since it could not exploit the closed-form of the one-dimensional Wasserstein distance. Another approach named the Projected Wasserstein distance (PWD), which was proposed in~\cite{rowland2019orthogonal}, uses sliced decomposition to find multiple one-dimension optimal transport maps. Then, it computes the average cost of those maps equally in the original dimension.


\textbf{Our contributions.} Our paper also follows the slicing approach. However, we address key friction in this general line of work: how to obtain a relatively small number of slices simultaneously to maintain the computational efficiency, but at the same time, cover the major differences between two high-dimensional distributions. We take a probabilistic view of slicing by using a probability measure on the unit sphere to represent how important each direction is. From this viewpoint, SW uses the uniform distribution while Max-SW searches for the best delta-Dirac distribution over the projections, both can be considered as special cases. In this paper, we propose to search for an optimal distribution of important directions. We regularize this distribution such that it prefers directions that are far away from one another, hence encouraging an efficient exploration of the space of directions. In the case of no regularization, our proposed method recovers max-(generalized) SW as a special case. In summary, our main contributions are two-fold:

\vspace{-0.1em}
\begin{enumerate}
     \item First, we introduce a novel distance, named \emph{Distributional Sliced-Wasserstein distance} (DSW), to account for the issues of previous sliced distances.  Our main idea is to search for not just a single most important projection, but an \emph{optimal} distribution over projections that could balance between an expansion of the area around important projections and the informativeness of projections themselves, i.e., how well they can distinguish the two target probability measures. We show that DSW is a proper metric in the probability space and possesses appealing statistical and computational properties as the previous sliced distances. 
     \item Second, we apply the DSW distance to generative modeling tasks based on the generative adversarial framework. The extensive experiments on real and large-scale datasets show that DSW distance significantly outperforms the SW and Max-SW distances under similar computational time on these tasks. Furthermore, the DSW distance helps model distribution converge to the data distribution faster and provides more realistic generated images than the SW and Max-SW distances.
 \end{enumerate}
 \vspace{-0.1em}
 \textbf{Organization.} The remainder of the paper is organized as follows. In Section~\ref{Sec:background}, we provide backgrounds for Wasserstein distance and its slice-based versions. In Section~\ref{sec:dist_sliced_Wasserstein}, we propose distributional (generalized) sliced-Wasserstein distance and analyze some of its theoretical properties. Section~\ref{sec:experiments} includes  extensive experiment results followed by discussions in Section~\ref{sec:conclusion}. Finally, we defer the proofs of key results and extra materials in the Appendices. \\
 \textbf{Notation.} For any $\theta,  \theta' \in \mathbb{R}^{d}$, $\text{cos}(\theta,\theta') = \frac{\theta^{\top} \theta'}{\|\theta \| \| \theta'\|}$, where $\|.\|$ is $\ell_2$ norm. For any $d \geq 2$, $\mathbb{S}^{d-1}$ denotes the unit sphere in $d$ dimension in $\ell_2$ norm . Furthermore, $\delta$ denotes the  Dirac delta function, and $\langle \cdot,\cdot \rangle$ is the Euclidean inner-product. For any $p \geq 1$, $\mathbb{L}^{p}(\mathbb{R}^d)$ is the set of real-valued functions on $\mathbb{R}^{d}$ with finite $p$-th moment. 
\section{Background}
\label{Sec:background}
In this section, we provide necessary backgrounds for the (generalized) Radon transform, the Wasserstein, and sliced-Wasserstein distances. 
\subsection{Wasserstein distance}
We start with a formal definition of Wasserstein distance. For any $p \geq 1$, we define $\mathcal{P}_p (\mathbb{R}^{d})$ as the set of Borel probability measures with finite $p$-th moment defined on a given metric space $(\mathbb{R}^{d},\|.\|)$. For any probability measures $\mu,\nu$ defined on $\mathcal{X}, \mathcal{Y} \subseteq \mathbb{R}^{d}$, we denote their corresponding probability density functions as $I_\mu$ and $ I_\nu$. The Wasserstein distance of order
$p$ between $\mu$ and $\nu$ is given by~\cite{Villani-09, peyre2019computational}:
\begin{equation*}
    W_p(\mu,\nu) : = \Big{(} \inf_{\pi \in \Pi(\mu,\nu)} \int_{\mathcal{X} \times \mathcal{Y}} \| x - y\|^{p} d \pi(x,y) \Big{)^{\frac{1}{p}}},
\end{equation*}
where $\Pi(\mu,\nu)$ is a set of all transportation plans $\pi$ such that the marginal distributions of $\pi$ are $\mu$ and $\nu$, respectively. In order to simplify the presentation, we abuse the notation by using both $W_p(\mu,\nu) $ and $W_p(I_{\mu},I_{\nu})$ interchangeably for the Wasserstein distance between $\mu$ and $\nu$. 

When $\mu$ and $\nu$ are \textit{one-dimension} measures, the Wasserstein distance between $\mu$ and $\nu$ has a closed-form expression $W_p(\mu,\nu) =
    ( \int_0^1 |F_\mu^{-1}(z) - F_{\nu}^{-1}(z)|^{p} dz )^{1/p}$
where $F_{\mu}$ and $F_{\nu}$  are  the cumulative
distribution function (CDF) of $I_{\mu}$ and $I_{\nu}$, respectively.
\subsection{(Generalized) Radon transforms}
\label{sec:random_transform}
Now, we review (generalized) Radon transform maps, which are key to the notion of (generalized) sliced-Wasserstein distance and its variants. The \emph{Radon transform} (RT) maps a function $I \in \mathbb{L}^1(\mathbb{R}^d)$ to the space of functions defined over space of lines in $\mathbb{R}^d$. In particular, for any  $t \in \mathbb{R}$ and direction $\theta \in \mathbb{S}^{d-1}$, the RT is defined as follows~\cite{helgason2010integral} :
$
    \mathcal{R}I(t,\theta) : = \int_{\mathbb{R}^d} I(x)\delta (t - \langle x,\theta \rangle) dx.
$

The \textit{generalized} Radon transform (GRT) \cite{beylkin1984inversion} extends the original one from integration over hyperplanes of $\mathbb{R}^d$ to integration over hypersurfaces. In particular, it is defined as:
$
    \label{eq:generalized_transform}
    \mathcal{G}I(t,\theta) := \int_{\mathbb{R}^d} I(x) \delta(t-g(x,\theta))dx,
$
where $t \in \mathbb{R}$ and $\theta \in \Omega_\theta$. Here, $\Omega_{\theta}$ is a compact subset of $\mathbb{R}^{d}$ and $g: \mathbb{R}^{d} \times \mathbb{S}^{d- 1} \mapsto \mathbb{R}$ is a defining function (cf. Assumptions H1-H4 in~\cite{kolouri2019generalized} for the definition of defining function) inducing the hypersurfaces. When $g(x, \theta) = \langle x,\theta \rangle$ and $\Omega_\theta = \mathbb{S}^{d-1}$, the generalized Radon transform becomes the standard Radon transform. 
\subsection{(Generalized) sliced-Wasserstein distances}
The sliced-Wasserstein distance (SW) between two probability measures $\mu$ and  $\nu$ is defined as~\cite{bonneel2015sliced} :
\begin{align*}
    SW_p(\mu,\nu) : = \biggr(\int_{\mathbb{S}^{d-1}} W_p^p \big(\mathcal{R}I_{\mu}(\cdot,\theta),\mathcal{R}I_{\nu}(\cdot,\theta )\big) d\theta \biggr)^{1/ p}.
\end{align*}
Similarly, the generalized sliced-Wasserstein distance~\cite{kolouri2019generalized} (GSW) is given by $\text{GSW}_p(\mu,\nu) : = \left( \int_{\Omega_\theta} W_p^p \big(\mathcal{G}I_{\mu}(\cdot,\theta),\mathcal{G}I_{\nu}(\cdot,\theta )\big) d\theta \right)^{1/p}$, where $\Omega_{\theta}$ is the compact set of feasible parameter. However, these integrals are usually intractable. Thus, they are often approximated by using Monte Carlo scheme to draw uniform samples $\{\theta_i\}_{i = 1}^{N}$ from
$\mathbb{S}^{d-1}$ and $\Omega_\theta$. In particular, $SW_p^{p}(\mu,\nu) \approx \frac{1}{N} \sum_{i=1}^N W_p^p\big(\mathcal{R}I_{\mu} (\cdot,\theta_i),\mathcal{R}I_{\nu} (\cdot,\theta_i)\big)$ and $\text{GSW}_p^{p}(\mu,\nu) \approx \frac{1}{N} \sum_{i=1}^N W_p^p\big(\mathcal{G}I_{\mu} (\cdot,\theta_i),\mathcal{G}I_{\nu} (\cdot,\theta_i)\big)$. In order to obtain a good approximation of (generalized) SW distances, $N$ needs to be  sufficiently large. However, important directions are not distributed uniformly over the sphere. Thus, this approach will draw potentially many unimportant projections that are not only expensive but also greatly reduce the effect of the SW
distance. 
\subsection{Max (generalized) sliced-Wasserstein distances}
 An approach to using only informative directions is to simply take the best slice in discriminating two given probability distributions. That distance is max sliced-Wasserstein distance (Max-SW)~\cite{deshpande2019max}, which is given by:
 \begin{align*}
    &\text{max} SW_p(\mu,\nu) := \max_{\theta \in \mathbb{S}^{d-1}} W_p (\mathcal{R}I_{\mu}(\cdot,\theta),\mathcal{R}I_{\nu}(\cdot,\theta)).
\end{align*}
By combining this idea with non-linear projections from generalized Radon transform, we obtain max generalized sliced-Wasserstein distance (Max-GSW)~\cite{kolouri2019generalized}. The formal definition of that distance is: $\text{max} GSW_p(\mu,\nu) := \max_{\theta \in \Omega_\theta} W_p (\mathcal{G}I_{\mu}(\cdot,\theta),\mathcal{G}I_{\nu}(\cdot,\theta))$. The (generalized) Max-SW distances focus on finding only the most important direction. Meanwhile, other informative directions play no role in the distance. Therefore, (generalized) Max-SW distances can ignore useful information about the structure of high dimensional probability measures.
\section{Distributional Sliced-Wasserstein Distance}
\label{sec:dist_sliced_Wasserstein}
With the aim of improving the limitations of the previous sliced distances, we propose a novel distance, named \emph{Distributional Sliced-Wasserstein distance} (DSW), that can search for not just a single but a distribution of important directions on the unit sphere. We prove that it is a well-defined metric and discuss its connection to the existing sliced-based distances in Section~\ref{sec:definition}. Then, we provide a procedure to approximate DSW based on its dual form in Section~\ref{subsec:computation_DSW}.
\subsection{Definition and metricity}
\label{sec:definition}
We first start with a definition of distributional sliced-Wasserstein distance. We say $C > 0$ \emph{admissible} if the set  $\conspace_{C}$ of probability measures $\mathcal{\sigma}$ on $\mathbb{S}^{d-1}$ satisfying $\mathbb{E}_{\theta,\theta' \sim \mathcal{\sigma}}\left[|\theta^\top \theta'|\right] \leq  C$ is not empty.
\begin{definition} \label{def:DSW}
Given two probability measures $\mu$ and $\nu$ on $\mathbb{R}^d$ with finite $p$-th moments where $p \geq 1$ and an admissible regularizing constant $C>0$. The \emph{distributional sliced-Wasserstein distance} (DSW) of order $p$ between $\mu$ and $\nu$ is given by:
\begin{align}
    \dislice_{p}(\mu,\nu; C) := \sup_{\mathcal{\sigma} \in \conspace_{C}} \biggr( \mathbb{E}_{\theta \sim \sigma} \biggr[ W_p^p(\mathcal{R}I_{\mu}(\cdot,\theta),\mathcal{R}I_{\nu}(\cdot,\theta)) \biggr] \biggr)^{\frac{1}{p}}, \label{eq:dist_slice_ori}
\end{align}
where $\mathcal{R}$ is the Radon transform operator.
\end{definition}
The DSW aims to find the optimal probability measure of slices on the unit sphere $\sphere^{d-1}$. Note that, the Max-SW distance is equivalent to searching for the best Dirac measure on a single point in $\mathbb{S}^{d-1}$, which puts all weights in only one direction. Meanwhile,  the uniform measure in the formulation of SW distance distributes the same weights in all directions. Indeed, the uniform and Dirac measures are two 
special cases, because they view that  either all directions are equally important or only one direction is important. That view is too restricted if the data actually lie on low dimensional space. Thus,  we aim to find a probability measure which concentrates only on areas around important directions. Furthermore, we do not want these directions to lie in only one small area, because  under the orthogonal projection of RT, their corresponding one-dimensional distributions will become similar.   In order to achieve this, we search for an optimal measure $\sigma$ that satisfies the regularization constraint
$\mathbb{E}_{\theta, \theta^{'}\sim \sigma} [\big| \theta^{\top} \theta^{'} \big|] \leq C$. By Cauchy-Schwarz inequality, $C$ is no greater than $1$, thus $\mathbb{M}_1$ contains all probability measures on the unit sphere. Optimizing over $\mathbb{M}_1$ simply returns the best Dirac measure corresponding to the Max-SW distance. When $C$ is 
 small, the constraint forces the measure $\sigma$ to distribute more weights to directions that are far from each other (in terms of their angles). Thus, a small appropriate value of $C$  will help to balance between the distinctiveness and informativeness of these targeted directions. For further discussion about $C$, see Appendix~\ref{subsec:constraint_DSW}. 

Next, we show that DSW is a well-defined metric on the probability space. 
\begin{theorem}\label{theorem:DSW-distance}
For any $p \geq 1$ and admissible $C>0$, $\text{DSW}_p(\cdot,\cdot;C)$ is a well-defined metric in the space of Borel probability measures with finite $p$-th moment. In particular, it is non-negative, symmetric, identity, and satisfies the triangle inequality.
\end{theorem}
The proof of Theorem~\ref{theorem:DSW-distance} is in Appendix~\ref{sec:proof:theorem:DSW-distance}. Our next result establishes the topological equivalence between DSW distance and (max)-sliced Wasserstein and Wasserstein distances.
\begin{theorem}
\label{theorem:property_DSW}
For any $p \geq 1$ and admissible $C>0$, the following holds
\begin{itemize}
    \item[(a)] $DSW_{p}(\mu, \nu; C) \leq \text{maxSW}_{p}(\mu, \nu) \leq W_{p}(\mu, \nu)$.
    \item[(b)] If $C \geq 1/ d$, we have $DSW_{p}(\mu, \nu; C) \geq \left(\frac{1}{d}\right)^{1/p} \text{maxSW}_{p}(\mu, \nu) \geq \left(\frac{1}{d}\right)^{1/p} \text{SW}_{p}(\mu, \nu)$.
\end{itemize}
As a consequence, when $p \geq 1$ and $C \geq 1/d$, $DSW_{p}(\cdot,\cdot;C)$, $\text{SW}_{p}$, $\text{maxSW}_{p}$, and $W_{p}$ are topologically equivalent, namely, the convergence of probability measures under $DSW_{p}(.,.;C)$ implies the convergence of these measures under other metrics and vice versa. 
\end{theorem}
The proof of Theorem~\ref{theorem:property_DSW} is in Appendix~\ref{sec:proof:theorem:property_DSW}. As a consequence of Theorem~\ref{theorem:property_DSW}, the statistical error of estimating the unknown distribution based on the empirical distribution of $n$ i.i.d data under DSW distance is $C_{d} \cdot n^{-1/2}$ with high probability where $C_{d}$ is some universal constant depending on dimension $d$ (see Appendix~\ref{subsec:stats_DSW}). Therefore, as other sliced-based Wasserstein distances, the DSW distance does not suffer from the curse of dimensionality.
\subsection{Computation of DSW distance} 
\label{subsec:computation_DSW}
Direct computation of DSW distance is challenging. Hence we consider a dual form of DSW distance and a reparametrization of $\sigma$  as follows. 
\begin{definition}
\label{def:dual_DSW}
For any $p \geq 1$ and admissible $C>0$, there exists a non-negative constant $\lambda_{C}$ depending on $C$ such that the dual form of DSW distance takes the following form
\begin{align*}
    \dislice_{p}^{*}(\mu,\nu; C) 
    & = \sup_{\mathcal{\sigma} \in \conspace} \Bigg\{ \biggr( \mathbb{E}_{\theta \sim \sigma} \biggr[ W_p^p(\mathcal{R}I_{\mu}(\cdot,\theta),\mathcal{R}I_{\nu}(\cdot,\theta)) \biggr] \biggr)^{\frac{1}{p}} - \lambda_{C} \mathbb{E}_{\theta,\theta' \sim \mathcal{\sigma}}\left[|\theta^\top \theta'|\right] \Bigg\} + \lambda_{C} C, \nonumber
\end{align*}
where $\conspace$ denotes the space of all probability measures on the unit sphere $\sphere^{d-1}$.
\end{definition}
By the Lagrangian duality theory, $\dislice_{p}(\mu,\nu; C) \geq \dislice_{p}^{*}(\mu,\nu; C)$ for any $p \geq 1$ and admissible $C > 0$. In Definition 
\ref{def:dual_DSW}, the set $\mathbb{M}_C$ disappears and $\lambda_C$ plays the tuning role for the regularized term $\mathbb{E}_{\theta,\theta^{\prime}\sim \sigma}\big[ |\theta^{\top} \theta^{\prime} | \big]$. When $\lambda_C$ is large, $\mathbb{E}_{\theta,\theta^{\prime}\sim \sigma}\big[ |\theta^{\top} \theta^{\prime} | \big]$ needs to be small, meaning that $C$ is small. When $\lambda_C$ is small, 
the value of $\mathbb{E}_{\theta,\theta^{\prime}\sim \sigma}\big[ |\theta^{\top} \theta^{\prime} | \big]$ becomes less important, i.e., $C$ is large.  

For reparametrizing the measure  $\sigma$, we use a pushforward of uniform measure on the unit sphere through some measurable function $f$. In particular, let $f$ be a Borel measurable  function from $\mathbb{S}^{d-1}$ to $\mathbb{S}^{d-1}$. For any Borel set $A \subset \mathbb{S}^{d-1}$, we define $ \sigma(A) = \sigma^{d-1}(f^{-1}(A)),$
where $\sigma^{d-1}$ is the uniform probability measure on $\mathbb{S}^{d-1}$. Then for any Borel measurable function $g:  \mathbb{S}^{d-1} \rightarrow \mathbb{R}$, we have $\int_{\theta \sim \sigma} g(\theta) d\sigma(\theta) = \int_{\theta \sim \sigma^{d-1}} (g\circ f)(\theta) d\sigma^{d-1}(\theta)$. Therefore, we obtain the equivalent dual form of DSW as follows:
\begin{align}
    \dislice_{p}^{*}(\mu,\nu; C) = \sup_{f\in \mathcal{F}}\Bigg\{ \biggr( \mathbb{E}_{\theta \sim \sigma^{d-1}} \big[W_p^p\big(\mathcal{R}I_{\mu}(\cdot, f(\theta)), \mathcal{R}I_{\nu}(\cdot, f(\theta))\big)  \big] \biggr)^{1/ p}  & \label{eq:dual_dsw_equiv} \\
 & \hspace{-18 em} - \lambda_{C} \mathbb{E}_{\theta,\theta^{'} \sim \sigma^{d-1}}\Big[\big|f(\theta)^{\top} f(\theta^{'}) \big|\Big] \Bigg\} + \lambda_{C} C : = \sup_{f \in \mathcal{F}} \text{DS}(f), \nonumber
\end{align}
where $\mathcal{F}$ is a class of all Borel measurable functions from $\sphere^{d-1}$ to $\sphere^{d-1}$.


\textbf{Finding the optimal $f$:} We parameterize $f$ in the dual form~(\ref{eq:dual_dsw_equiv}) by using a deep neural network with parameter $\phi$, defined as $f_{\phi}$.  Then, we estimate the gradient of the objective function $\text{DS}(f_{\phi})$ in equation~(\ref{eq:dual_dsw_equiv}) with respect to $\phi$ and use stochastic gradient ascent algorithm to update $\phi$. Since there are expectations over uniform distribution in the gradient of $\text{DS}(f_{\phi})$, we use the Monte Carlo method to approximate these expectations. Note that, we can use the fixed point from the stochastic ascent algorithm to approximate the dual value of DSW in equation~(\ref{eq:dual_dsw_equiv}). A detailed argument for this point is in Appendix~\ref{subsec:gradient_DSW}. Finally, in generative model applications with DSW being the loss function, we only need to use the gradient of the function $\text{DS}(.)$ to update the parameters of interest. Therefore, we can treat $\lambda_{C}$ as a regularized parameter and tune it to find suitable value in these applications. 

\begin{figure}[!t]
\begin{center}

  \begin{tabular}{c}
    \widgraph{0.95\textwidth}{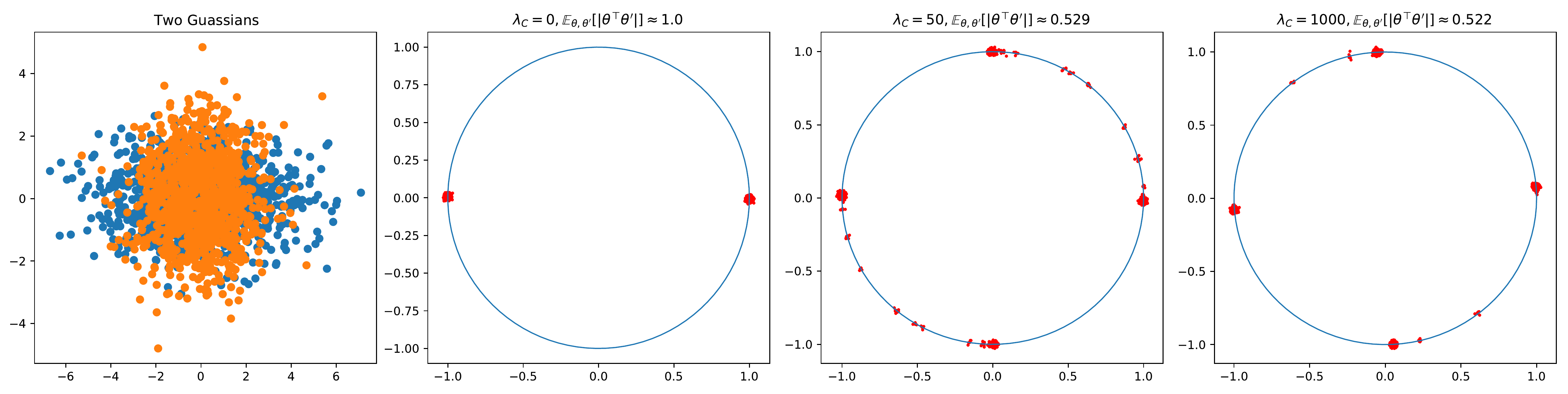}
   
  \end{tabular}
 \end{center}
  \caption{
  \footnotesize{
  Empirical behavior of optimal measure $\sigma$, approximated by 1000 samples, on a circle for different values of $\lambda_{C}$ (the constant in the dual form of DSW in Definition \ref{def:dual_DSW}) when $\mu$ and $\nu$ are bivariate Gaussian distributions sharing the same eigenvectors. When $\lambda_C=0$, $C = 1$. When $\lambda_{C}$ increases, $C$ becomes small.
    }}
  \label{fig:guassians}
\vspace{-0.2 em}
\end{figure}

\textbf{Illustration of the roles of $\lambda_C$ and $C$:} To illustrate the roles of $\lambda_C$ and $C$ in finding optimal distribution $\sigma$, we conduct a simple experiment on two Gaussian distributions with zero means and covariance matrices given by $ \begin{pmatrix} 2&0\\0&2 \end{pmatrix}$ and $\begin{pmatrix} 5 & 0 \\ 0 & 1\end{pmatrix}$. The experiment  optimizes the empirical form of Definition \ref{def:dual_DSW} with different choices of $\lambda_C$. The results  are  shown in Figure~\ref{fig:guassians} with the reported value of $\lambda_C$ and $\mathbb{E}_{\theta,\theta^{\prime}\sim \sigma}\big[ |\theta^{\top} \theta^{\prime} | \big]$. For  $\lambda_C=0$, the obtained distribution concentrates only on one direction. When $\lambda_C = 50$, optimal $\sigma$ distributes more weights to other directions on the circle. When $\lambda_C = 1000$, optimal $\sigma$ 
is close to the discrete distribution concentrated on two eigenvectors of the covariance matrices, 
which are the main directions differentiating the two Gaussian distributions. 

\paragraph{Extension of DSW and comparison of DSW to Max-GSW-NN:} Similar to SW, we extend DSW to distributional generalized sliced Wasserstein (DGSW) by using the non-linear projecting operator via GRT. The definition of the DGSW and its properties are in Appendix~\ref{sec:extension_DGSW}. Finally, in Appendix~\ref{subsec:exp_generative_models}, we show the distinction of the DSW to Max-GSW-NN~\cite{kolouri2019generalized} when the neural network defining function in Max-GSW-NN is $g(x,\theta) = \langle x,f(\theta) \rangle$ where $f: \mathbb{S}^{d-1} \rightarrow \mathbb{S}^{d-1}$.

\section{Experiments}
\label{sec:experiments}
In this section, we conduct extensive experiments comparing the performance in both generative quality and computational speed of the proposed DSW  distance with other sliced-based distances, namely the SW, Max-SW,  Max-GSW-NN~\cite{kolouri2019generalized} and  projected robust subspace Wasserstein (PRW) ~\cite{paty2019subspace,lin2020projection}
using the minimum expected distance estimator (MEDE)~\cite{bernton2019parameter} on  MNIST \cite{lecun1998gradient}, CIFAR10 \cite{Krizhevsky09learningmultiple}, CelebA \cite{liu2015faceattributes} and LSUN~\cite{yu15lsun} datasets.  The details of the MEDE framework are described in Appendix~\ref{sec:application}. On MNIST dataset, we train generative models with different distances and then evaluate their performances by comparing Wasserstein-2 distances between 10000 random generated images and all images from the MNIST test set. Due to the very large size of other datasets, e.g., 3 million images in LSUN,  it is expensive to compute empirical Wasserstein-2 distance as its complexity is of order $\mathcal{O}(k^2 \log k)$ where $k$ is the number of support points. Therefore, after we train generative models, we use FID score~\cite{heusel2017gans} to evaluate the generative quality of these generators. 
 The FID score is calculated from  $10000$ random generated images and all training samples using precomputed statistics in~\cite{heusel2017gans}. Finally, for $\lambda_C$ in DSW (see Definition~\ref{def:dual_DSW}), it is chosen in the set  $\{1,10,100,1000\}$ such that its  Wasserstein-2 (FID score) (between 10000 random generated images and all images from corresponding validation set) is the lowest among the four values. Detailed experiment settings are in Appendix~\ref{sec:setting}.


\subsection{Results on MNIST}


    
\begin{figure}[!t]
\begin{center}

  \begin{tabular}{cc}
\widgraph{0.35\textwidth}{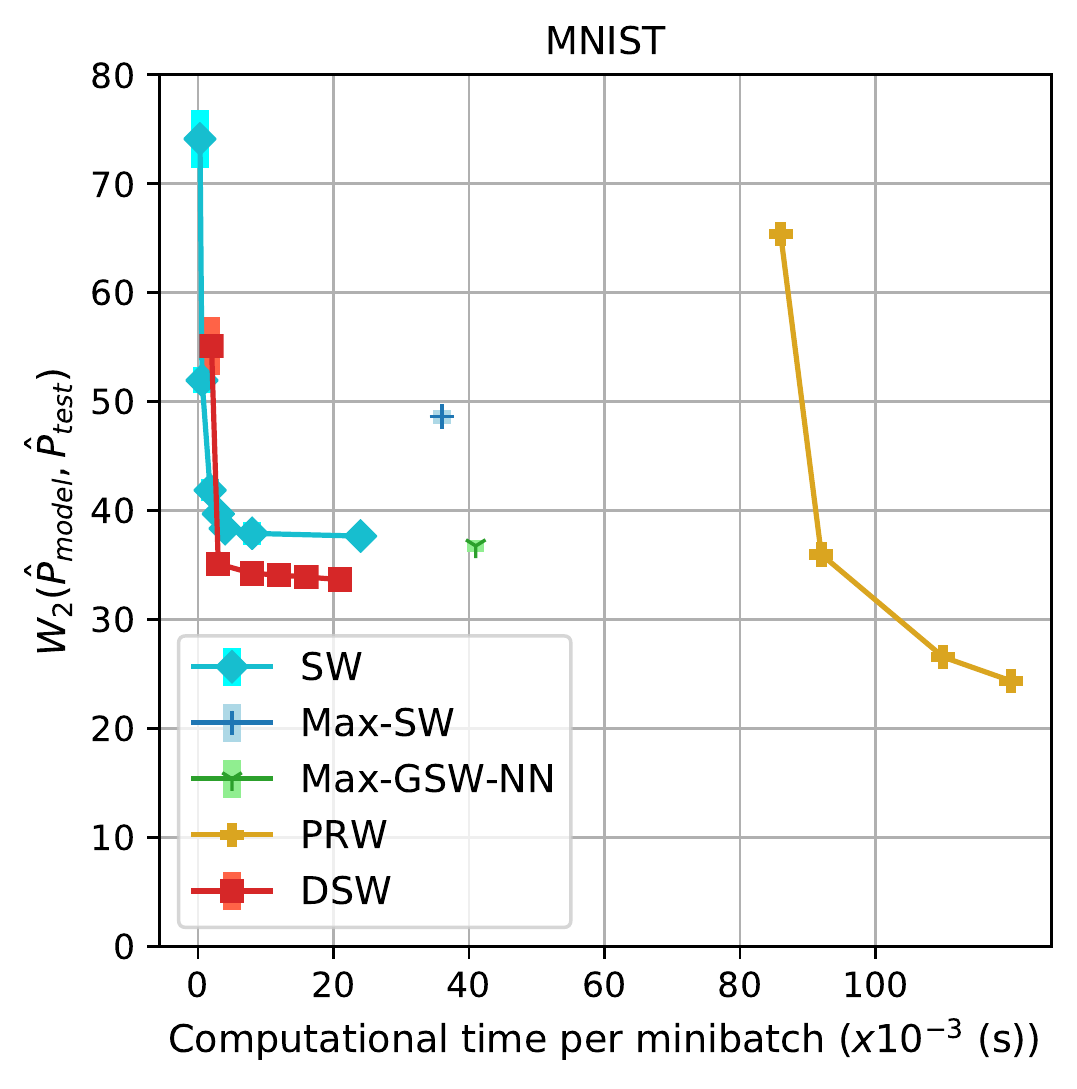} \label{fig:sub1}
&
     \widgraph{0.35\textwidth}{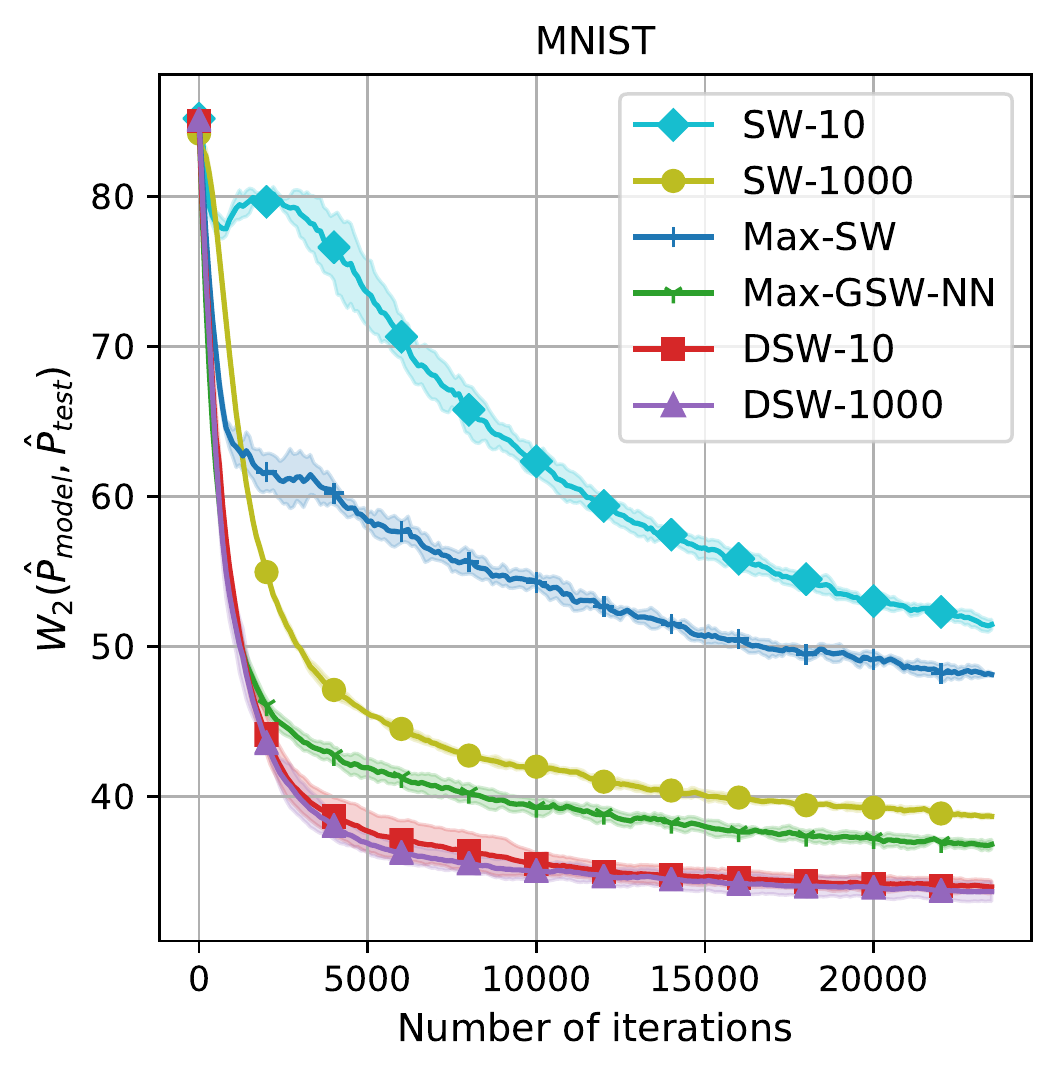} \label{fig:sub2}\\
     (a)&(b)
    \end{tabular}
    \begin{tabular}{cc}
    \widgraph{0.35\textwidth}{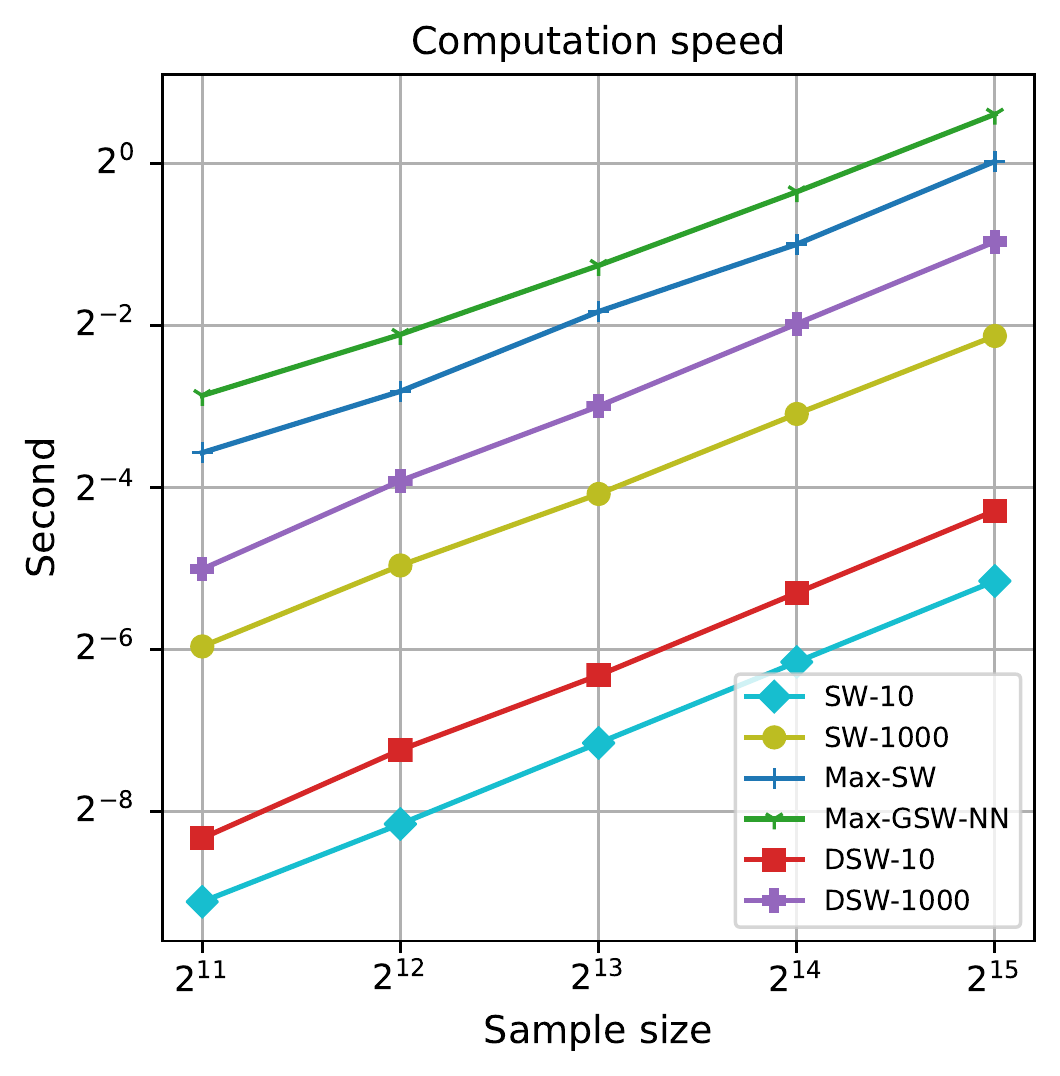} \label{fig:sub3}

  &
    \widgraph{0.35\textwidth}{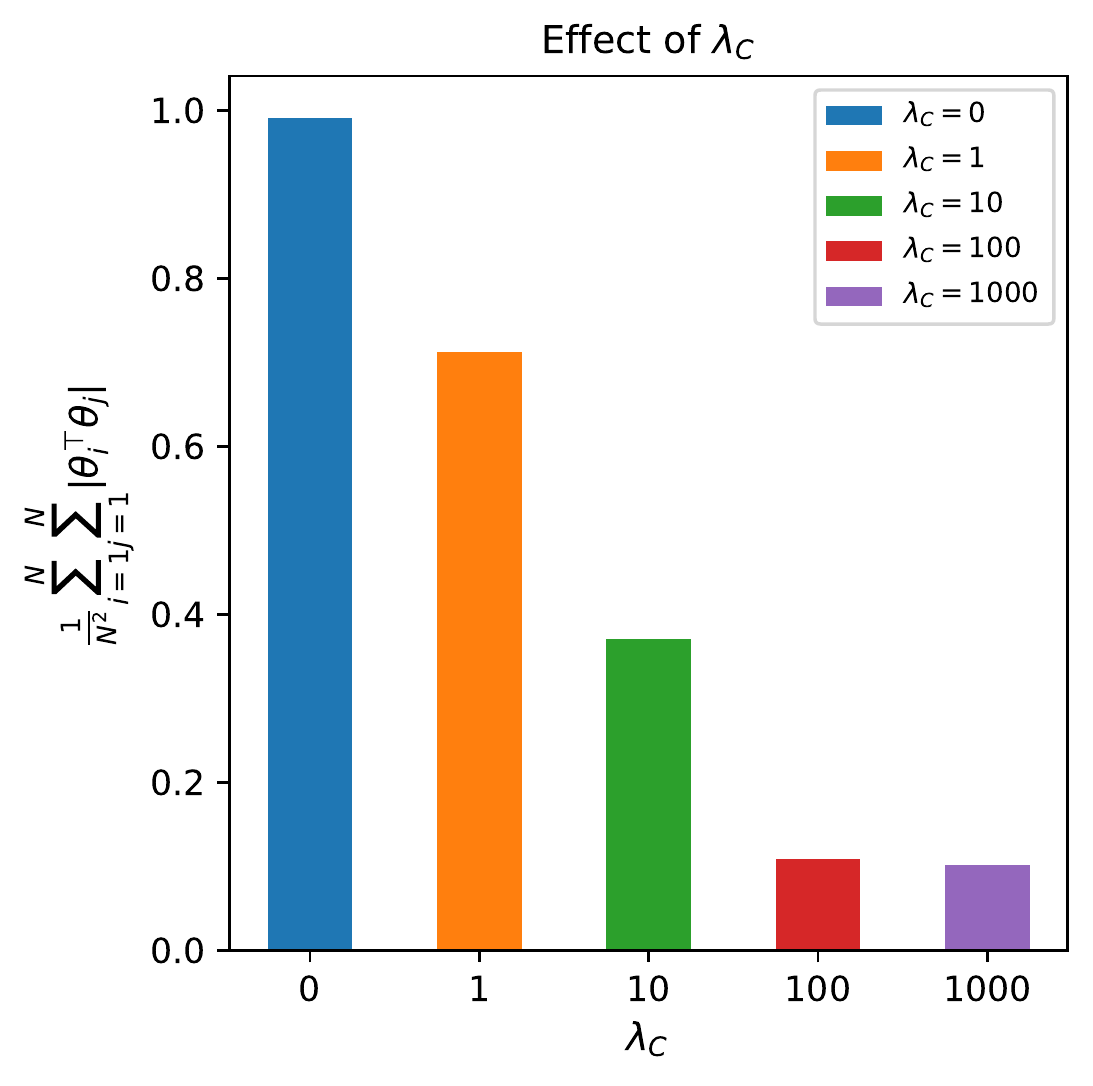} \label{fig:sub4}
     \\
     (c)& (d)
    
  \end{tabular}
  \end{center}
  \caption{
  \footnotesize{(a) Comparison between DSW, SW, Max-SW and Max-GSW-NN based on execution time and performance. Here, each dot of SW and DSW corresponds to the number of projections chosen in $\{1,10,10^2,5\times10^2,10^3,5\times 10^3,10^4\}$. Each dot of PRW corresponds to the dimension of the subspace chosen in $\{2,5,10,50 \}$; (b) Comparison between SW, DSW, Max-SW and Max-GSW-NN based on Wasserstein-2 distance
between distributions of learned model and test set over iterations; (c)  Computation speed of distances based on the number of minibatch's samples (log-log scale); (d) Effect of $\lambda_C$ on the mean of absolute values of pairwise cosine similarity between 10 random directions from the found distribution $\sigma$ of DSW.}
}
  \label{fig:MNISTgraphs}
\end{figure}

\textbf{Generative quality and computational speed:} We report the performance of the learned generative models for MNIST in Figure~\ref{fig:MNISTgraphs}(a). To plot this figure, we vary the number of projections $N \in \{1,10,10^2,5\times 10^2,10^3,5\times 10^3,10^4\}$ for the SW, and $N \in \{1,10,10^2,5\times10^2,10^3,5\times10^3\}$ for the DSW. Then we measure the computational time per minibatch and the Wasserstein-2 score of the learned generators for each $N$. We plot the Wasserstein-2 score and computational time of Max-SW and Max-GSW-NN in their standard settings~\cite{kolouri2019generalized}. Except for the regime with very fast but low-quality learned models, DSW is better than all the existing slice-based baselines in terms of both model quality and computational speed. Moreover, DSW can learn good models with very few projections, e.g., DSW-10 achieves better model quality than Max-GSW-NN and Max-SW and is one order-of-magnitude faster than these sliced distances. Finally, with a similar computational time, a learned generator by DSW has the Wasserstein-2 score that is roughly $10\%$ lower than the one got from SW. For the qualitative comparison between these distances, we show random generated images from their generative models in Figure~\ref{fig:MNISTgenimages} in Appendix~\ref{subsec:exp_generative_models}. We observe that generated images from DSW are sharper and easier to classify into numbers than those from other baseline distances.

\vspace{0.5 em}
\noindent
\textbf{Comparison with projected robust subspace Wasserstein (PRW):} In Figure~\ref{fig:MNISTgraphs}(a), we plot the Wasserstein-2 score and computational time of PRW, where the subspace dimension of PRW varies in the range $\{2,5,10,50 \}$. PRW is able to improve upon the model quality of slice-based methods including DSW, however at the cost of being an order of magnitude slower than DSW with 10 projections (DSW-10). We observe that DSW-10 obtains a better Wasserstein-2 score than PRW with 5-dimensional subspace, while its corresponding computational time is 30 times faster than that of PRW-5. Using 50 dimension, PRW's Wasserstein-2 score improves about $29\%$ to that of DSW-10 but the computational cost is also around 40 times slower. The main computational advantage of DSW comes from the exact calculation of Wasserstein distance in one-dimension. The visual comparison between PRW and DSW based on their generated images is in Figure~\ref{fig:PRWimages} in Appendix~\ref{subsec:prw}.

\vspace{0.5 em}
\noindent
\textbf{Convergence behavior:} Figure~\ref{fig:MNISTgraphs}(b) shows that DSW learns better models at a faster speed of convergence than other baseline distances with a very small number of projections, e.g., DSW-10 is the second lowest curve compared to curves from other sliced-based distances. 

\vspace{0.5 em}
\noindent
\textbf{Scalability over sample size of minibatch: } Results in Figure~\ref{fig:MNISTgraphs}(c) show that DSW has a computational complexity of the order $\mathcal{O}(k \log k)$, which is similar to those of other sliced-based distances, where $k$ is the number of samples per batch. 
 \vspace{0.5 em}
 \noindent
 \textbf{Effect of the regularization parameter $\lambda_C$:} For each value of $\lambda_C \in \{1,10,100,1000\}$, we find the optimal distribution $\sigma$ of DSW with $N = 10$ projections, and then calculate $A_{N} = \frac{1}{N^2}\sum_{i, j = 1}^N |\theta_i^\top \theta_j|$, an approximation of the regularized term $\mathbb{E}_{\theta,\theta' \sim \mathcal{\sigma}}\left[|\theta^\top \theta'|\right]$ in the dual form of DSW in equation~(\ref{eq:dual_dsw_equiv}), where $\{\theta_{i}\}_{i = 1}^{N} \sim \sigma$. The results are shown in Figure~\ref{fig:MNISTgraphs}(d).  We observe that when $\lambda_C$ 
increases, $A_{N}$ goes down. When $\lambda_C =0$, i.e., no regularization, $A_{N}$ gets close to 1, meaning that all 
projected directions collapse to one direction.  When $\lambda_C =1000$,  $A_{N}$ is close to 0.1, suggesting that all projected directions are nearly orthogonal. 

\vspace{0.5 em}
\noindent
\textbf{Additional experiments: } We also investigate how the number of gradient-steps used for updating distribution of directions $\sigma$, and how the size of minibatches affects the quality of DSW (see Appendix~\ref{subsec:exp_generative_models}).  The results show that an increasing number of gradient steps to update $\sigma$ leads to better performance of DSW but also slows down the computation speed. Furthermore, we carry out experiments with DGSW, an extension of DSW to non-linear projections,  and test the new proposed distances in training encoder-generator models on MNIST using joint contrastive inference (JCI) in Appendices~\ref{subsec:exp_generative_models} and~~\ref{subsec:exp_joint_contrastive}. The description of these models is in Appendix~\ref{sec:application}.
 

\subsection{Results on Large-scale Datasets}
\label{sec:largescale_exp}
Next, we conduct large-scale experiments on a range of more realistic image datasets. We train generative models using CIFAR10, CelebA, and LSUN datasets (all these datasets are rescaled to 64x64 resolution).
When working with high dimensional distributions,~\cite{deshpande2018generative} proposed a trick to improve the quality of the generator by learning a feature function which maps data to a new feature space that is more manageable in size. When the feature function is fixed, the generator is trained to match the distribution of features. When the generator is fixed, the feature function tries to tease apart the data empirical features from the generated feature distribution. For the experiments in this section, we use the same technique with DSW and all other baseline distances.

\begin{figure}[t]
\begin{center}

  \begin{tabular}{ccc}
    \widgraph{0.29\textwidth}{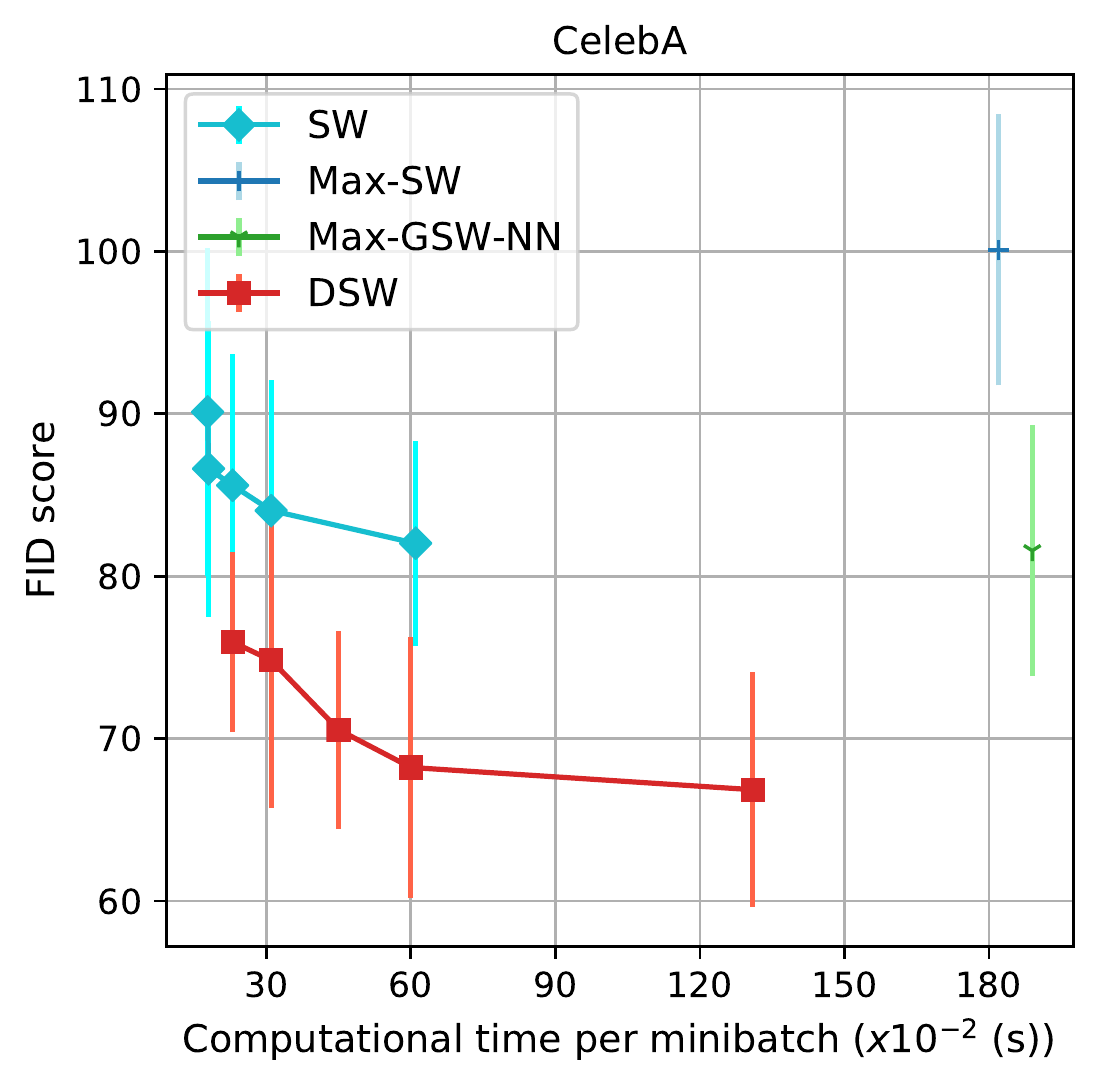}
    &\widgraph{0.29\textwidth}{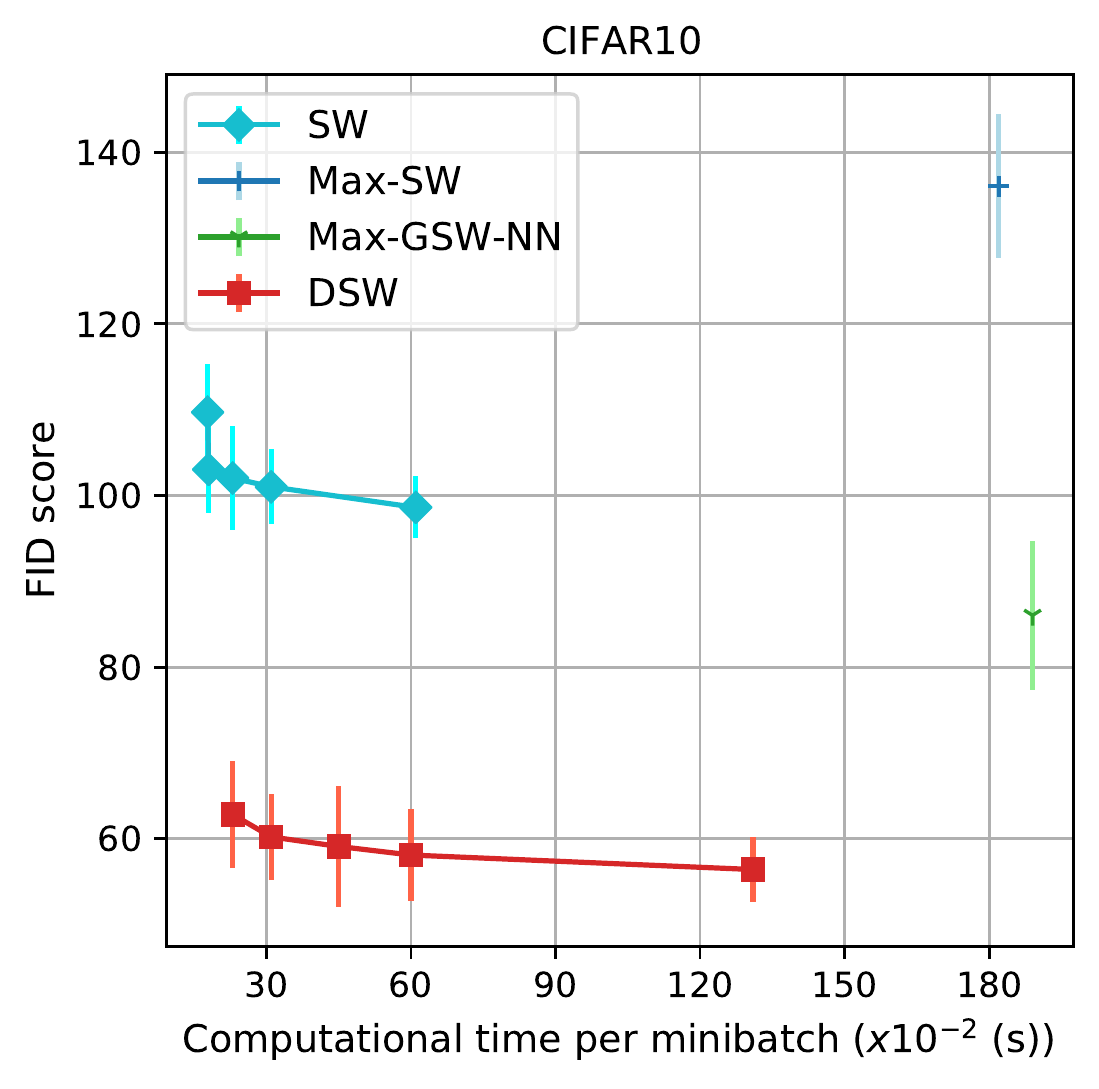}
    &\widgraph{0.29\textwidth}{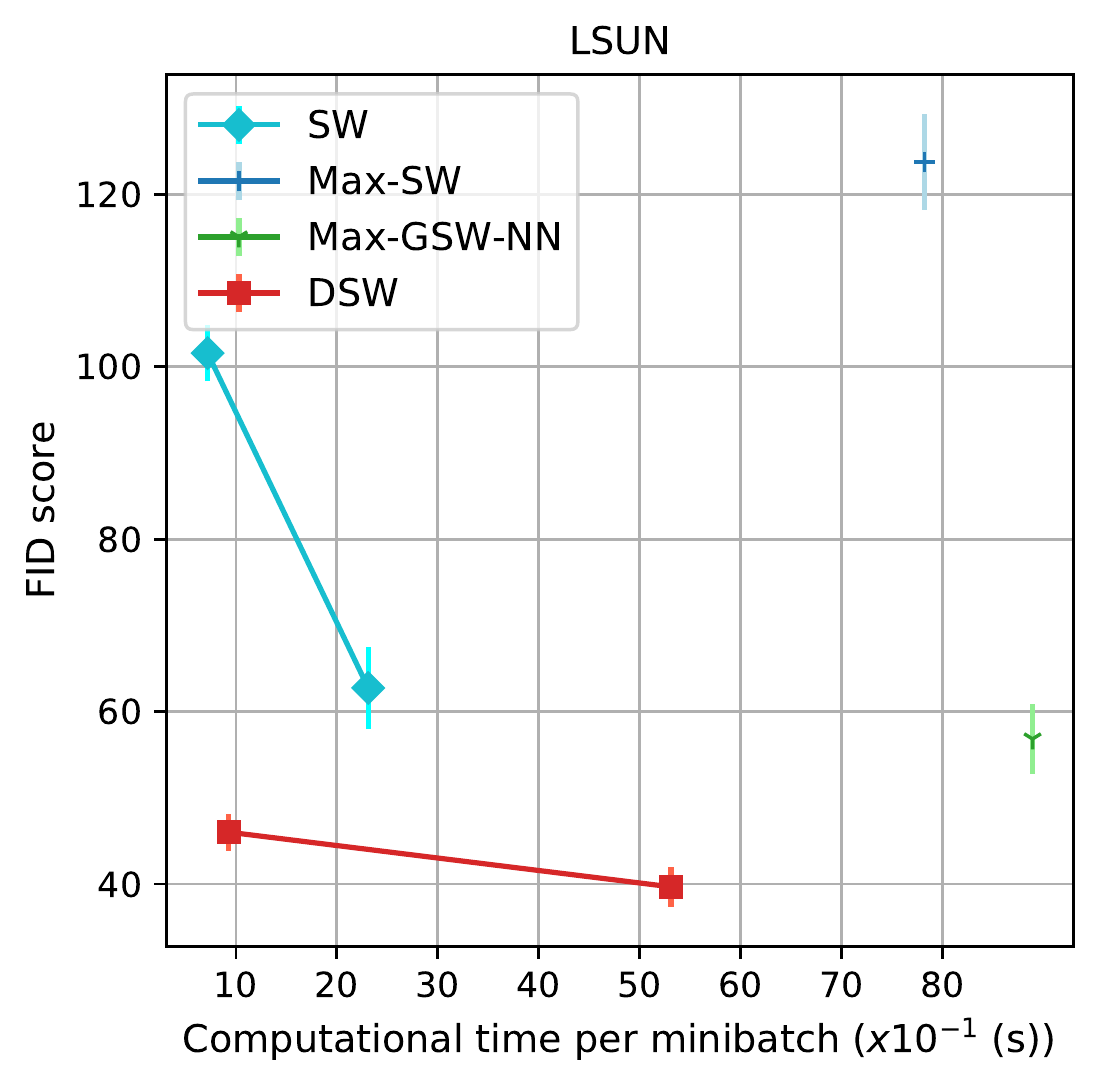}

  \end{tabular}
  \end{center}
  \caption{\footnotesize{
  Comparison between DSW, SW, Max-SW and Max-GSW-NN in  terms of execution time and performance. Here, each dot of SW and DSW corresponds to the number of projections chosen in $\{10^2,5\times10^2,10^3,5\times10^3,10^4\}$. We set the minibatch size be 512 on CelebA and CIFAR, and be 4096 on LSUN.
    }}
  \label{fig:DSW_SW_comparison}
\end{figure}

We compare DSW with SW, Max-SW, and Max-GSW-NN in both generative quality (FID score) and computational time in Figure~\ref{fig:DSW_SW_comparison}. We could not compare DSW with PRW on the large-scale datasets since PRW is computationally expensive to train to obtain good generated images. On CelebA and CIFAR10, we let $N$, the number of projections of both DSW and SW, vary in the set $\{10^2,5\times10^2,10^3,5\times10^3,10^4\}$. 
For LSUN, since it takes considerably longer time to train each model, we only vary $N$ in the set $\{10^2,10^4\}$. 
On all these large datasets, DSW outperforms all the other baselines in both FID score of the learned model and computational efficiency.
The gap of FID scores between DSW and other methods is especially large on CIFAR10 and LSUN. 
For example, on CIFAR10, with the same computational time, FID scores of DSW are always lower than those of SW about 20 units. 
On LSUN, with 100 projections, DSW can achieve an FID score of 46 while SW with 10000 projections still has a worse FID score of over 60. 
It is interesting to note that on these high-dimensional datasets, Max-SW performs rather poorly: it obtains the highest FID scores among all distances while requires heavy computation. Max-GSW-NN has better FID scores than (Max)-SW; however, it is still worse than DSW and while being slower. This is consistent with the intuition that as the number of dimension of the data grows, the use of a single important slice in Max-SW becomes a less efficient approximation. DSW, on the other hand, is able to make use of more important slices, and at the same time avoids SW's inefficiency of uniform slice-sampling.

Generated images from CelebA, CIFAR10 and LSUN are deferred to Appendix~\ref{subsec:exp_generative_models}. Comparing to other sliced-Wasserstein distances, generated samples obtained from the DSW's generative model are also more visually realistic. Further experiments to compare DGSW with GSW, Max-GSW, and Max-GSW-NN are also given in the Appendix~\ref{subsec:exp_generative_models}. Based on these experiments, we can conclude that the distributional approach also improves the generative quality of non-linear slicing distances.
\section{Conclusion}
\label{sec:conclusion}
In this paper, we have presented the novel distributional sliced-Wasserstein (DSW) distances between two probability measures. Our main idea is to search for the best distribution of important directions while regularizing towards orthogonal directions. We prove that they are well-defined metrics and provide their theoretical and computational properties. We compare our proposed distances to other sliced-based distances in a variety of generative modeling tasks, including estimating generative models and jointly estimating both generators and inference models. Extensive experiments demonstrate that our new distances yield significantly better models and convergence behaviors during training than the previous sliced-based distances. One important future direction is to investigate theoretically the optimal choice of the regularization parameter $\lambda_{C}$ such that the DSW distance can capture all the important directions that can distinguish two target probability measures well.

\bibliographystyle{abbrv}
\bibliography{iclr2021_conference}

\newpage
\appendix
\begin{center}
\textbf{\Large{Supplement to ``Distributional Sliced-Wasserstein and Applications to Generative Modeling''}}
\end{center}
In this supplementary material, we collect several proofs and remaining materials that were deferred from the main paper. In Appendix~\ref{subsection:Proof}, we provide the proofs of the main results in the paper. In Appendix~\ref{subsec:additional_study}, additional properties of distributional sliced-Wasserstein (DSW) distance are provided. In Appendix~\ref{sec:extension_DGSW}, we discuss distributional generalized sliced-Wasserstein distance (DGSW) and its dual form and properties. We describe in detail the applications of DSW and DGSW to generative modelings in Appendix~\ref{sec:application}. Furthermore, we provide additional experiments and experiment settings in Appendices~\ref{sec:addExp} and~\ref{sec:setting}.
\section{Proofs}
\label{subsection:Proof}
In this appendix, we collect the proofs for all the results in the main text.
\subsection{Proof of Theorem~\ref{theorem:DSW-distance}}
\label{sec:proof:theorem:DSW-distance}
We first show that the distributional sliced-Wasserstein distance satisfies the triangle inequality property for any three probability measures $\mu_{1}, \mu_{2}$, and $\mu_{3}$. In fact, from the definition of distributional sliced-Wasserstein distance for admissible $C>0$, for any $\epsilon > 0$ we find that 
\begin{align}
   \dislice_{p} (\mu_1,\mu_2;C) 
    & \stackrel{(i)} {\leq}  \Big{\{}\mathbb{E}_{\theta \sim \sigma^*_{\epsilon}}  W_p^p\big(\mathcal{R}I_{\mu_1}(\cdot,\theta),\mathcal{R}I_{\mu_2}(\cdot,\theta)\big)\Big{\}}^\frac{1}{p} + \epsilon \nonumber \\
    & \stackrel{(ii)} {\leq} \Big{\{}\mathbb{E}_{\theta \sim \sigma^*_{\epsilon}} \big{[} W_p (\mathcal{R}I_{\mu_1}(\cdot,\theta),\mathcal{R}I_{\mu_3}(\cdot,\theta)) + W_p^p (\mathcal{R}I_{\mu_3}(\cdot,\theta),\mathcal{R}I_{\mu_2}(\cdot,\theta)) \big{]} \Big{\}}^\frac{1}{p} + \epsilon \nonumber \\
    & \stackrel{(iii)} {\leq} \Big{\{}\mathbb{E}_{\theta \sim \sigma^*_{\epsilon}}   W_p^p (\mathcal{R}I_{\mu_1}(\cdot,\theta),\mathcal{R}I_{\mu_3}(\cdot,\theta))  \Big{\}}^\frac{1}{p} \nonumber \\& \quad+  \Big{\{}\mathbb{E}_{\theta \sim \sigma^*_{\epsilon}}  W_p^p\big(\mathcal{R}I_{\mu_3}(\cdot,\theta),\mathcal{R}I_{\mu_2}(\cdot,\theta)\big)  \Big{\}}^\frac{1}{p} + \epsilon \nonumber \\
    &\leq \sup_{\sigma \in \mathbb{M}_{C}} \Big{\{}\mathbb{E}_{\theta \sim \sigma} \big[ W_p^p\big(\mathcal{R}I_{\mu_1}(\cdot,\theta),\mathcal{R}I_{\mu_3}(\cdot,\theta)\big) \big] \Big{\}}^{\frac{1}{p}} \nonumber \\ 
     & \quad+ \sup_{\sigma \in \mathbb{M}_{C}} \Big{\{}\mathbb{E}_{\theta \sim \sigma} \big[  W_p^p \big(\mathcal{R}I_{\mu_3}(\cdot,\theta),\mathcal{R}I_{\mu_2}(\cdot,\theta)\big) \big]\Big{\}}^\frac{1}{p}  + \epsilon \nonumber \\
    &= \dislice_{p} (\mu_1, \mu_3;C) + \dislice_{p} (\mu_2, \mu_3;C) + \epsilon, \nonumber
\end{align}
where the existence of $\sigma_{\epsilon}^{*}$ in $(i)$ is from the definition of supremum; inequality in $(ii)$ is due to the triangle inequality with Wasserstein distance of order $p$; inequality in $(iii)$ follows from the application of the
Minkowski inequality. By letting $\epsilon \to 0$ in the above inequality, we obtain the conclusion with the triangle inequality of distributional sliced-Wasserstein distance.

The non-negativity and symmetry of distributional sliced-Wasserstein distance follow directly from the non-negativity and symmetry of Wasserstein distance. For the identity property, it is straight-forward that if $ \mu_{1} \equiv \mu_{2} $ then $ \dislice_p (\mu_{1},\mu_{2})=0 $. On the other hand, if $ \dislice_{p}(\mu_1,\mu_2) = 0 $, an application of Fourier transform as that in~\cite{bonnotte2013unidimensional} leads to $\mu_1 \equiv \mu_2$.

As a consequence, for any $p \geq 1$ and admissible $C > 0$, $\text{DSW}_p(.,.;C)$ is a well-defined metric in the space of Borel probability measures with finite $p$-th moment.
\subsection{Proof of Theorem~\ref{theorem:property_DSW}}
\label{sec:proof:theorem:property_DSW}
(a) From the definition of distributional sliced-Wasserstein distance, for any $p \geq 1$ and admissible $C > 0$ we find that
\begin{align*}
    \dislice_{p}(\mu,\nu; C) \leq \sup_{\mathcal{\sigma} \in \conspace} \biggr( \mathbb{E}_{\theta \sim \sigma} \biggr[ W_p^p(\mathcal{R}I_{\mu}(\cdot,\theta),\mathcal{R}I_{\nu}(\cdot,\theta)) \biggr] \biggr)^{\frac{1}{p}} = \text{maxSW}_{p}(\mu, \nu),
\end{align*}
where $\conspace$ is the space of all probability measures. The inequality is due to the fact that $\conspace_{C} \subseteq \conspace$ for all admissible $C > 0$. The second equality is true  because $W_p(\mathcal{R}I_{\mu}(\cdot,\theta),\mathcal{R}I_{\nu}(\cdot,\theta)) \leq \text{maxSW}_{p}(\mu, \nu)$ for all $\theta \in \sphere^{d-1}$, which leads to $\mathbb{E}_{\theta \sim \sigma} \bigr[ W_p^p(\mathcal{R}I_{\mu}(\cdot,\theta),\mathcal{R}I_{\nu}(\cdot,\theta))\bigr] \leq \text{maxSW}_{p}^p(\mu, \nu)$. The  inequality becomes equality when $\sigma$ is the Dirac measure at $\theta^{*}$ that maximizes the value of $W_p(\mathcal{R}I_{\mu}(\cdot,\theta),\mathcal{R}I_{\nu}(\cdot,\theta))$. 

Furthermore, we have
\begin{align*}
    W_p^{p}(\mathcal{R}I_{\mu}(\cdot,\theta),\mathcal{R}I_{\nu}(\cdot,\theta)) & = \inf_{\pi \in \Pi(\mu, \nu)} \int_{\mathcal{X} \times \mathcal{Y}} | x^{\top} \theta - y^{\top} \theta |^{p} d \pi(x,y) \\
    & \leq \inf_{\pi \in \Pi(\mu, \nu)} \int_{\mathcal{X} \times \mathcal{Y}} \| x - y \|^{p} d \pi(x,y) = W_{p}^{p}(\mu, \nu),
\end{align*}
where the last inequality is due to the fact that length of side of right triangle $|(x - y)^{\top} \theta|$ is less than length of its hypotenuse  $ \| x - y \|$ for all $\theta \in \sphere^{d-1}$. Therefore, $\text{maxSW}_{p}(\mu, \nu) \leq W_{p}(\mu, \nu)$ for any $p \geq 1$.

Putting the above results together, we obtain the conclusion of part (a) of the theorem. \\

\noindent
(b) Denote $\bar{\sigma} = \sum_{i = 1}^{d} \frac{1}{d} \delta_{\theta_{i}}$ where $\theta_{1} = \theta^{*}$, which maximizes the value of $W_p(\mathcal{R}I_{\mu}(\cdot,\theta),\mathcal{R}I_{\nu}(\cdot,\theta))$ and $\theta_{1}, \ldots, \theta_{d}$ form an orthonormal basis in $\mathbb{R}^{d}$. Simple algebra shows that
\begin{align*}
    \mathbb{E}_{\theta,\theta' \sim \mathcal{\bar{\sigma}}}\left[|\theta^\top \theta'|\right] = \sum_{1 \leq i,j \leq d} \bigr(\frac{1}{d}\bigr)^{2} |\theta_{i}^{\top}\theta_{j}| = \frac{1}{d}.
\end{align*}
Since $C \geq \frac{1}{d}$, the above equation indicates that $\bar{\sigma} \in \conspace_{C}$. Therefore, we find that
\begin{align*}
    \dislice_{p}(\mu,\nu; C) & \geq \biggr( \mathbb{E}_{\theta \sim \bar{\sigma}} \biggr[ W_p^p(\mathcal{R}I_{\mu}(\cdot,\theta),\mathcal{R}I_{\nu}(\cdot,\theta)) \biggr] \biggr)^{\frac{1}{p}} \\
    & = \biggr(\sum_{i = 1}^{d} \frac{1}{d} W_p^p(\mathcal{R}I_{\mu}(\cdot,\theta_{i}),\mathcal{R}I_{\nu}(\cdot,\theta_{i})) \biggr)^{\frac{1}{p}} \\
    & \geq \bigr(\frac{1}{d}\bigr)^{\frac{1}{p}} W_p(\mathcal{R}I_{\mu}(\cdot,\theta_{1}),\mathcal{R}I_{\nu}(\cdot,\theta_{1})) = \bigr(\frac{1}{d}\bigr)^{\frac{1}{p}} \text{maxSW}_{p}(\mu, \nu).
\end{align*}
Moreover, for any $p \geq 1$, $\text{SW}_{p}(\mu, \nu) \leq \text{maxSW}_{p}(\mu, \nu)$. Collecting the previous results, we reach the conclusion of part (b). 
\paragraph{Equivalence of $\text{DSW}_{p}(\cdot,\cdot;C)$ to other distances:} Based on the result of Theorem 2.1 in~\cite{bayraktar2019strong}, $\text{maxSW}_{p}$, $\text{SW}_{p}$, and $W_{p}$ are equivalent distances for any $p \geq 1$. In particular, for any sequence $(\mu_{n})_{n \geq 1} \in \mathcal{P}_p (\mathbb{R}^{d})$ and $\mu \in \mathcal{P}_p (\mathbb{R}^{d})$, the following holds
\begin{align}
    \lim_{n \to \infty} \text{maxSW}_{p}(\mu_{n}, \mu) = 0 \iff \lim_{n \to \infty} \text{SW}_{p}(\mu_{n}, \mu) = 0 \iff \lim_{n \to \infty} W_{p}(\mu_{n}, \mu) = 0. \label{eq:equiv_one}
\end{align}
Now, if we have $\lim_{n \to \infty} \text{maxSW}_{p}(\mu_{n}, \mu) = 0$ for $p \geq 1$, the result of part (a) shows that $\lim_{n \to \infty} \text{DSW}_{p}(\mu_{n}, \mu; C) = 0$. On the other hand, when $\lim_{n \to \infty} \text{DSW}_{p}(\mu_{n}, \mu; C) = 0$, as long as $C \geq \frac{1}{d}$ and $p \geq 1$, the result of part (b) leads to $\lim_{n \to \infty} \text{maxSW}_{p}(\mu_{n}, \mu) = 0$. As a consequence, when $C \geq \frac{1}{d}$ and $p \geq 1$ we have
\begin{align}
    \lim_{n \to \infty} \text{DSW}_{p}(\mu_{n}, \mu; C) = 0 \iff \lim_{n \to \infty} \text{maxSW}_{p}(\mu_{n}, \mu) = 0. \label{eq:equiv_two}
\end{align}
Combining the results in equations~(\ref{eq:equiv_one}) and~(\ref{eq:equiv_two}), we reach the conclusion that when $C \geq \frac{1}{d}$ and $p \geq 1$, $\text{DSW}_{p}(\cdot,\cdot;C)$, $\text{maxSW}_{p}$, $\text{SW}_{p}$, and $W_{p}$ are equivalent distances.  

\section{Additional Studies with Distributional Sliced-Wasserstein Distance}
\label{subsec:additional_study}
In this appendix, we provide further studies with distributional sliced-Wasserstein distance.
\subsection{Discussion of the constraint in DSW}
\label{subsec:constraint_DSW}
We first compute $\mathbb{E}_{\theta,\theta' \sim \sigma^{d-1}}\left[|\theta^\top \theta'|\right]$ where $\sigma^{d-1}$ is the uniform distribution on the unit sphere $\mathbb{S}^{d - 1}$.
\begin{theorem} 
\label{theorem:uniform}
For uniform measure $\sigma^{d-1}$ on the unit sphere $\sphere^{d-1}$, we have
\begin{align*}
    \int_{\theta,\theta^{\prime}\sim \sigma^{d-1}} |\theta^{\top}\theta^{\prime} | d\sigma^{d-1}(\theta) d\sigma^{d-1}(\theta^{\prime}) = \frac{\Gamma(\frac{d}{2})}{\pi^{\frac{1}{2}} \Gamma(\frac{d+1}{2})},
\end{align*}
where $\Gamma(.)$ is the Gamma function.
\end{theorem}
\begin{remark}
The result of Theorem~\ref{theorem:uniform} indicates that as long as $C \geq \frac{\Gamma(\frac{d}{2})}{\pi^{\frac{1}{2}} \Gamma(\frac{d+1}{2})}$, we have $\sigma^{d- 1} \in \mathbb{M}_{C}$. Furthermore, by Gautschi's inequality for the Gamma function, we find that
\begin{align*}
    \frac{1}{\pi^{\frac{1}{2}} (\frac{d+1}{2})^{\frac{1}{2}}} < \frac{\Gamma(\frac{d}{2})}{\pi^{\frac{1}{2}} \Gamma(\frac{d+1}{2})} < \frac{1}{\pi^{\frac{1}{2}} (\frac{d-1}{2})^{\frac{1}{2}}}
\end{align*}
For $d\geq 3$, we have $2d^2/(d+1) > \pi$. Hence, we obtain that 
\begin{align*}
    \frac{1}{\pi^{\frac{1}{2}} (\frac{d+1}{2})^{\frac{1}{2}}} > \frac{1}{d}.
\end{align*}
Given the above bound, when $C \geq \frac{\Gamma(\frac{d}{2})}{\pi^{\frac{1}{2}} \Gamma(\frac{d+1}{2})}$, the set $\mathbb{M}_{C}$ contains both $\sigma^{d - 1}$ and $\bar{\sigma} = \sum_{i = 1}^{d} \frac{1}{d} \delta_{\theta_{i}}$ where $\theta_{1}, \ldots, \theta_{d}$ form any orthonormal basis in $\mathbb{R}^{d}$. Furthermore, for $d =2$, we have
\begin{align*}
    \frac{\Gamma(1)}{\pi^{\frac{1}{2}} \Gamma(\frac{2+1}{2})} = \frac{2}{\pi} > \frac{1}{2}.
\end{align*}
Therefore, when $d = 2$ and $C \geq \frac{\Gamma(\frac{d}{2})}{\pi^{\frac{1}{2}} \Gamma(\frac{d+1}{2})}$, the set $\mathbb{M}_{C}$ also contains $\bar{\sigma} = \sum_{i = 1}^{d} \frac{1}{d} \delta_{\theta_{i}}$.
\end{remark}
\begin{proof}
Since $\sigma^{d-1}$ is the uniform measure on the unit sphere $\sphere^{d-1}$,  the integral
\begin{align*}
    \int_{\theta\sim \sigma^{d-1}} |\theta^{\top} \theta^{\prime}|d\sigma^{d-1}(\theta)
\end{align*}
is the same for all fixed $\theta^{\prime}$. Hence for any fixed $\theta^*\in \sphere^{d-1}$, we obtain
\begin{align*}
    I = \int_{\theta,\theta^{\prime}\sim \sigma^{d-1}} |\theta^{\top}\theta^{\prime} | d\sigma^{d-1}(\theta) d\sigma^{d-1}(\theta^{\prime}) =\int_{\theta\sim \sigma^{d-1}} |\theta^{\top} \theta^{*}|d\sigma^{d-1}(\theta).
\end{align*}
Without loss of generality, we choose $\theta^* = (1,0,\ldots,0)$, $I$ is equal to  
\begin{align*}
\int_{\theta\sim \sigma^{d-1}} |\theta^{(1)}|d\sigma^{d-1}(\theta),
\end{align*}
where $\theta = (\theta^{(1)},\ldots,\theta^{(d)})$. For any measurable subset $S$ of $\sphere^{d-1}$, let $A(S)$ be the area of $S$ on the surface of $\sphere^{d-1}$ and $A(\sphere^{d-1})$ be the area of the surface of $\sphere^{d-1}$ which is equal to
\begin{align*}
    A(\sphere^{d-1}) = \frac{d\pi^{\frac{d}{2}}}{\Gamma(\frac{d}{2}+1)}. 
\end{align*}
Now, we have 
\begin{align*}
  \int_{\theta\sim \sigma^{d-1}} |\theta^{(1)}|d\sigma^{d-1}(\theta)
   = \frac{1}{A(\sphere^{d-1})} \int_{\theta \in \sphere^{d-1}} |\theta^{(1)}| d A(\sphere^{d-1}(\theta)).
\end{align*}
Let $H_1$ be the hyperplane formed by $\theta^{(2)},\ldots,\theta^{(d)}$ and $H_{\theta}$ be the hyperplane tangent to the sphere $\sphere^{d-1}$ at $\theta$. Then $\theta$ is the normal vector to $H_{\theta}$ and $\theta^* = (1,0,\ldots,0)$ is orthogonal to $H_1$. Let $\alpha$ be the angle between $\theta$ and $\theta^*$. Then
\begin{align*}
    dA(\sphere^{d-1}(\theta)) \cos(H_1,H_{\theta}) &= d\theta^{(2)} \ldots d\theta^{(d)} \\
    dA(\sphere^{d-1}(\theta)) &=\frac{1}{\cos(\theta,\theta^*)} d\theta^{(2)}\ldots d\theta^{(d)}  = \frac{1}{|\theta^{(1)}|}d\theta^{(2)}\ldots d\theta^{(d)}.
\end{align*}
Return to the integral, we find that
\begin{align*}
I = \int_{\theta\sim \sigma^{d-1}} |\theta^{(1)}| d\sigma^{d-1}(\theta) &=  \frac{1}{A(\sigma^{d-1})} \int_{\sum_{i=1}^d (\theta^i)^2 = 1}|\theta^{(1)}| \frac{1}{|\theta^{(1)}|}d\theta^{(2)}\ldots d\theta^{(d)} \\
&= \frac{2}{A(\sigma^{d-1})}\int_{\theta^{(1)} > 0, \sum_{i=2}^d (\theta^{(i)})^2 \leq 1}  d\theta^{(2)}\ldots d\theta^{(d)} \\
&= \frac{2}{A(\sphere^{d-1})} V(\mathbb{B}^{d-1})  \\
&= \frac{2 \Gamma(\frac{d}{2})}{2\pi^{\frac{d}{2}}} \times \frac{\pi^{\frac{d-1}{2}}}{\Gamma(\frac{d-1}{2}+1)} \\
&= \frac{\Gamma(\frac{d}{2})}{\pi^{\frac{1}{2}} \Gamma(\frac{d+1}{2})},
\end{align*}
where $\mathbb{B}^{d-1}$ is the unit ball in the $d-1$ dimensional space and $V(\mathbb{B}^{d-1})$ is its corresponding volume. As a consequence, we obtain the conclusion of the theorem.
\end{proof}
   
\subsection{Approximation of dual value of DSW}
\label{subsec:gradient_DSW}

Now, we give a detailed form of the objective function $\text{DS}(f_{\phi})$ in the dual form of DSW in equation~(\ref{eq:dual_dsw_equiv}). In particular, simple calculation shows that
\begin{align}
    \nabla_\phi \text{DS}(f_{\phi}) 
    &= \frac{1}{p} \Big{\{} \mathbb{E}_{\theta \sim \sigma^{d-1}}\big[W_p^p (\mathcal{R}I_\mu (.,f_\phi(\theta)), \mathcal{R}I_\nu (.,f_\phi(\theta))\big]  \Big{\}}^{\frac{1}{p}-1} \label{eq:gradient_compute} \\ & \hspace{-0.5 em} \times \mathbb{E}_{ \theta \sim \sigma^{d-1}}\big[  \nabla_\phi W_p^p (\mathcal{R}I_\mu (.,f_\phi(\theta)), \mathcal{R}I_\nu (.,f_\phi(\theta))\big] - \lambda_C \mathbb{E}_{\theta,\theta' \sim \sigma^{d-1}} \big[\nabla_\phi |f_{\phi}(\theta)^\top f_{\phi}(\theta')|\big]. \nonumber
\end{align}
Since the outer expectations in equation~(\ref{eq:gradient_compute}) are intractable to compute, we employ the standard Monte Carlo scheme to approximate these expectations. Therefore, we obtain the following approximation:
\begin{align*}
    &\nabla_\phi \text{DS}(f_{\phi})  \approx  \frac{1}{p} \Big{\{} \frac{1}{n} \sum_{i=1}^n \big[W_p^p (\mathcal{R}I_\mu (.,f_\phi(\theta_i)), \mathcal{R}I_\nu (.,f_\phi(\theta_i))\big]  \Big{\}}^{\frac{1}{p}-1} \\ 
    & \hspace{- 1 em} \times \Bigg\{\frac{1}{n}\sum_{i=1}^n \Big[ \nabla_\phi W_p^p (\mathcal{R}I_\mu (.,f_\phi(\theta_i)), \mathcal{R}I_\nu (.,f_\phi(\theta_i)) \Big]\Bigg\} - \frac{\lambda_{C}}{n (n - 1)} \sum_{1 \leq i \neq j \leq n} \nabla_\phi \big|(f_\phi(\theta_{i}))^{\top} f_\phi(\theta_{j})\big|,  \nonumber
\end{align*}
where $\theta_{1}, \ldots, \theta_{n}$ are i.i.d. samples from the unit sphere $\mathbb{S}^{d - 1}$.

Denote $\phi^{*}$ as the fixed point of the stochastic gradient ascent algorithm. Then, we can use $f_{\phi^{*}}$ as the local maxima of the optimization problem~(\ref{eq:dual_dsw_equiv}). By using Monte Carlo method to approximate the expectation in equation~(\ref{eq:dual_dsw_equiv}), we obtain the following approximation:
\begin{align*}
    \dislice_{p}^{*}(\mu,\nu; C) \approx \Big\{ \frac{1}{n} \sum_{i = 1}^{n} \big[W_p^p\big(\mathcal{R}I_{\mu}(\cdot, f_{\phi^{*}}(\theta_{i}),\mathcal{R}I_{\nu}(\cdot, f_{\phi^{*}}(\theta_{i}))\big)  \big] \Big\}^{1/ p} \nonumber \\ &
    & \hspace{-16 em} - \frac{\lambda_{C}}{n(n-1)} \sum_{1 \leq i \neq j \leq n} \big|(f_{\phi^{*}}(\theta_{i}))^{\top} f_{\phi^{*}}(\theta_{j}) \big| + \lambda_{C} C.
\end{align*}
\subsection{Statistical guarantee of DSW}
\label{subsec:stats_DSW}
In this appendix, we provide the statistical guarantee of DSW. 
\begin{theorem}
\label{theorem:stats_DSW}
Given probability measure $P$ supported on a compact subset $\Theta \subset \mathbb{R}^{d}$. Assume that $X_{1}, \ldots, X_{n}$ are i.i.d. data from $P$. Denote $P_{n} = \frac{1}{n} \sum_{i = 1}^{n} \delta_{X_{i}}$ the empirical measure of the data points $X_{1}, \ldots, X_{n}$. Then, for any admissible regularizing constant $C > 0$ and for any $p \geq 1$, we obtain that
\begin{align*}
     \mathbb{E} \biggr[\dislice_{p}(P_{n},P; C)\biggr] 
\leq c \sqrt{\frac{d \log n}{n}},
\end{align*}
where $c > 0$ is some universal constant.
\end{theorem}
\begin{remark} The result of Theorem~\ref{theorem:stats_DSW} demonstrates that DSW has similar statistical guarantees as other sliced distances and does not suffer from the curse of dimensionality. Therefore, it is an appealing distance for applications in generative modeling.
\end{remark}
\begin{proof}
The proof of Theorem~\ref{theorem:stats_DSW} is a direct application of Theorem~\ref{theorem:property_DSW} and statistical guarantee of max-sliced Wassertein distance. Here, we provide the proof for the completeness. In particular, based on the result of Theorem~\ref{theorem:property_DSW}, we obtain that
\begin{align*}
    \mathbb{E} \biggr[\dislice_{p}(P_{n},P; C)\biggr] \leq \mathbb{E} \biggr[ \text{maxSW}_{p}(P_{n}, P)\biggr].
\end{align*}
Therefore, it is sufficient to demonstrate that $\mathbb{E} \biggr[ \text{maxSW}_{p}(P_{n}, P)\biggr] \leq c \sqrt{\frac{d \log n}{n}}$ for some universal constant $c > 0$. In order to simplify the presentation, we denote a few notation. First, we define $\mathcal{H}$ the set
of half-spaces $H_{\theta,x} = \{y\in \mathbb{R}^d: \langle y, \theta \rangle \leq x\}$ for any $\theta \in \mathbb{S}^{d-1}$ and $x \in \mathbb{R}$. Then, it has been shown that $\mathcal{H}$ has at most $d+1$ Vapnik–Chervonenkis (VC) dimension~\cite{wainwrighthigh}. The VC inequality implies that
\begin{align*}
\sup_{H \in \mathcal{H}} |P_n(H) - P(H)| \leq 
     \sqrt{\frac{32}{n}\big[(d+1)\log(n+1) + \log(8/\delta)\big]}=: c_{n,\delta}
\end{align*}
with probability at least $1-\delta$, for any $\delta \in(0,1)$. On the other hand, we have
\begin{align*}
\sup_{H \in \mathcal{H}} |P_n(H) - P(H)| = 
    \sup_{\substack{x \in \mathbb{\mathbb{R}} , \theta \in \mathbb{S}^{d-1}}} |F_{n,\theta}(x)
     - F_\theta(x)|,
\end{align*}
where $F_{n,\theta}$ and $F_\theta$ are respectively the cumulative distribution functions (CDF) of 
$\mathcal{R}I_{P_{n}}(\cdot,\theta)$ and $\mathcal{R}I_{P}(\cdot,\theta)$. Given the above equation and the close-form of Wasserstein distance in one dimension, we find that
\begin{align*}
    \text{maxSW}_{p}^{p}(P_{n}, P) & = \max_{\theta \in \mathbb{S}^{d - 1}} \int_{0}^{1} |F_{n,\theta}^{-1} (u)
     - F_\theta^{-1} (u)|^{p} du \\
     & = \max_{\theta \in \mathbb{S}^{d - 1}} \int_{\mathbb{R}} |F_{n,\theta} (x)
     - F_\theta (x)|^{p} dx \\
     & \leq \text{diam}(\Theta) \sup_{\substack{x \in \mathbb{\mathbb{R}} , \theta \in \mathbb{S}^{d-1}}} |F_{n,\theta}(x)
     - F_\theta(x)|^{p} \leq \text{diam}(\Theta) c_{n,\delta}^{p}.
\end{align*}
By using the above inequality, we obtain that $\mathbb{E} \biggr[ \text{maxSW}_{p}(P_{n}, P)\biggr] \leq c \sqrt{\frac{d \log n}{n}}$ for some universal constant $c > 0$. As a consequence, we reach the conclusion of Theorem~\ref{theorem:stats_DSW}.
\end{proof}
\section{An extension to distributional generalized sliced-Wasserstein distance}
\label{sec:extension_DGSW}
We now consider an extension of DSW to non-linear projections via generalized Radon transform. The constant $C > 0$ is \emph{generalized admissible} if the set  $\overline{\conspace}_{C}$ of probability measures $\mathcal{\sigma}$ on  the compact set of feasible parameters $\Omega_{\theta}$ satisfying $\mathbb{E}_{\theta,\theta' \sim \mathcal{\sigma}}\left[|\cos(\theta, \theta')|\right] \leq  C$ is not empty.
\begin{definition}
Given two probability measures $\mu$ and $\nu$ on $\mathbb{R}^d$ with finite $p$-th moments where $p \geq 1$ and a generalized admissible regularizing constant $C>0$. The \emph{distributional generalized sliced-Wasserstein distance} (DGSW) of order $p$ between $\mu$ and $\nu$ is defined as follows:
\begin{align}
   & \digslice_{p}(\mu,\nu; C) \nonumber := \sup_{\sigma \in \overline{\conspace}_{C}} \Big{\{}\mathbb{E}_{\theta \sim \sigma} W_p^p(\mathcal{G}I_{\mu}(\cdot,\theta),\mathcal{G}I_{\nu}(\cdot,\theta)) \Big{\}}^{1/ p}, 
\end{align}
where $\mathcal{G}$ is generalized Radon transform defined in Section~\ref{sec:random_transform}.
\end{definition}
The DGSW distance uses the advantage of non-linear projections to capture more complex structures of the target probability measures. We show that as long as the generalized Radon transform is injective, DGSW is a proper metric in the probability space.
\begin{theorem}
\label{theorem:generalized_sliced}
For any $p \geq 1$ and generalized admissible $C>0$, as long as the generalized Radon transform is injective, the distributional generalized sliced-Wasserstein is a well-defined metric in the space of Borel probability measures with finite $p$-th moment.
\end{theorem}
The proof of Theorem~\ref{theorem:generalized_sliced} simply follows the proof argument of Theorem~\ref{theorem:DSW-distance} under the injectivity of GRT; thus, it is omitted. In order to compute DGSW, we also utilize the dual form of DGSW as that of DSW distance.
\paragraph{Dual form of distributional generalized sliced-Wasserstein distance:} Similar to the distributional sliced-Wasserstein distance, we use the dual form of distributional generalized sliced-Wasserstein distance to approximate the value of distributional generalized sliced-Wasserstein distance. Recall that, for any $\theta,  \theta' \in \mathbb{R}^{d}$, $\text{cos}(\theta,\theta') = \frac{\theta^{\top} \theta'}{\|\theta \| \| \theta'\|}$.
\begin{definition}
\label{def:dual_DGSW}
For any $p \geq 1$ and generalized admissible $C > 0$, there exists a non-negative constant $\lambda_{C}$ depending on $C$ such that the dual form of DGSW distance takes the following form
\begin{align*}
    \digslice_{p}^{*}(\mu,\nu; C) & : = - \sup_{\lambda \geq 0} \inf_{\mathcal{\sigma} \in \bar{\conspace}} \Biggr\{ - \biggr( \mathbb{E}_{\theta \sim \sigma} \biggr[ W_p^p(\mathcal{G}I_{\mu}(\cdot,\theta),\mathcal{G}I_{\nu}(\cdot,\theta)) \biggr] \biggr)^{1/p} \\
    & \hspace{16 em} + \lambda \left( \mathbb{E}_{\theta,\theta' \sim \sigma}\biggr[\frac{|\theta^\top \theta'|}{\norm{\theta} \norm{\theta'}}\biggr] - C \right) \Biggr\} \\ 
    & = \sup_{\mathcal{\sigma} \in \bar{\conspace}} \Bigg\{ \biggr( \mathbb{E}_{\theta \sim \sigma} \biggr[ W_p^p(\mathcal{G}I_{\mu}(\cdot,\theta),\mathcal{G}I_{\nu}(\cdot,\theta)) \biggr] \biggr)^{1/p} - \lambda_{C} \mathbb{E}_{\theta,\theta' \sim \sigma}\biggr[\frac{|\theta^\top \theta'|}{\norm{\theta} \norm{\theta'}}\biggr] \Biggr\} \\
    & \hspace{16 em} + \lambda_{C} C, \nonumber
\end{align*}
where $\bar{\conspace}$ denotes the space of all probability measures on the compact set of feasible parameter $\Omega_\theta$.
\end{definition}
From the duality theory, we obtain that $\digslice_{p}(\mu,\nu; C) \geq \digslice_{p}^{*}(\mu,\nu; C)$ for any $p \geq 1$ and admissible $C > 0$. Similar to DSW distance, the dual form of DGSW provides an efficient way to approximate the DGSW distance. We show that when the compact set of feasible parameter $\Omega_{\theta} = \sphere^{d - 1}$, similar reparametrization trick like that of the dual form of DSW distance can be applied to the dual form of DGSW distance. In particular, when $\Omega_{\theta} = \sphere^{d - 1}$, we obtain the equivalent dual form of DGSW as follows:
\begin{align}
    \digslice_{p}^{*}(\mu,\nu; C) = \sup_{f\in \mathcal{F}}\Bigg\{ \biggr( \mathbb{E}_{\theta \sim \sigma^{d-1}} \big[W_p^p\big(\mathcal{G}I_{\mu}(\cdot, f(\theta)), \mathcal{G}I_{\nu}(\cdot, f(\theta))\big)  \big] \biggr)^{1/ p}  & \label{eq:dual_dgsw_equiv} \\
 & \hspace{-14 em} - \lambda_{C} \mathbb{E}_{\theta,\theta^{'} \sim \sigma^{d-1}}\Big[\big|f(\theta)^{\top} f(\theta^{'}) \big|\Big] \Bigg\} + \lambda_{C} C, \nonumber
\end{align}
where $\mathcal{F}$ is a class of Borel measurable functions from $\sphere^{d-1}$ to $\sphere^{d-1}$ and $\lambda_{C} > 0$ is some positive constant given in Definition~\ref{def:dual_DGSW}. Then, in order to find an optimal $f$, we can parameterize $f$ as $f_{\phi}$, which we can think as (deep) neural network. From here, with similar argument as that of equation~(\ref{eq:gradient_compute}), we can approximate the gradient of the objective function in equation~(\ref{eq:dual_dgsw_equiv}) with respect to $\phi$ and then use stochastic gradient ascent algorithm to update $\phi$. Finally, we can use the fixed point of the algorithm to approximate the dual value of DGSW in equation~(\ref{eq:dual_dgsw_equiv}). 

\section{Applications to Generative Modeling}
\label{sec:application}
The DSW and DGSW distances can potentially be applied in settings where there is a benefit of employing an optimal-transport type of distance in a computationally efficient manner. In this section, we discuss two general settings where the DSW and DGSW distances can be immediately applied. The first setting is a standard generative modeling task using the minimum expected distance estimator framework \cite{bernton2019parameter} where a generative model is fitted to a data distribution by minimizing an appropriate divergence. The second setting is a joint contrastive inference task where both a generative model and inference model are learned jointly, again by minimizing some divergence in the joint space of observed variable and latent variable. In each setting, we apply the DSW to these tasks as well as its generalized version, the DGSW.

\subsection{Minimum expected distributional sliced-Wasserstein estimator}
\label{subsec:MEDE}
Minimum expected distance estimators \cite{bernton2019parameter} are widely used recently due to its efficiency in learning implicit generative models. Popular estimators include those based on OT distances~\cite{ arjovsky2017wasserstein, genevay2018learning,tolstikhin2018wasserstein} due to their smooth and differentiable objectives especially when the supports of the data and the generative distributions are not the same. In sliced-Wasserstein cases, SW and Max-SW have been employed with rigorous theoretical analyses in various works~\cite{bayraktar2019strong,deshpande2019max,deshpande2018generative,nadjahi2019asymptotic}. They enjoy the benefits of
the Wasserstein distance in one dimension and obtain fast speed in training the
model. In this paper, we introduce a new novel estimator by replacing SW and Max-SW by our new DSW distance, which we refer to as \emph{minimum expected distributional sliced-Wasserstein estimator}. The new estimator is defined as follows: 
\begin{align}
    \hat{\theta}_{n} = \argmin{}_{\theta \in \Theta } \mathbb{E}[\dislice_{p}(\hat{\mu}_n,\hat{\mu}_{\theta,m} )|X_{1:n}],
\end{align}
where $\Phi$ is the parameter space, $\hat{\mu}_n = \frac{1}{n} \sum_{i=1}^n \delta_{X_i}$ is the empirical measure, and $\hat{\mu}_{\theta,m}= \frac{1}{m} \sum_{i=1}^m \delta_{Y_i}$ denotes the empirical distribution that is obtained by sampling i.i.d samples from $\mu_\theta$. In practice, $\mu_\theta$ is created by pushing a simple distribution $\epsilon$ (such as the standard Gaussian)  through a neural net, parameterized by $\theta$, i.e., $\mu_\theta = T_\theta \sharp \epsilon$.  

\subsection{Distributional sliced-Wasserstein joint contrastive inference}

Learning both a generator and an inference model, i.e., an encoder, is a central task in latent-variable modeling.  A general framework for performing this task is called joint contrastive inference~\cite{dumoulin2016adversarially}. Let $p_\theta(z,x)= p(z)p_\theta(x|z)$ be a generative model, $q_\phi(z|x)$ be an amortized inference model and define the data-induced aggregated joint inference model as $\hat{q}_\phi(z,x)=p_{data}(x)q_\phi(z|x)$. The joint contrastive inference framework then minimizes some divergence between the two structured joint distributions $p_\theta(z,x)$ and $\hat{q}_\phi(z,x)$. This can be seen as a generalized version of amortized inference. There are some well-known examples of this kind of inference such as the Variational Autoencoder \cite{kingma2013auto}, Adversarially Learned Inference \cite{dumoulin2016adversarially}, and Wasserstein Variational Inference \cite{ambrogioni2018wasserstein}. By using the DSW distance, we obtain a new joint contrastive inference method which inherits the benefits of optimal transport family of distances, yet remains scalable and computationally efficient. In particular, we learn both a generator and an inference model by solving:
\begin{equation}
    (\theta_m, \phi_m) = \argmin{}_{\theta \in \Theta, \phi \in \Phi} \mathbb{E}_{\hat{q}_\phi (z,x), p_\theta (z,x)} [ \dislice_{p}( \hat{q}_{\phi,m} (x,z),\hat{p}_{\theta,m} (z,x))],
\end{equation}
where $\Theta, \Phi$ are the parameter spaces, $\hat{q}_{\phi,m} (z,x)$ and $\hat{p}_{\theta,m} (z,x)$ are empirical distributions that sampled i.i.d data from $\hat{q}_\phi (z,x)$ and $ p_\theta (z,x) $ respectively.
\section{Additional Experiments}
\label{sec:addExp}
In this appendix, we provide additional experimental results to yield more understandings about the minimum expected distance framework, which uses the new proposed distances. The appendix is divided into three parts, namely Appendices E.1, E.2 and E.3.  Appendix E.1 is devoted to showing the  performances of DGSW (see Appendix~\ref{sec:extension_DGSW} for its definition) versus the generalized versions of other sliced distances on various factors which could affect the effectiveness of those methods. We also compare DSW to the recent augumented sliced Wasserstein method  (ASW)~\cite{chen2020augmented}. Then we show the generated images from slice-based distances method for MNIST, CelebA and LSUN, when the number of projections varies.   In Appendix E.2, we compare DSW to the projected robust subspcaae Wasserstein (PRW) in~\cite{paty2019subspace,lin2020projection} on MNIST dataset . The comparison is to show  Wasserstein-2 distance between the learned distribution and the data distribution versus the execution time. Finally, Appendix E.3 includes a comparison between DSW, DGSW, SW, Max-SW, Max-GSW, and Max-GSW-NN for the joint contrastive inference task on MNIST dataset. 

\subsection{Generative models}
\label{subsec:exp_generative_models}
\begin{figure}[!ht]
\begin{center}

  \begin{tabular}{cc}
    \widgraph{0.3\textwidth}{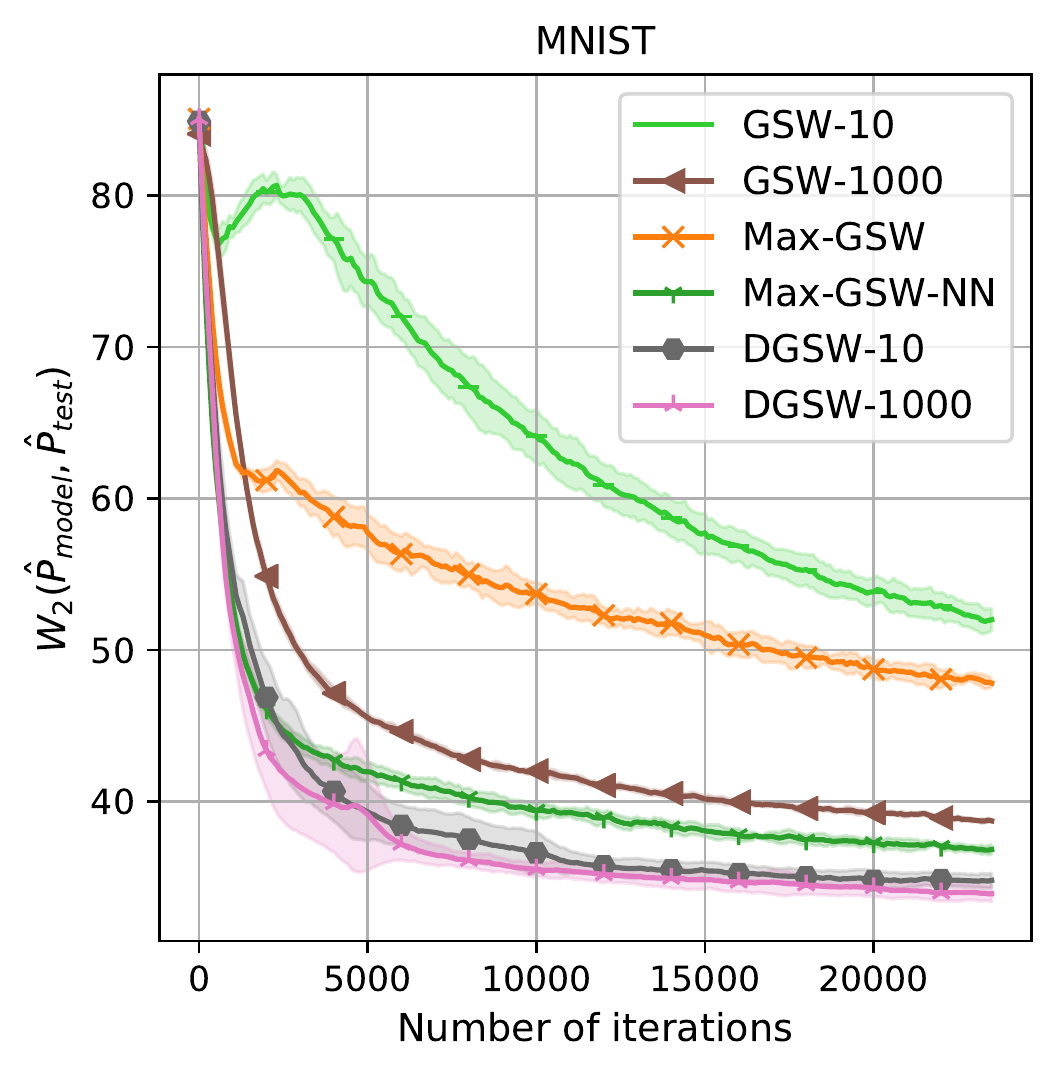}&
   \widgraph{0.3\textwidth}{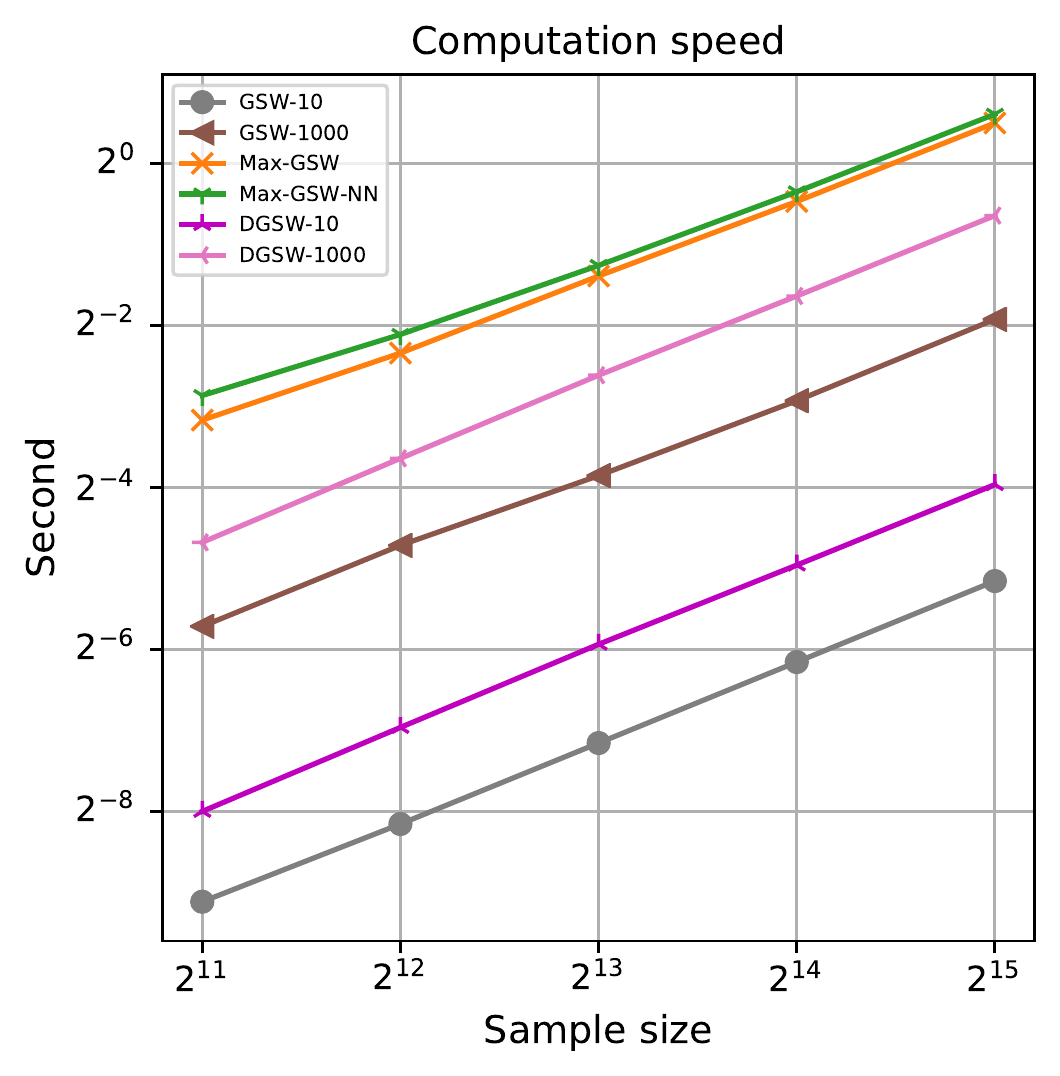} 
    \\
    (a) &(b) 
  \end{tabular}
  \end{center}
  \caption{
  \footnotesize{(a) Comparison between DGSW, GSW, Max-GSW and Max-GSW-NN using $W_2$ distance as metric. Here, GSW, Max-GSW and DGSW use circular function. (b) The computional speed over size of samples.
    }}
  \label{fig:gswMNISTgraphs}
\end{figure}

\textbf{DGSW results on MNIST: } 
Figure~\ref{fig:gswMNISTgraphs}(a) shows the convergence of estimators of the learned distribution to the data distribution based on ``generalized" sliced distances in the sense of  Wasserstein-2 distance. Here, we use the circular function as the defining function for both GSW, Max-GSW, and DGSW (the polynomial function is very expensive in high-dimension). With 10 projections, DGSW produces better performance than  GSW with 1000 projections, Max-GSW and Max-GSW-NN. There is a little improvement in the Wasserstein-2 score with DGSW when we increase the number of projections from 10 to 1000.   For the computational speed shown in Figure~\ref{fig:gswMNISTgraphs}(b), DGSW-10 is much faster than other reported methods, except the GSW-10 which has the worst Wasserstein-2 score.

\begin{figure}[!ht]
\begin{center}

  \begin{tabular}{cccc}
    \widgraph{0.21\textwidth}{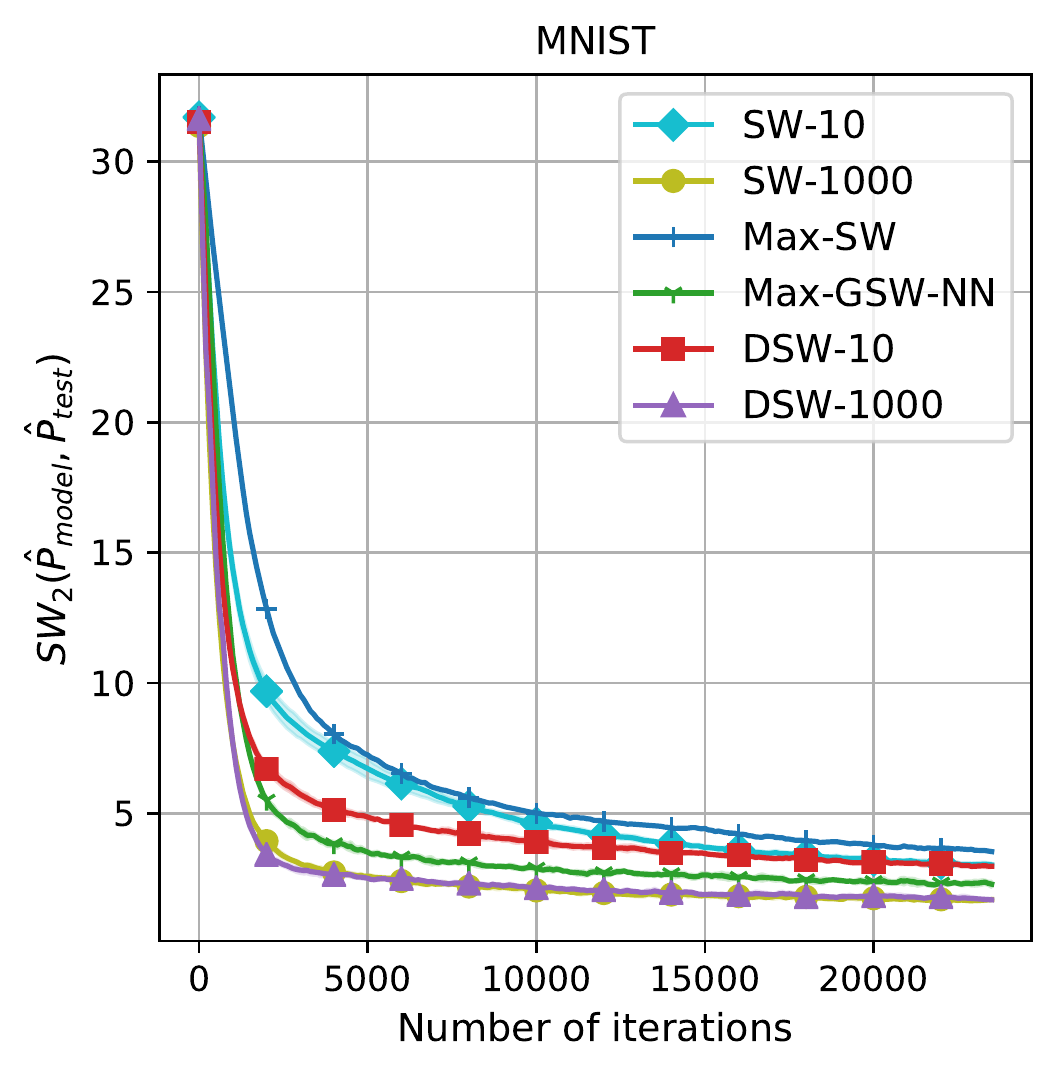}
    \label{fig:sub5}&
   \widgraph{0.21\textwidth}{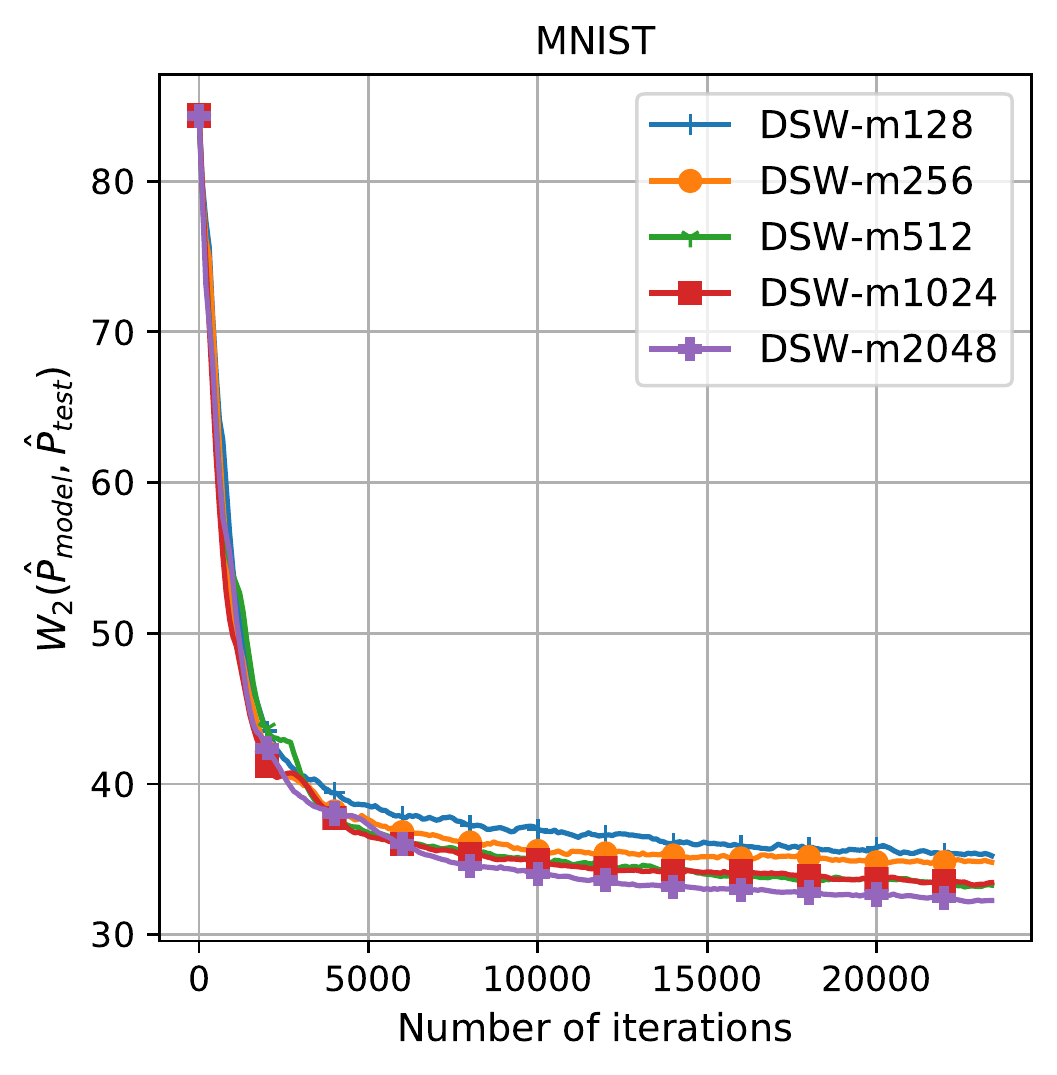}
    \label{fig:sub6}&
    \widgraph{0.21\textwidth}{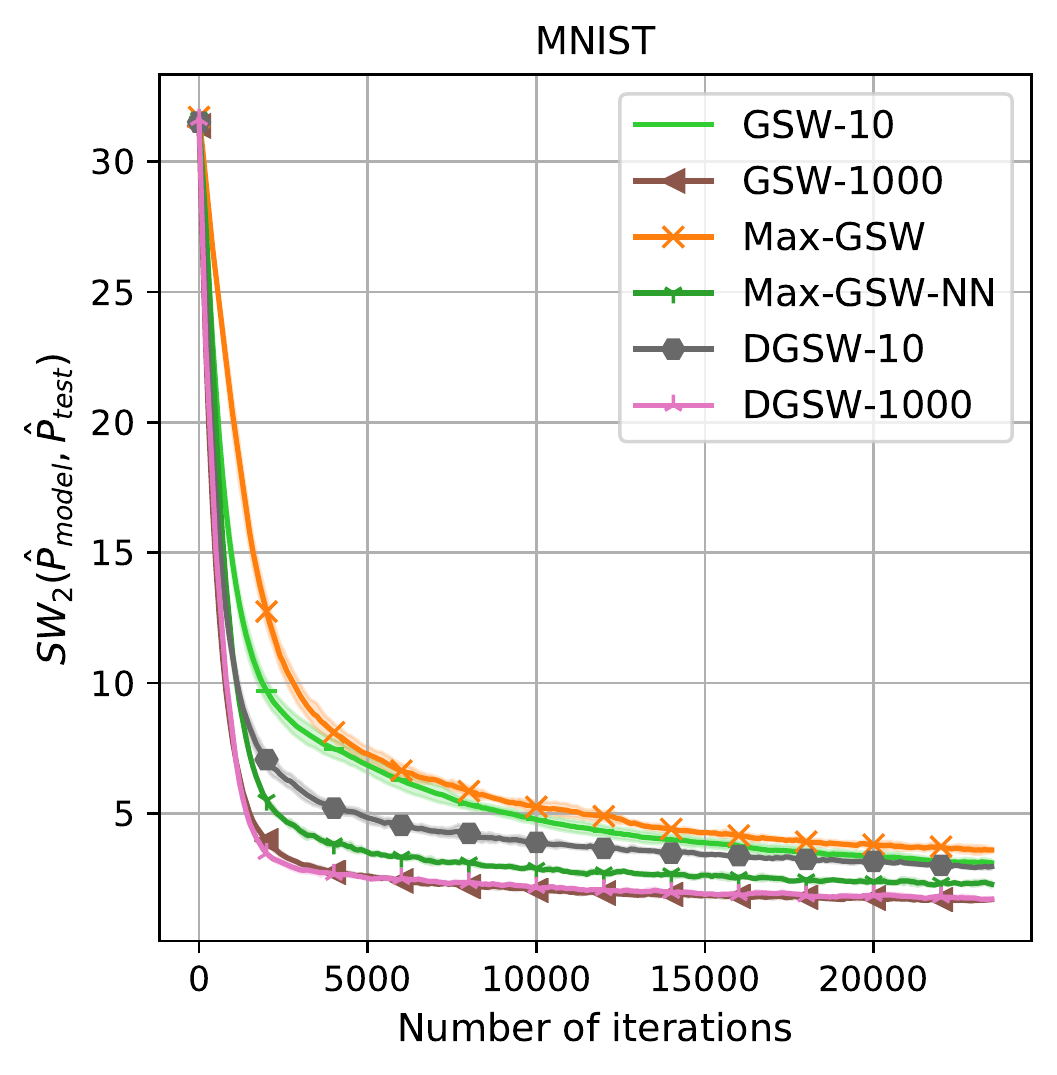}
    \label{fig:sub7}
    &
    \widgraph{0.21\textwidth}{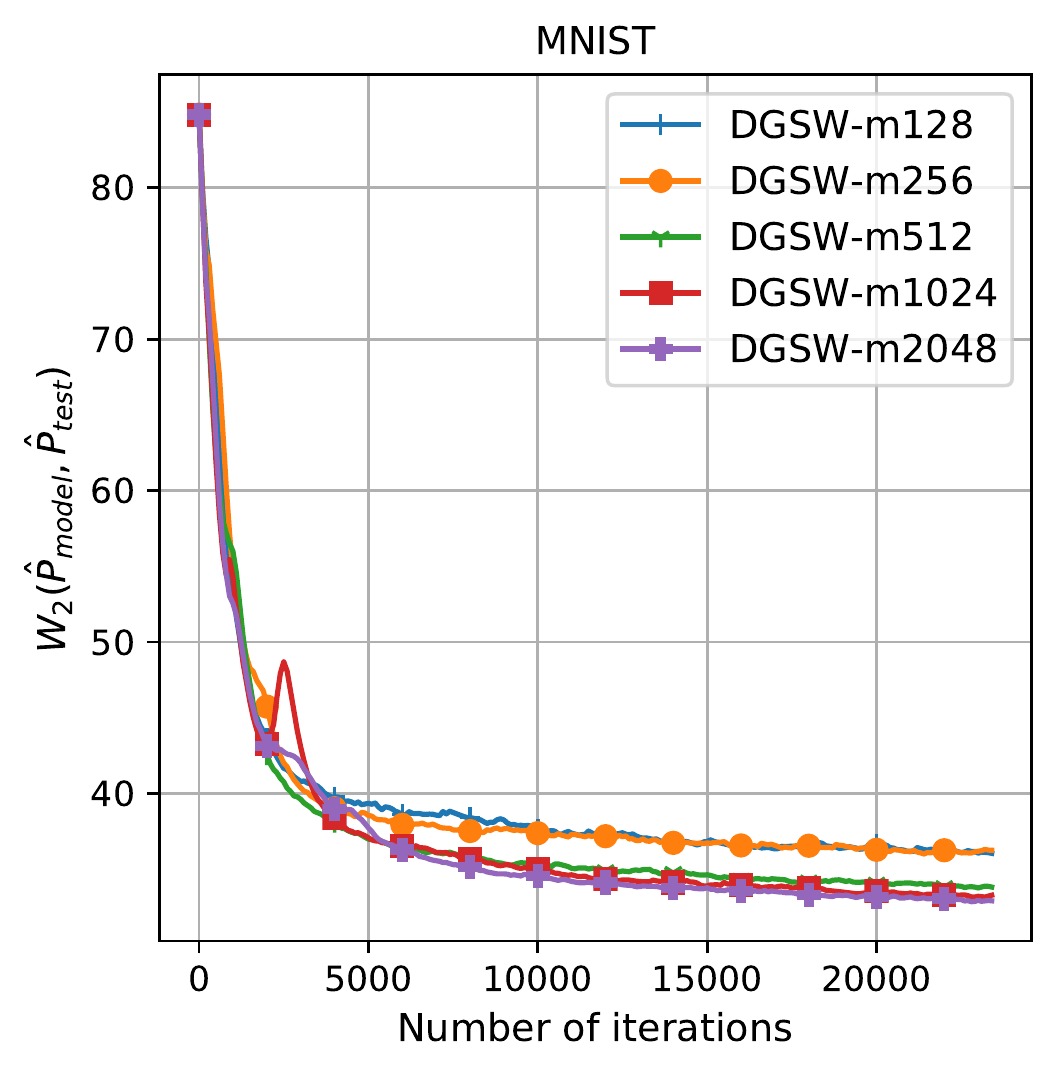}
    \label{fig:sub8}
    \\
    (a) &(b)&(c) &(d)
  \end{tabular}
 \end{center}
  \caption{
  \footnotesize{(a) Comparison between performances of DSW to SW, Max-SW and Max-GSW-NN using $SW_2$ distance as metric. (b) The effect of number of samples in minibatch to the convergence of DSW. (c) Comparing DGSW to GSW and Max-GSW-NN using $SW_2$ distance as metric. Here, GSW and DGSW use circular function. (d) The effect of number of samples in minibatch to the convergence of DGSW.
    }}
  \label{fig:addMNISTgraphs}
\end{figure}

\begin{figure}[!ht]
  \begin{tabular}{cccc}
    \widgraph{0.22\textwidth}{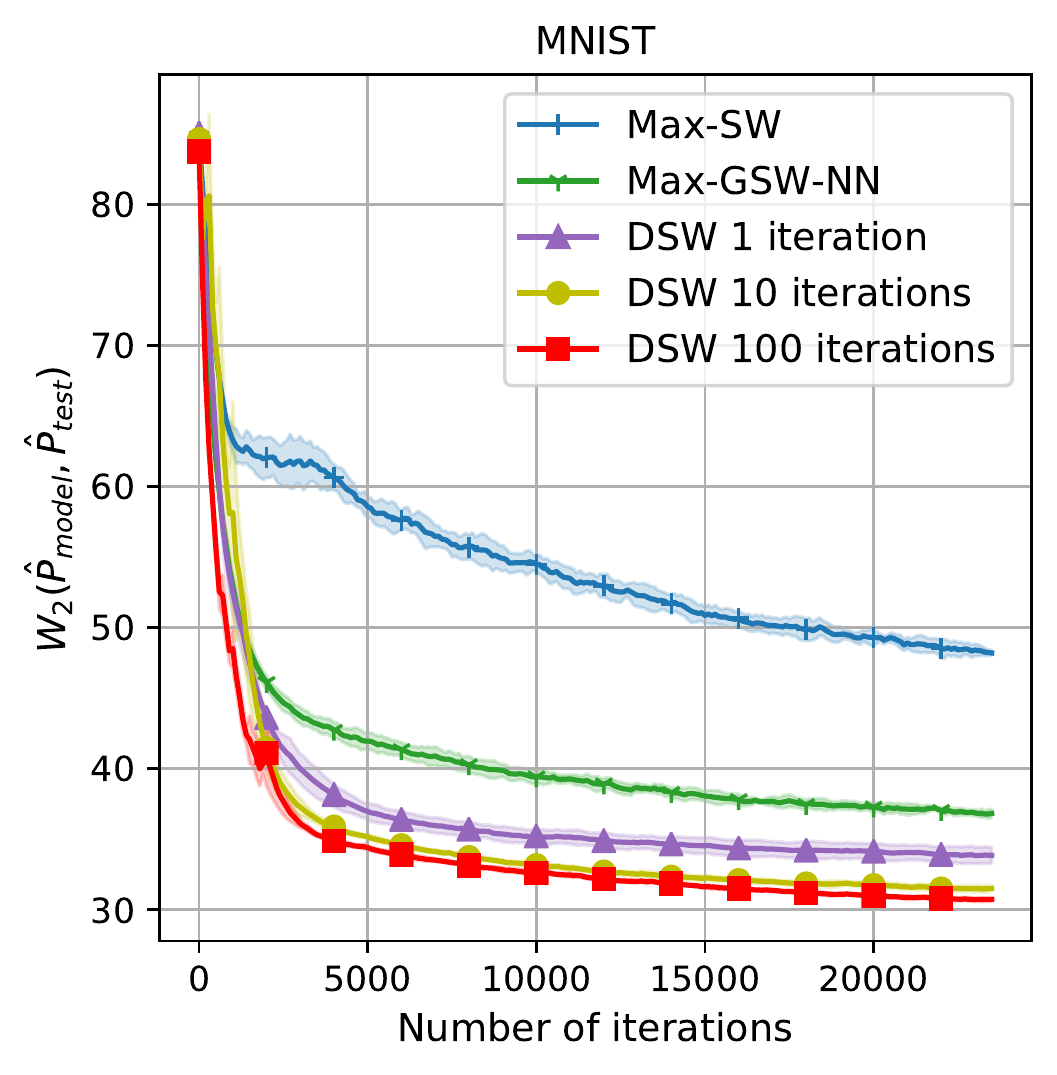}
    \label{fig:sub9}
  &
    \widgraph{0.22\textwidth}{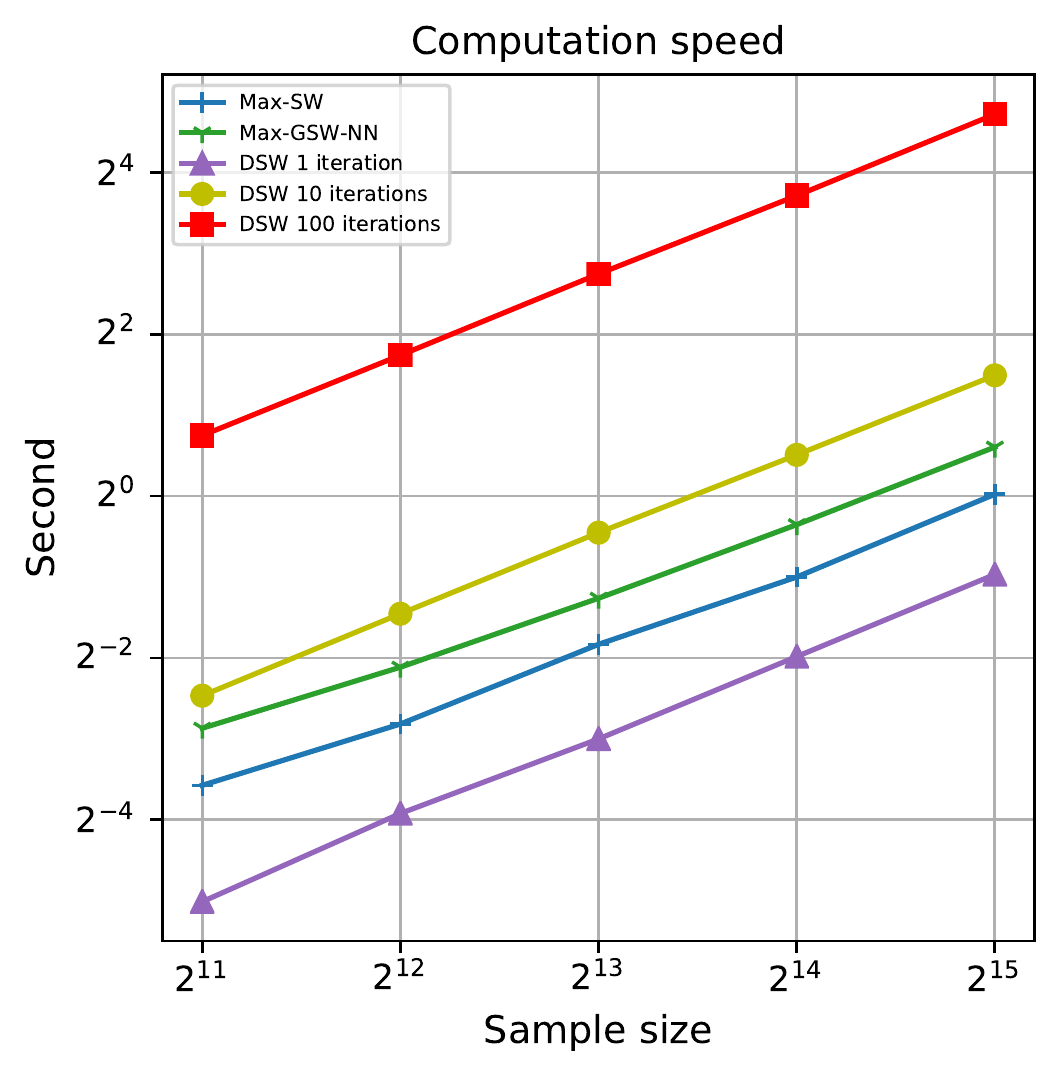}
    \label{fig:sub10}
    &
    \widgraph{0.22\textwidth}{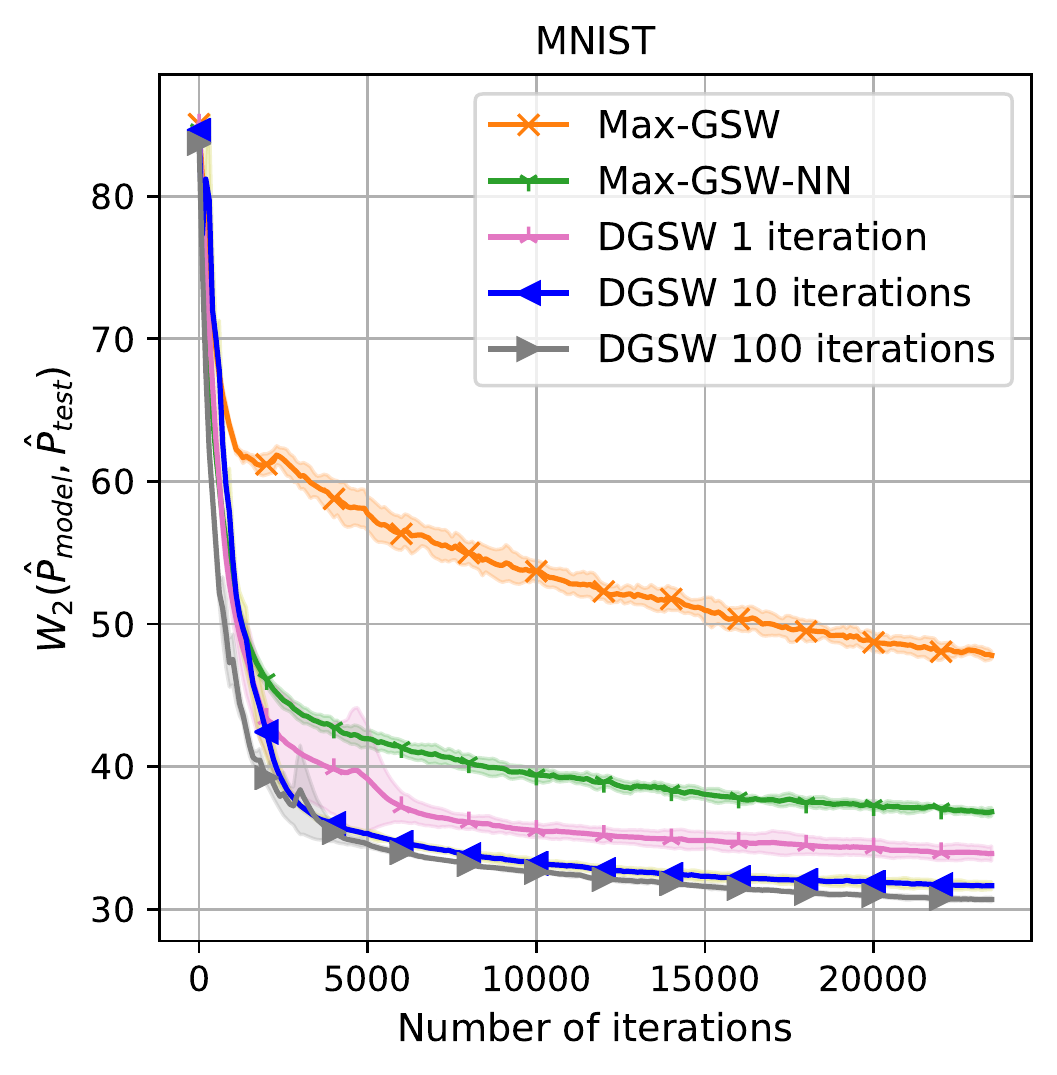}
    \label{fig:sub11}
    &
    \widgraph{0.22\textwidth}{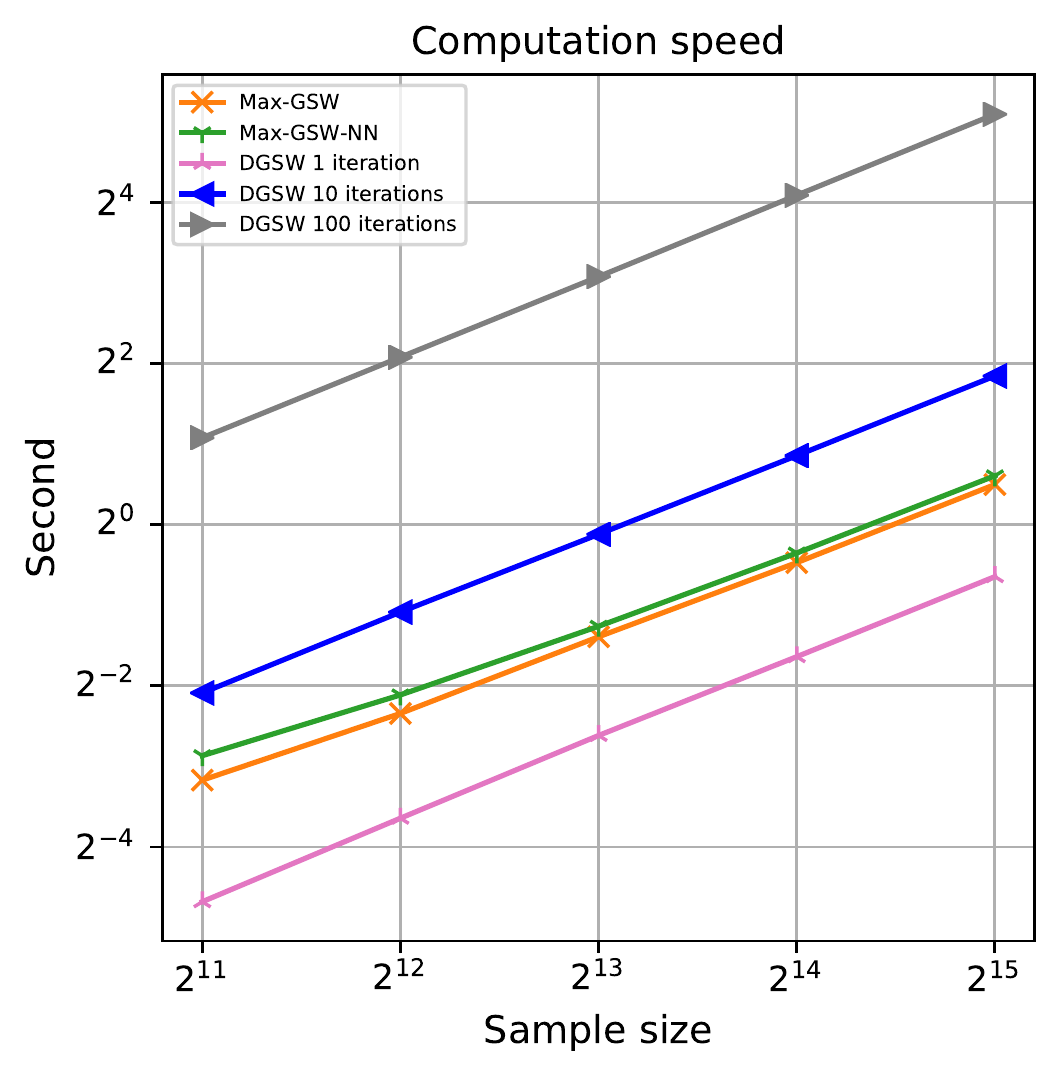}
    \label{fig:sub12}
    \\
    (a) & (b) & (c) &(d)
  \end{tabular}
  \caption{
  \footnotesize{(a) and (c): Increasing the number of times to update push forward measure can improve the performance of both DSW and  DGSW; (b) and (d): However, increasing the number of times to update push forward measure leads to much slower computation speed. 
    }}
  \label{fig:addMNISTexperiment}
\end{figure}

\textbf{Effects of the number of samples:} We conduct experiments to show how sample size  ($m$ in Appendix \ref{subsec:MEDE}) affects the results of DSW and DGSW in the MEDE framework. According to Figure \ref{fig:addMNISTgraphs}(b), increasing the sample size leads to better performance of DSW. Similarly, increasing the sample size in the MEDE framework that uses DGSW (with circular defining function) helps improve the results, see in Figure~\ref{fig:addMNISTgraphs}(d).
\paragraph{Effects of the number of gradient-updates:}  In both DSW and DGSW cases, we use a pushforward measure for the distribution over the sphere, and we use neural nets to find it. To learn these neural nets, we use gradient ascent to update their weights. In this experiment, we aim to find out how the number of iterations to update these neural net, affects the performance including the convergence behavior and computation speed. By increasing the number of updates from 1 to 10, both in DSW and DGSW, model distributions are much closer to data distribution; from 10 to 100 updates the results are improved but not too much, see the results in Figures~\ref{fig:addMNISTexperiment}(a) and~\ref{fig:addMNISTexperiment}(c). However, increasing update steps also lead to a computation problem as the al time increases considerably. When using 10 or 100 update steps, DSW and DGSW are slower than Max-SW, Max-GSW (50 gradient updates to find the max direction), and Max-GSW-NN (50 update times for the defining neural net function).  

\begin{table}[!h]

\caption{FID score of generator models trained on CIFAR10 (100 epochs), CelebA (50 epochs), and LSUN (20 epochs) datasets in 64x64 resolution. Results are averaged from 5 different runs. }
\vskip 0.15in
\begin{center}
\begin{tabular}{lllll}
\toprule
Model & $n$ &CIFAR-10  &CelebA&LSUN\\
\midrule
SW & $10^2$ & 109.7 $\pm$ 5.64 & 90.11 $\pm$ 10.11   &  101.57 $\pm$ 3.24 \\
GSW &$10^2$ & 103.11 $\pm$ 6.92 & 87.18 $\pm$ 8.97 &92.58 $\pm$ 4.78\\
ASW &$10^2$ & 138.26 $\pm$ 8.31 &122.11 $\pm$9.09 \\
DSW  & $10^2$ & \textbf{62.83 $\pm$ 6.24}   & 75.94$\pm$ 5.54 & \textbf{46.02 $\pm$ 2.15}  \\ 
DGSW &$10^2$ & 68.01 $\pm$  7.74&  \textbf{71.08$\pm$ 4.24 } & 46.91 $\pm$ 3.98\\

\midrule
Max-SW&   & 136.04 $\pm$ 8.35 &100.09 $\pm$ 8.34 & 123.74 $\pm$ 5.51 \\
Max-GSW-NN& &86.04$\pm$ 8.68  &81.57$\pm$7.72  & 56.83 $\pm$ 4.04 \\
SW & $10^4$ &98.61 $\pm$ 3.62   & 82.02 $\pm$ 6.33   & 62.75 $\pm$ 4.77 \\
GSW & $10^4$ &93.51 $\pm$ 6.12   & 84.22 $\pm$ 7.93   & 68.04 $\pm$ 2.17 \\
ASW &$10^4$ & 121.38 $\pm$ 6.83 & 101 $\pm$ 7.36\\
DSW& $10^4$ & \textbf{56.42 $\pm$ 3.78} & 66.85$\pm$ 7.22 & \textbf{39.68 $\pm$ 2.33} \\
DGSW& $10^4$ & 60.01 $\pm$ 5.58 & \textbf{65.8$\pm$ 4.42} & 42.04 $\pm$ 4.21 \\

\bottomrule

\end{tabular}
\label{table:appendGANFID}

\end{center}
\vskip -0.15in
\end{table}

\begin{table}[!h]

\caption{Computational speed per minibatch on CelebA and CIFAR10 dataset}
\vskip 0.15in
\begin{center}
\begin{tabular}{llc}
\toprule
Model & $n$ &Second/Minibatch\\
\midrule
SW & $10^2$ &0.178  \\
GSW &$10^2$ & 0.181\\
ASW &$10^2$ &  0.298\\
DSW  & $10^2$ & 0.21 \\ 
DGSW &$10^2$ & 0.212\\

\midrule
Max-SW&   & 1.821 \\
Max-GSW-NN& &1.895\\
SW & $10^4$ &0.615\\
GSW & $10^4$ & 0.632\\
ASW &$10^4$ & 1.561\\
DSW& $10^4$ & 1.312\\
DGSW& $10^4$ & 1.384\\

\bottomrule

\end{tabular}
\label{table:appendspeed}

\end{center}
\vskip -0.15in
\end{table}

\paragraph{Quantitative results:} We provide full FID scores of all distances mentioned in the papers and also the recent augmented sliced Wasserstein (ASW)~\cite{chen2020augmented} in Table~\ref{table:appendGANFID}. Based on the results in that table, DSW and DGSW (circular) achieve the best performance among all sliced distances. We also report the computational speed per minibatch in Table \ref{table:appendspeed}. The results show that DSW-100 is faster than DSW-10000 while its FID is lower. Regarding ASW, in our experiment, we find that the injective neural network, which is used to transform two target measures, is quite unstable to train and our obtained results with that distance are not good. Moreover, ASW is slower than DSW because ASW needs to double the dimension and still utilizes the uniform measure to slice on the new space. Note that, we use the implementation of ASW in \url{https://github.com/ShwanMario/ASWD}.

\begin{figure}[!ht]
\begin{center}
  \begin{tabular}{cccc}
    \widgraph{0.21\textwidth}{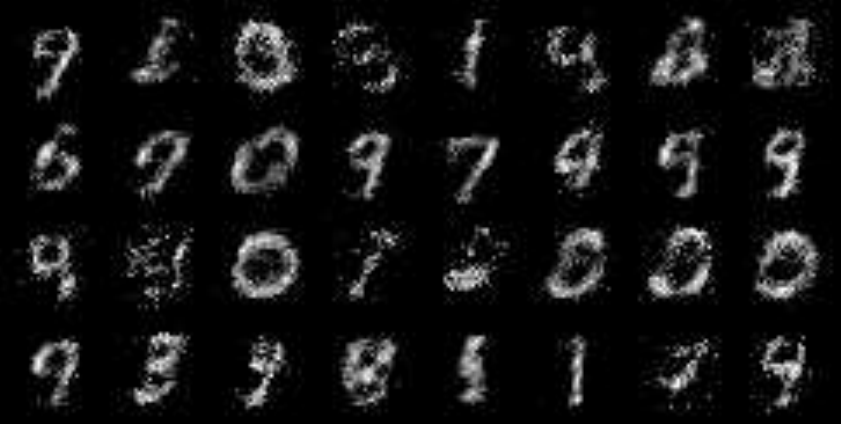}
    &
    \widgraph{0.21\textwidth}{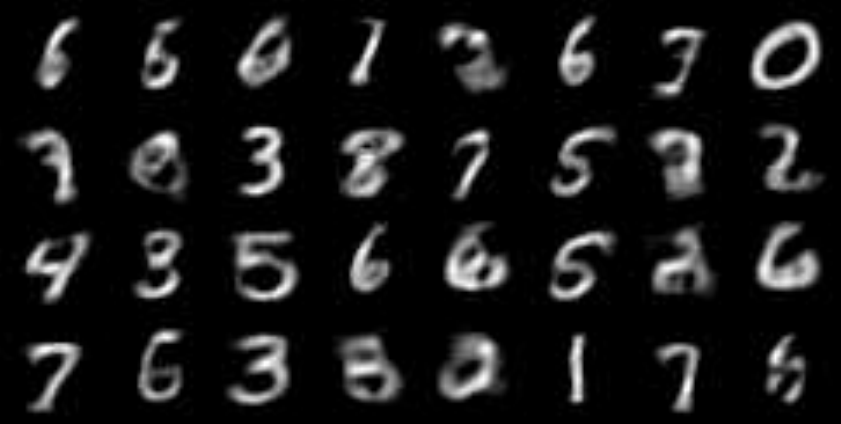}
    
    &
     \widgraph{0.21\textwidth}{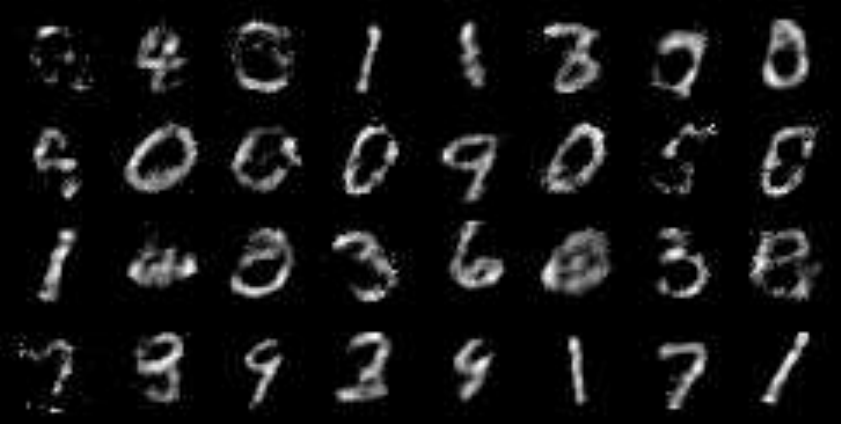}
    &
    \widgraph{0.21\textwidth}{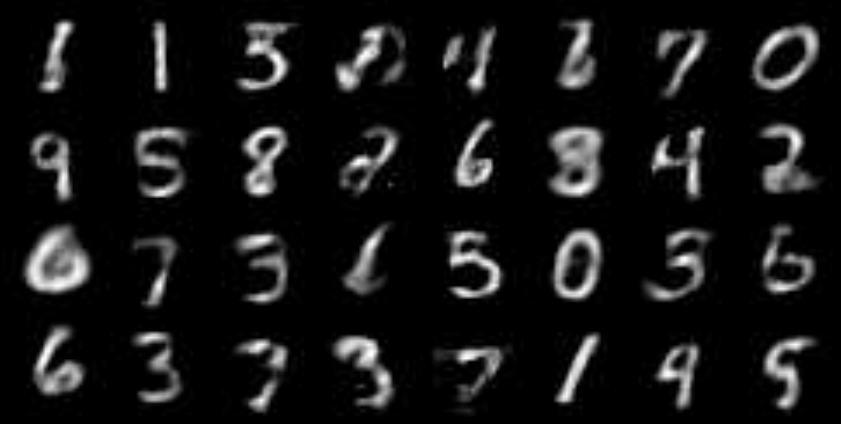}
    \\
      SW $n$=10 &  DSW $n$=10  &SW $n$=1000 &  DSW $n$=1000 
\\
     \widgraph{0.21\textwidth}{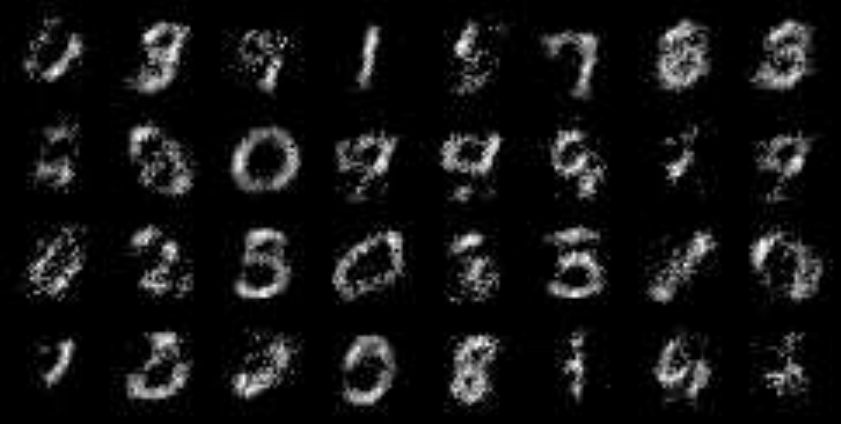}
    &
 \widgraph{0.21\textwidth}{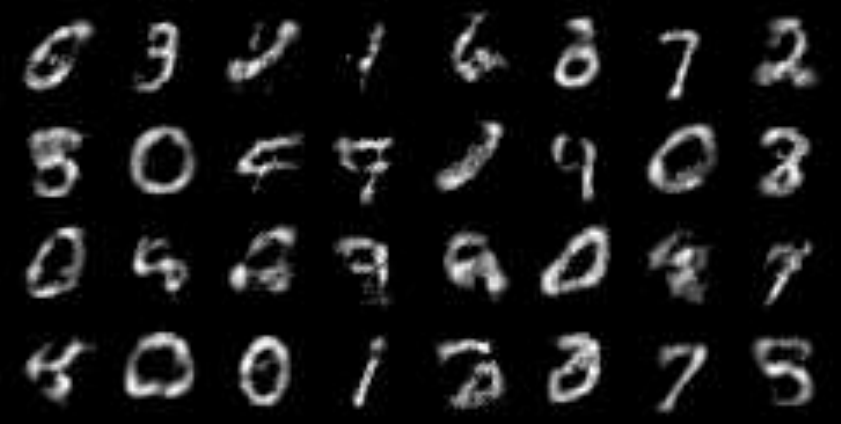}
    
    &
     \widgraph{0.21\textwidth}{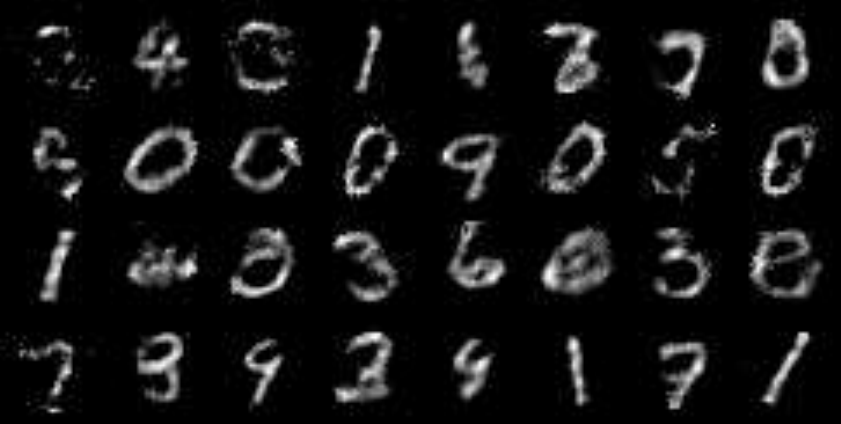}
    &
     \widgraph{0.21\textwidth}{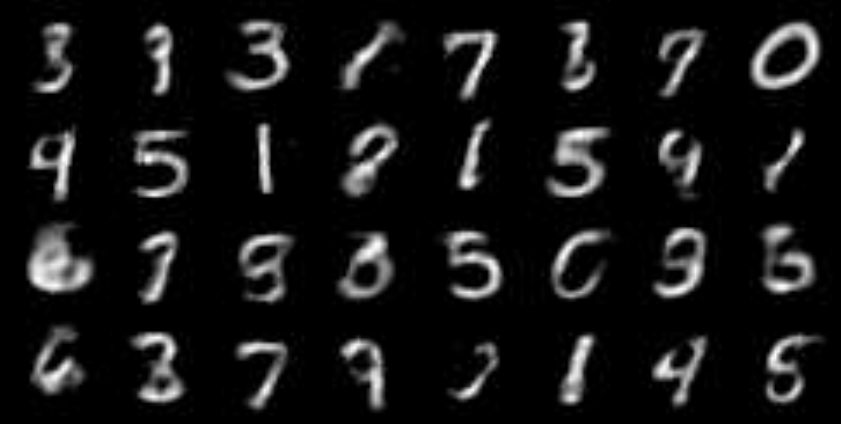}
    \\
      Max-SW&   Max-GSW-NN & GSW $n$=1000 &  DGSW $n$=1000 
 
  \end{tabular}
      
\end{center}
  \caption{ MNIST  generated images from different generators, $n$ is the number of projections.
    }
  \label{fig:MNISTgenimages}
\end{figure}

\begin{figure}[t]
\begin{center}

  \begin{tabular}{cccc}
    \widgraph{0.21\textwidth}{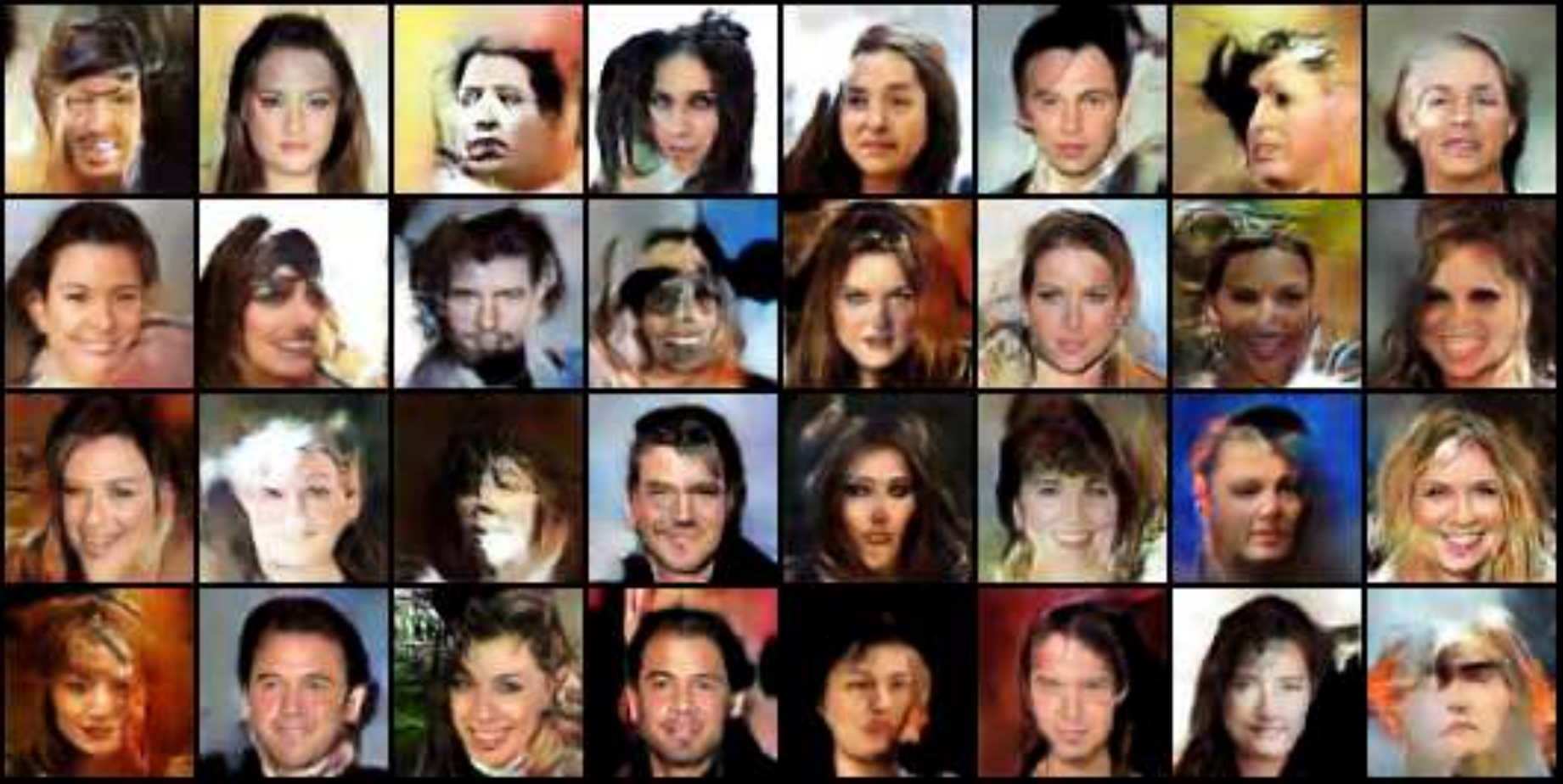}
    &
    \widgraph{0.21\textwidth}{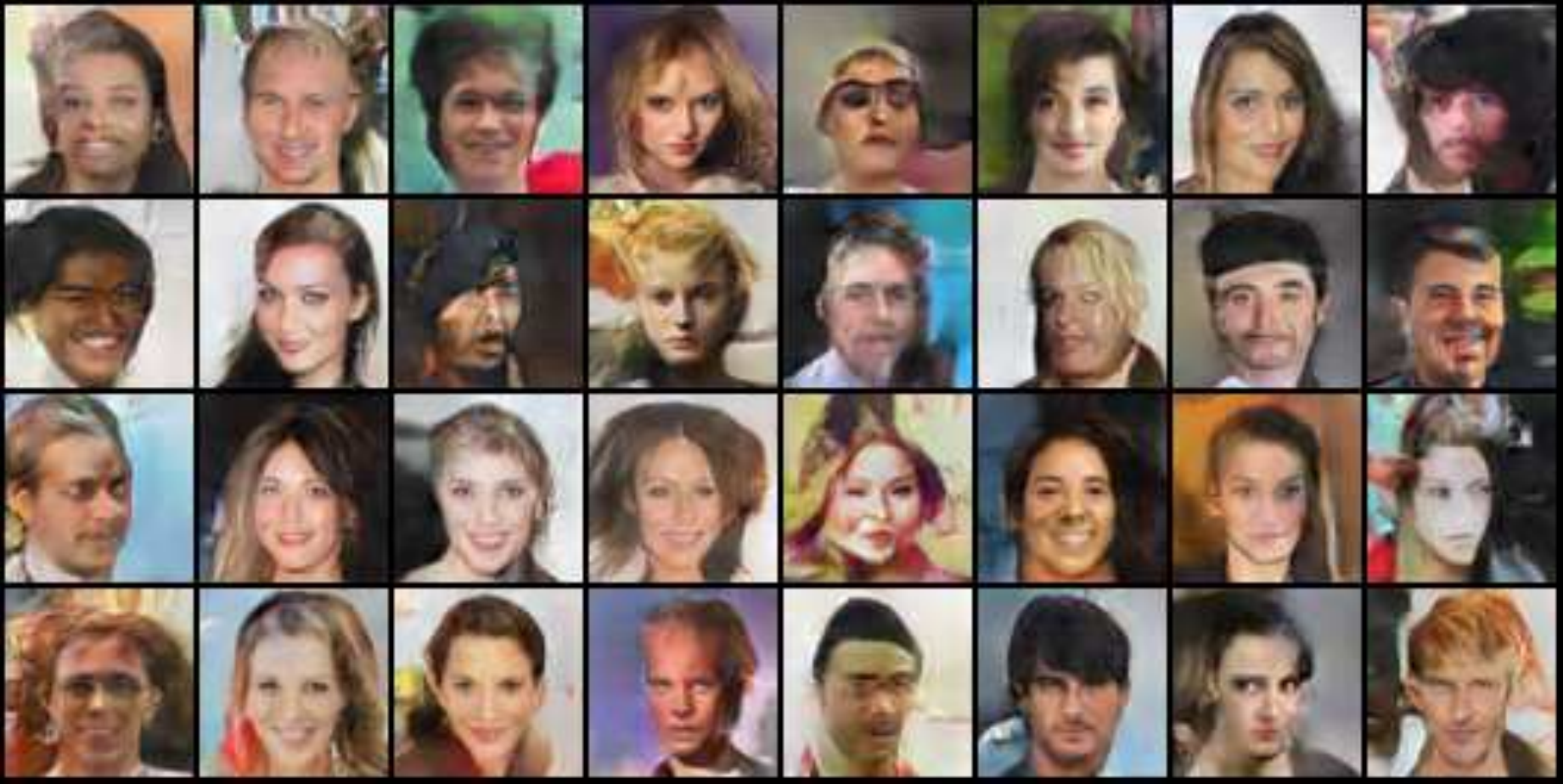}
    
    &
     \widgraph{0.21\textwidth}{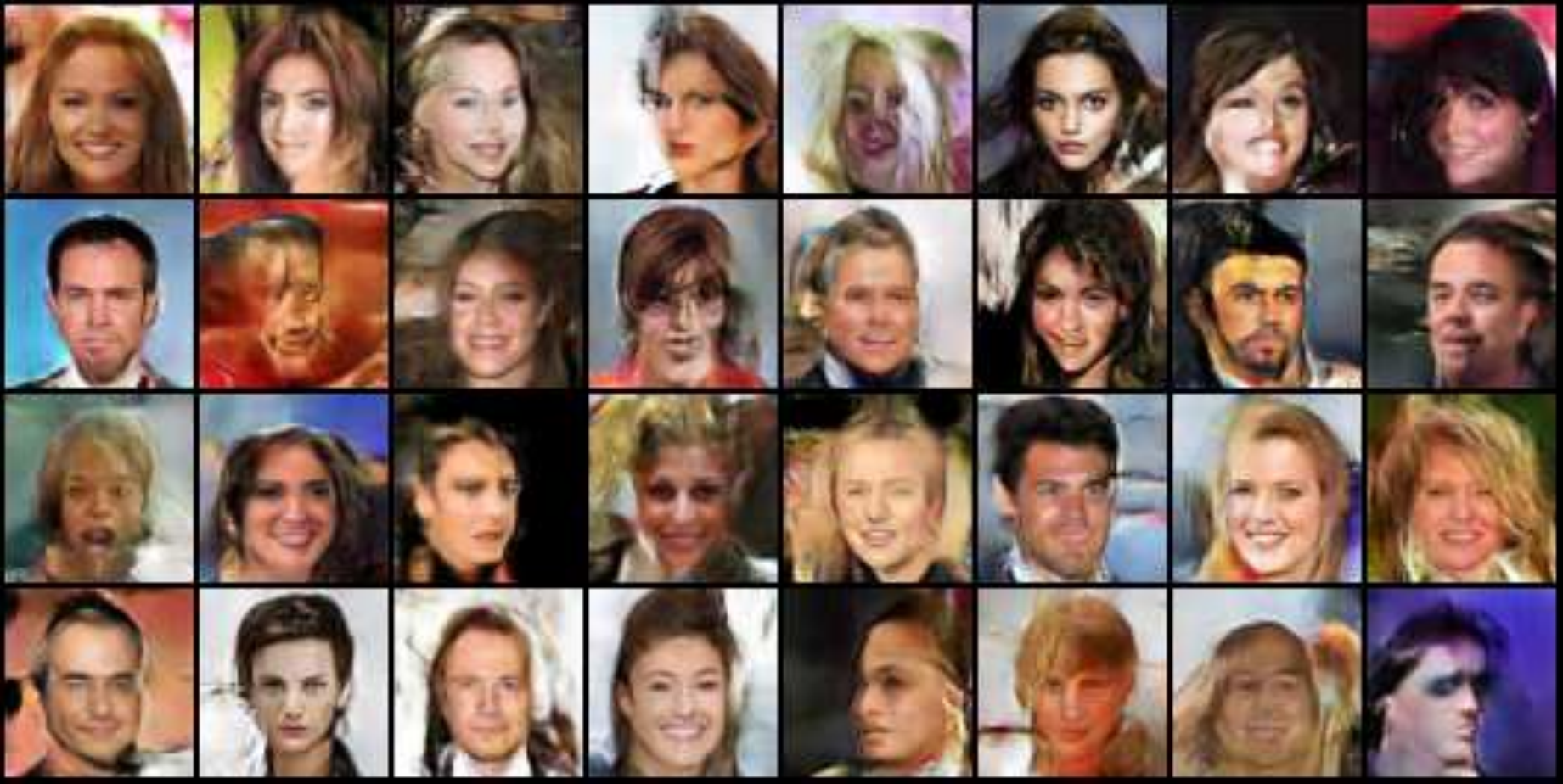}
    &
    \widgraph{0.21\textwidth}{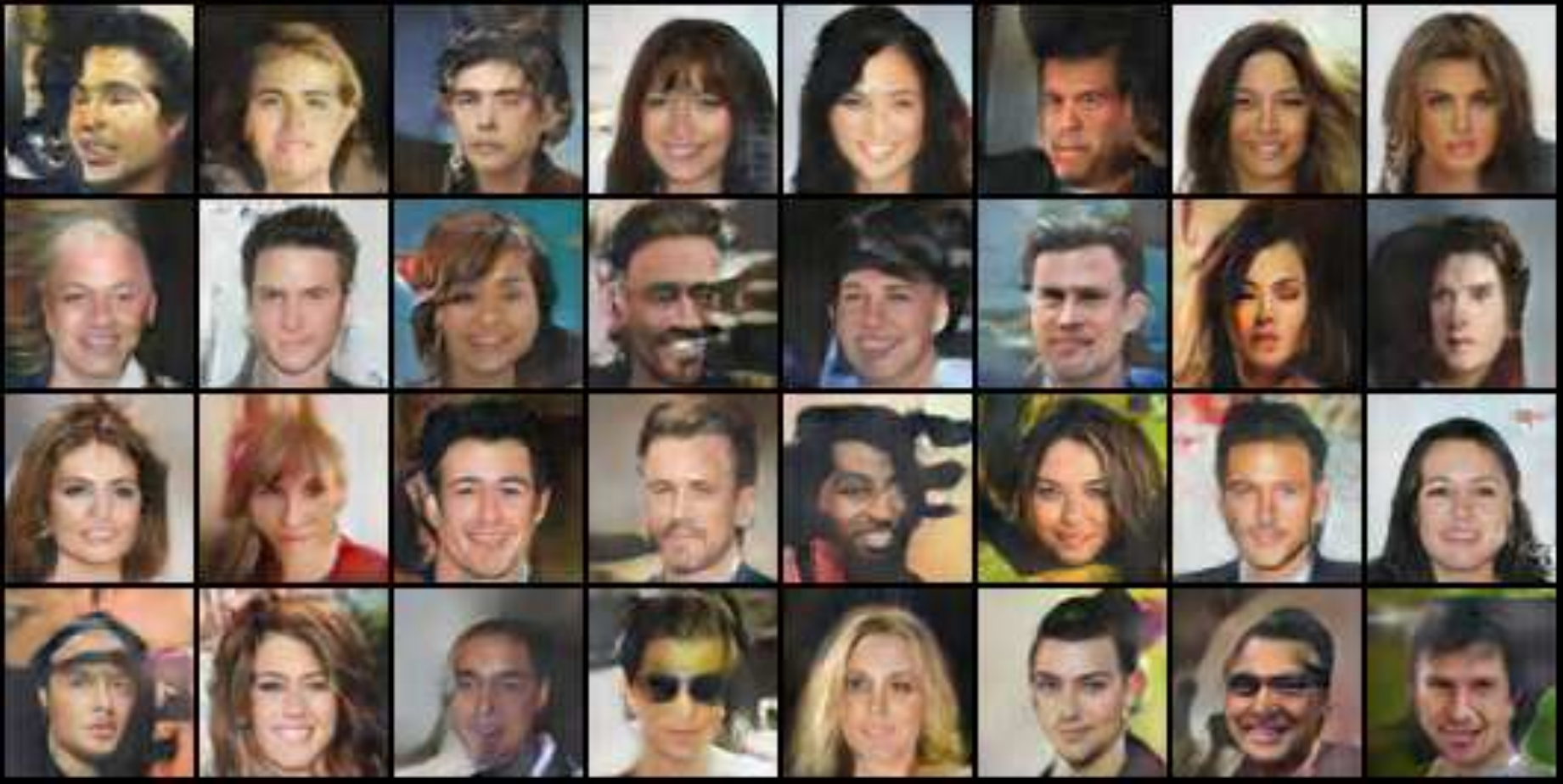}
    \\
      SW $n$=100 &  DSW $n$=100  &SW $n$=10000 &  DSW $n$=10000 
\\
     \widgraph{0.21\textwidth}{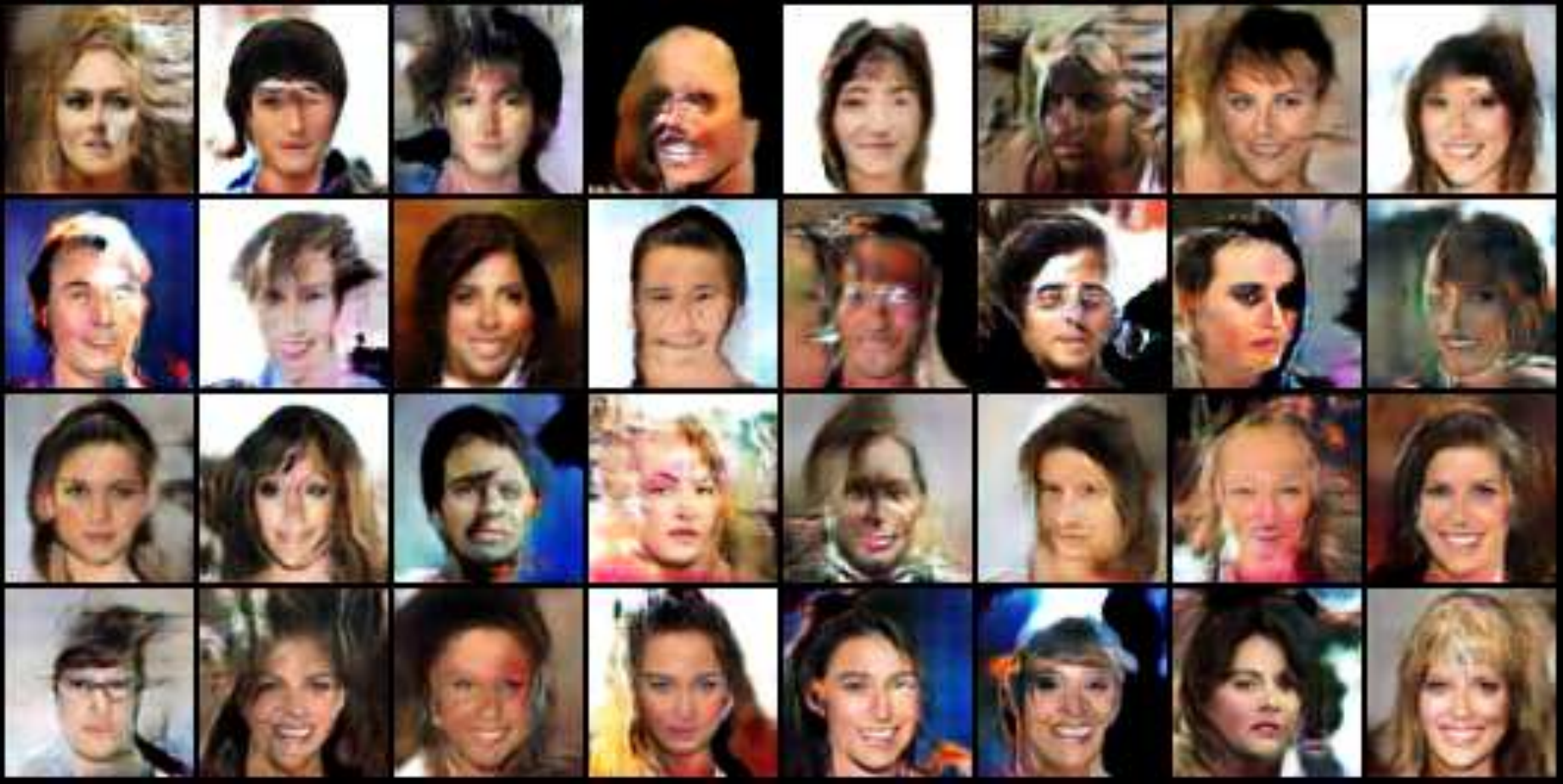}
    &
 \widgraph{0.21\textwidth}{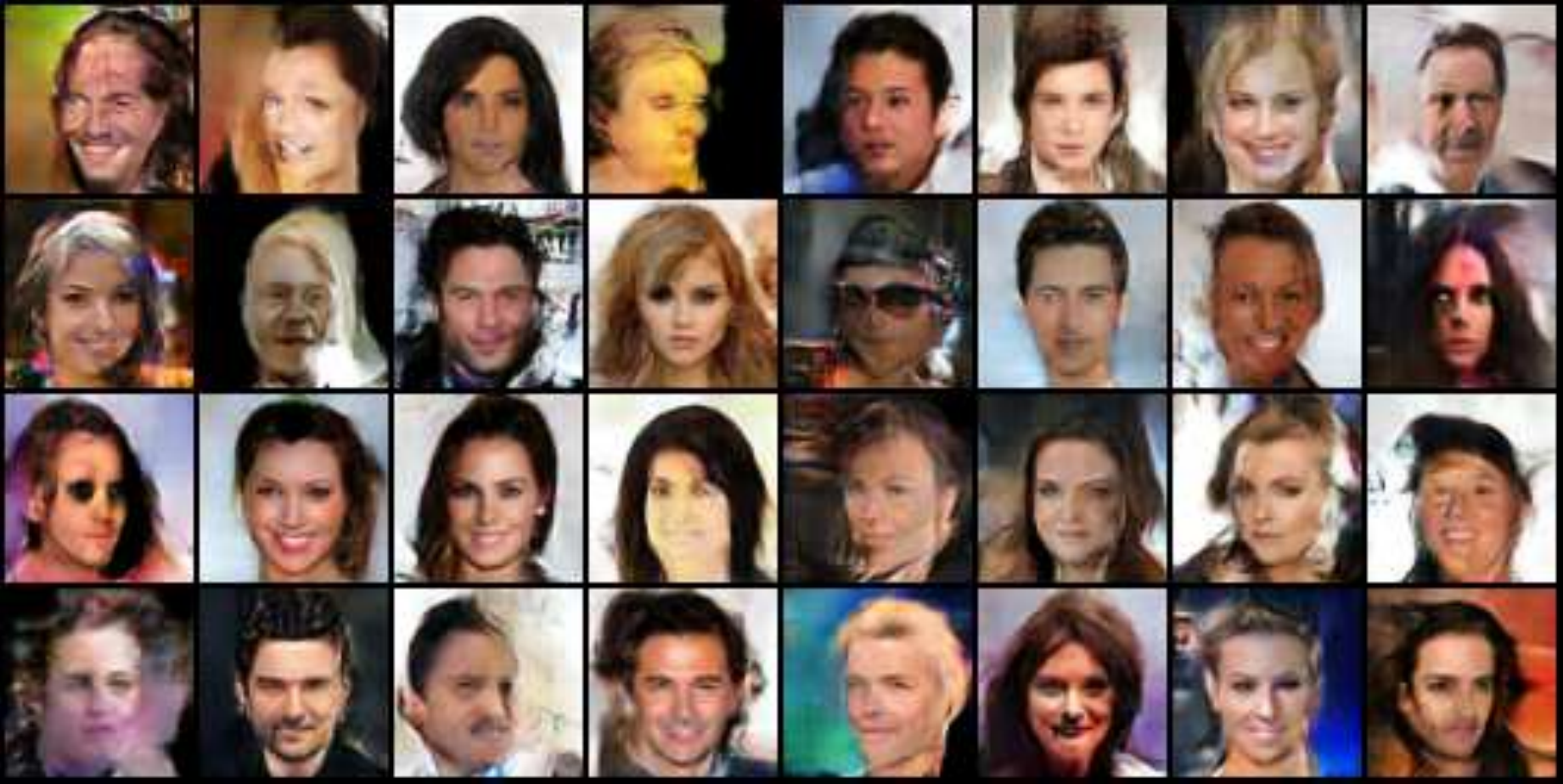}
    
    &
     \widgraph{0.21\textwidth}{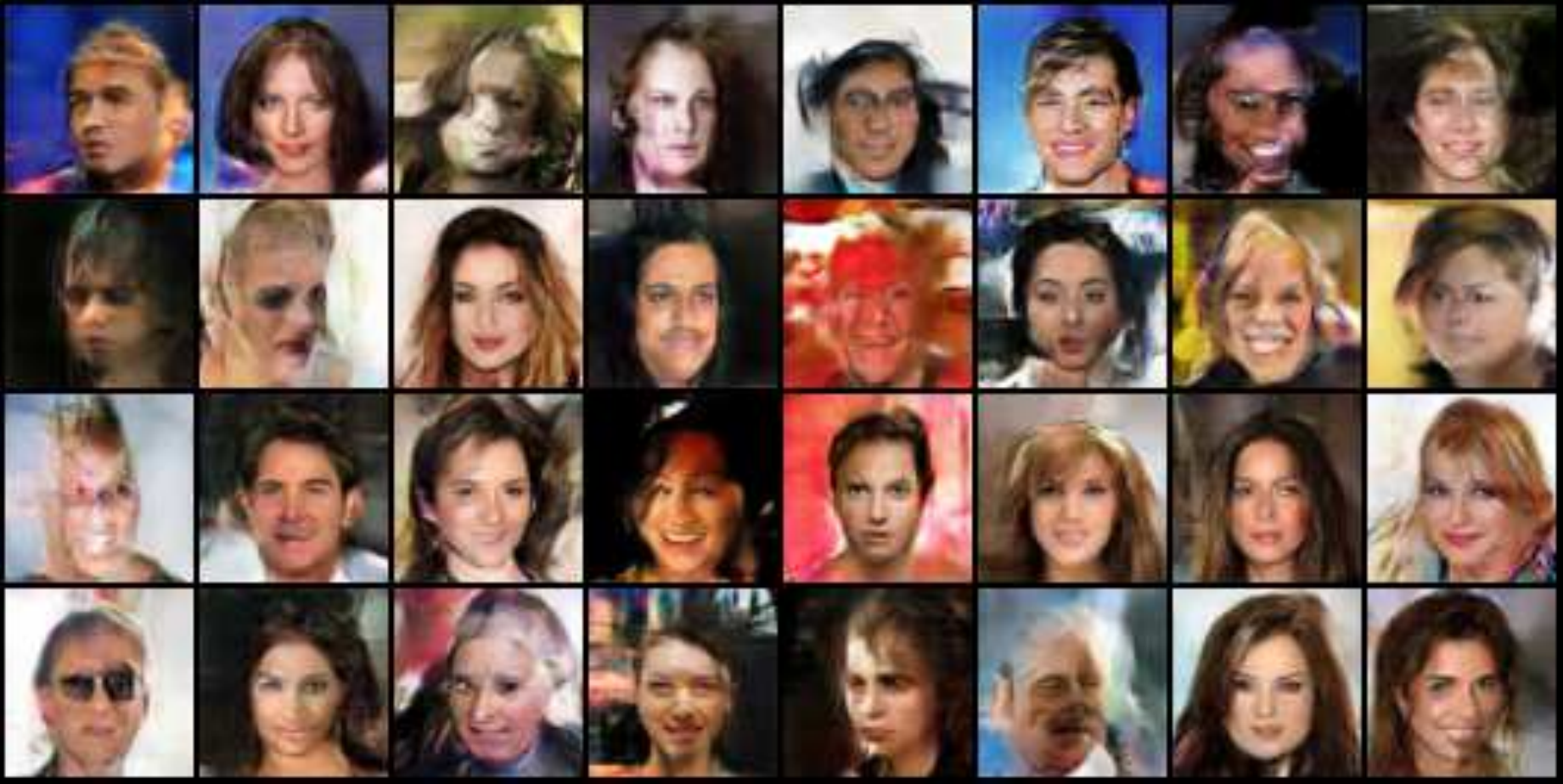}
    &
     \widgraph{0.21\textwidth}{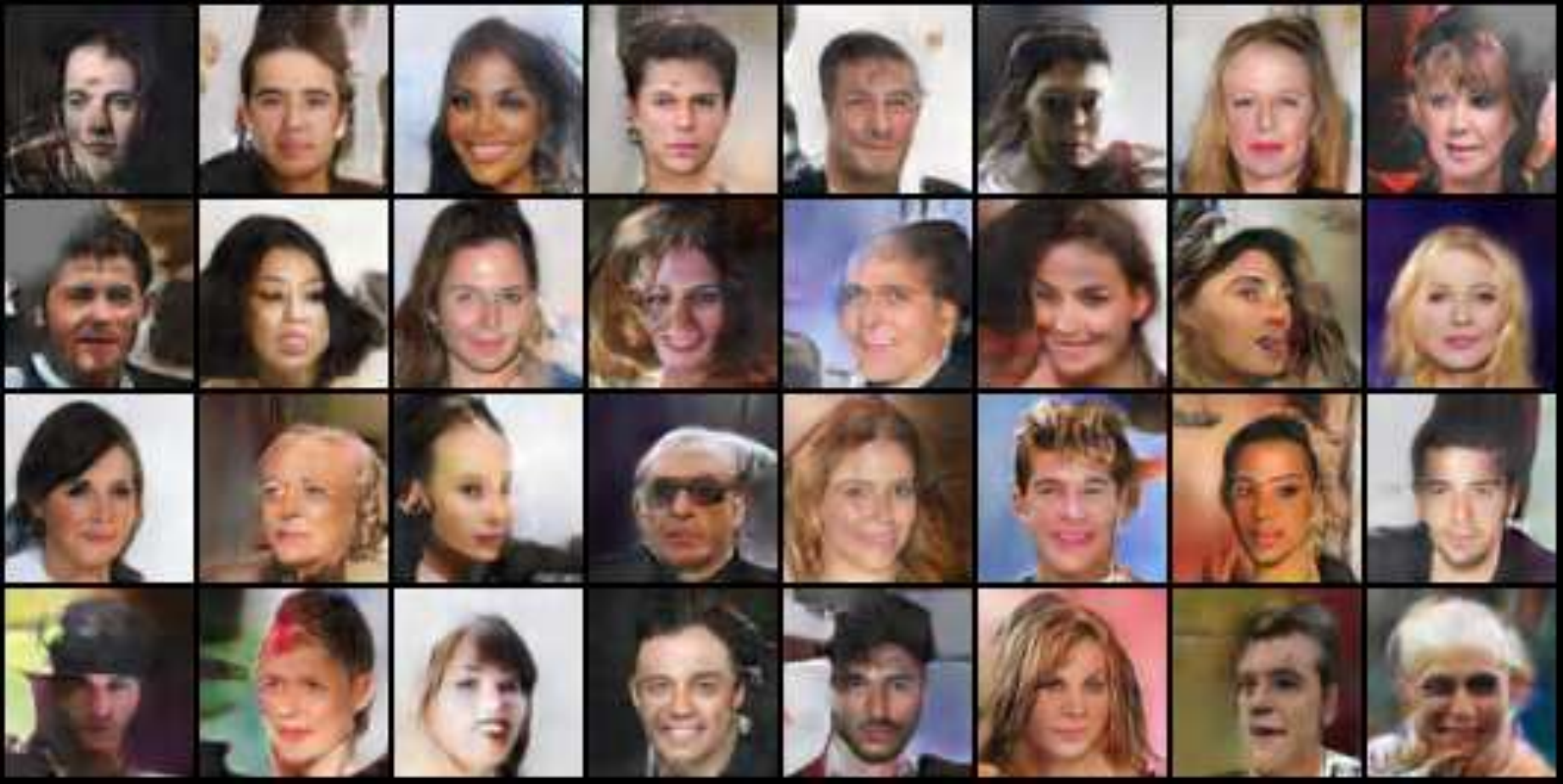}
    \\
      Max-SW&   Max-GSW-NN & GSW $n$=10000 &  DGSW $n$=10000 
 
  \end{tabular}
  \end{center}
  \caption{ CelebA  generated images from different generators, $n$ is the number of projections.
    }
    \vspace{-1 em}
  \label{fig:CELEBgenimages}
\end{figure}

\begin{figure}[!t]
 \begin{center}

  \begin{tabular}{cccc}
    \widgraph{0.21\textwidth}{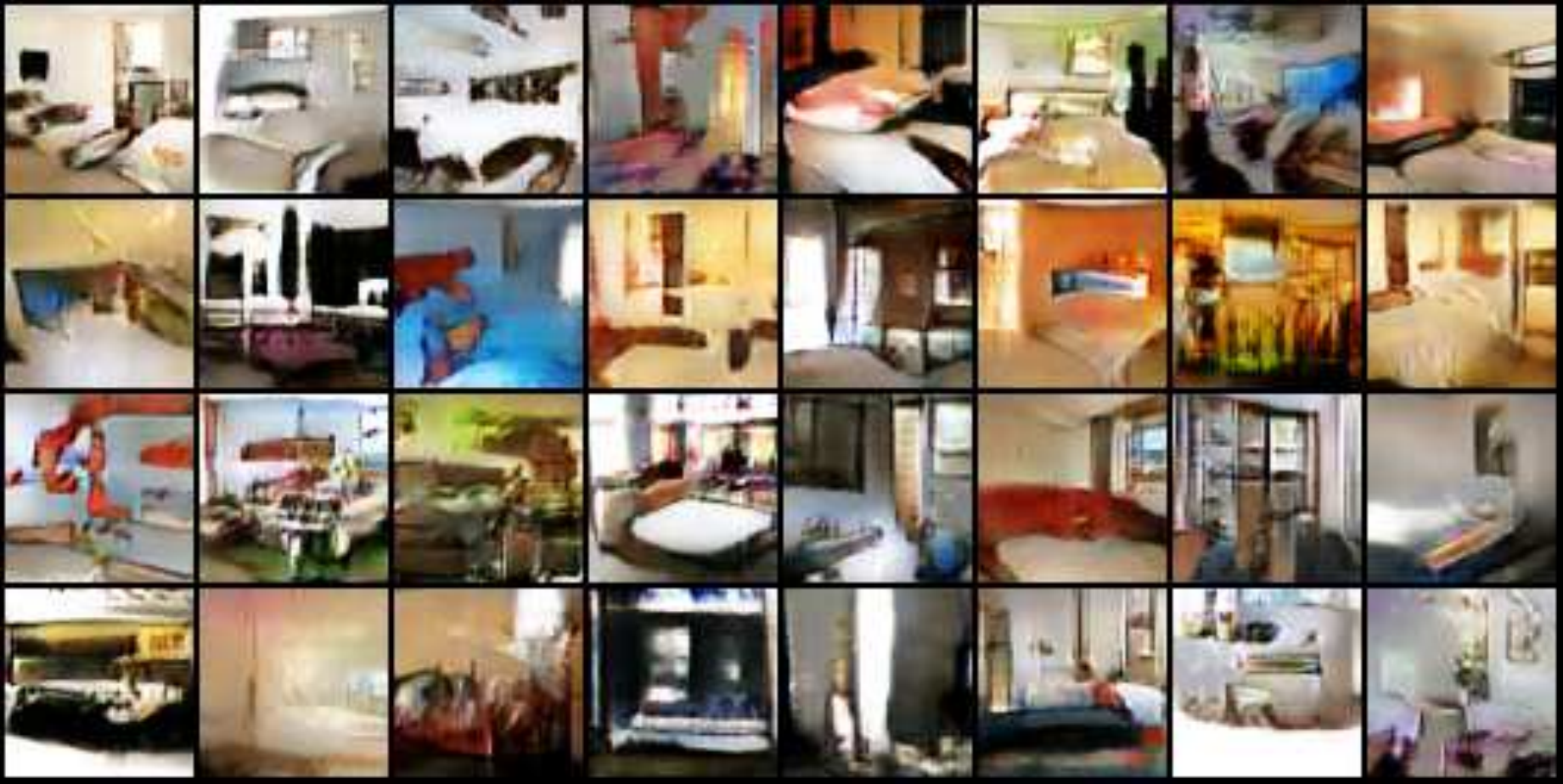}
    &
    \widgraph{0.21\textwidth}{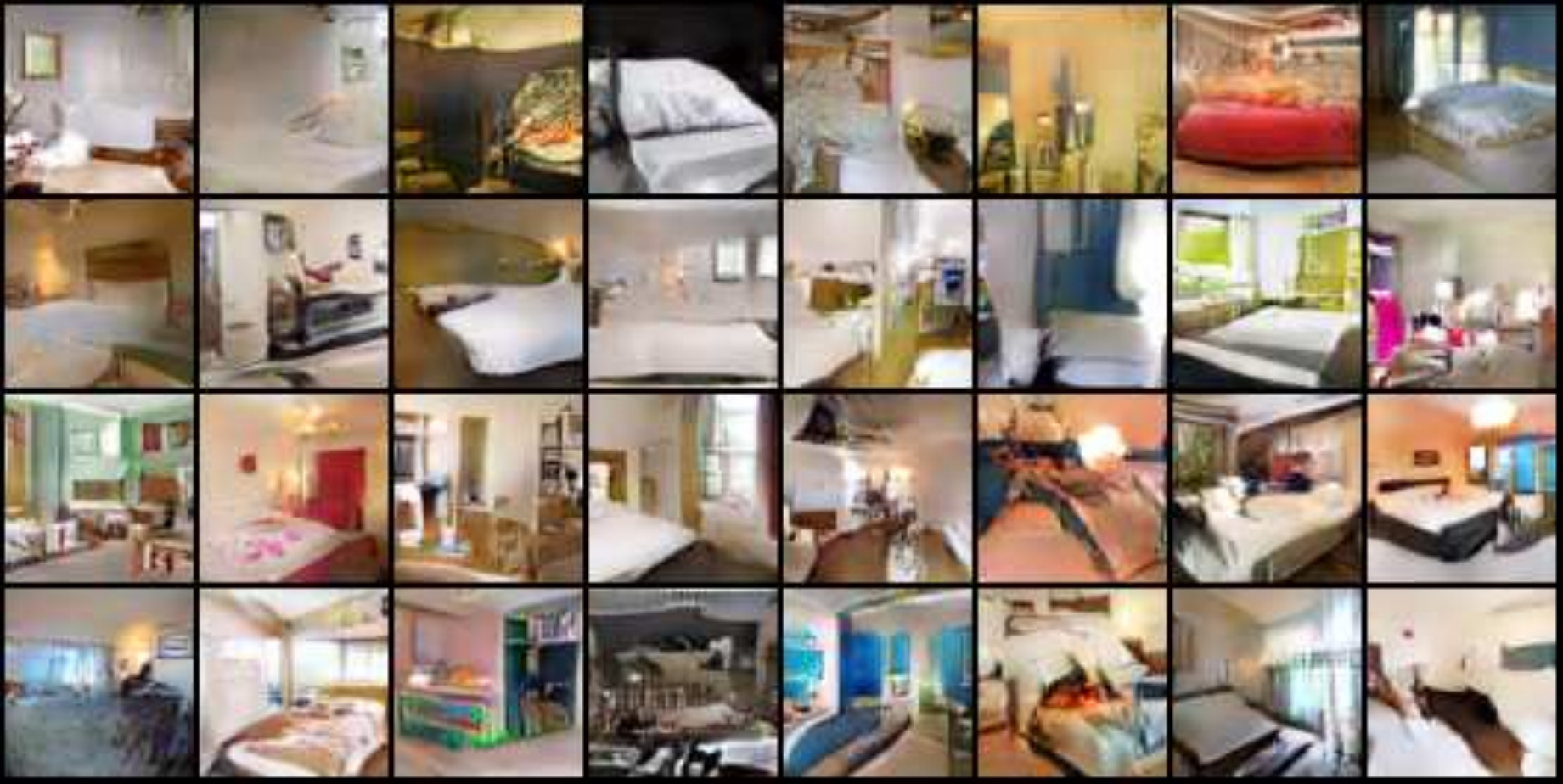}
    
    &
     \widgraph{0.21\textwidth}{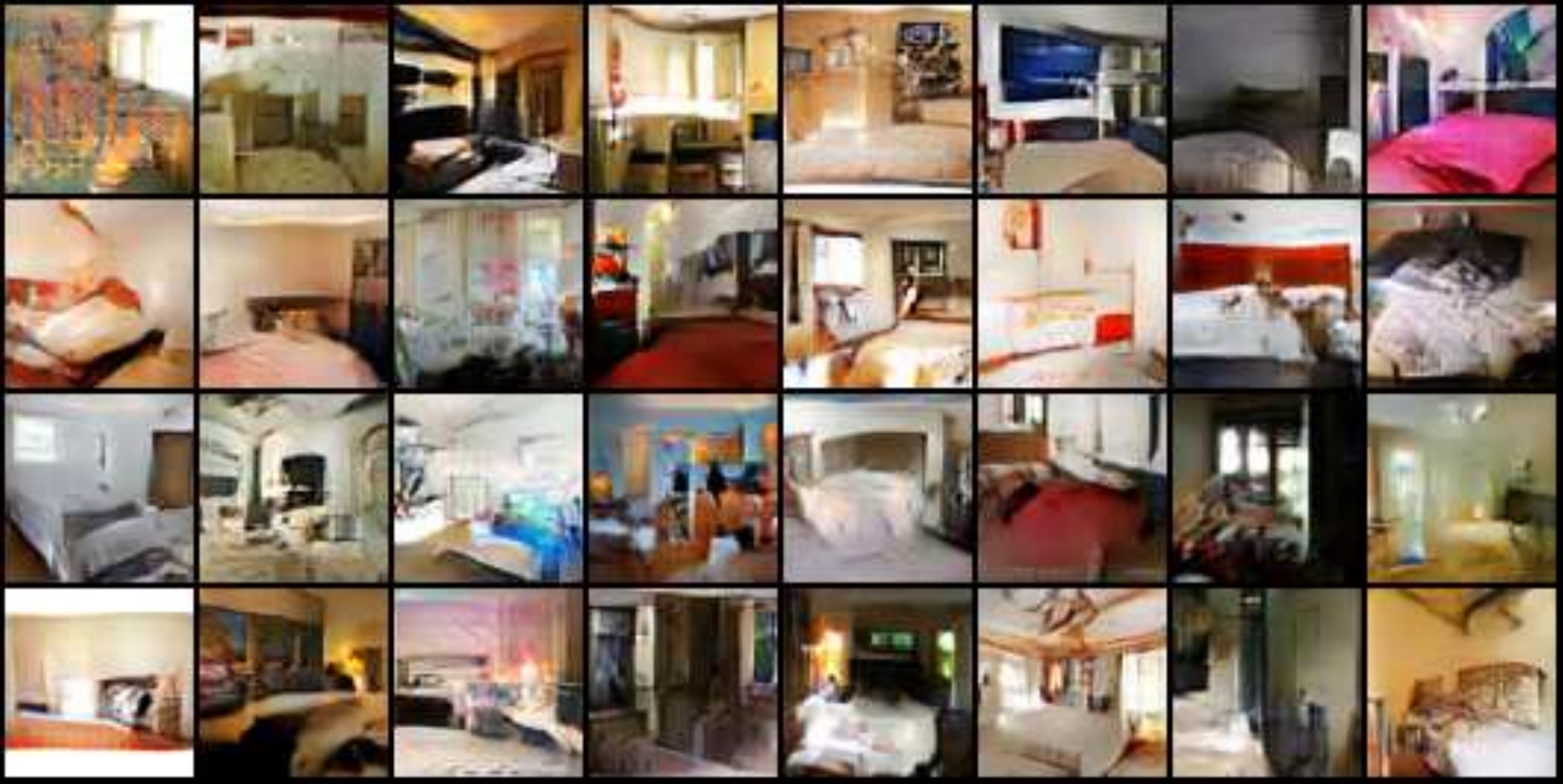}
    &
    \widgraph{0.21\textwidth}{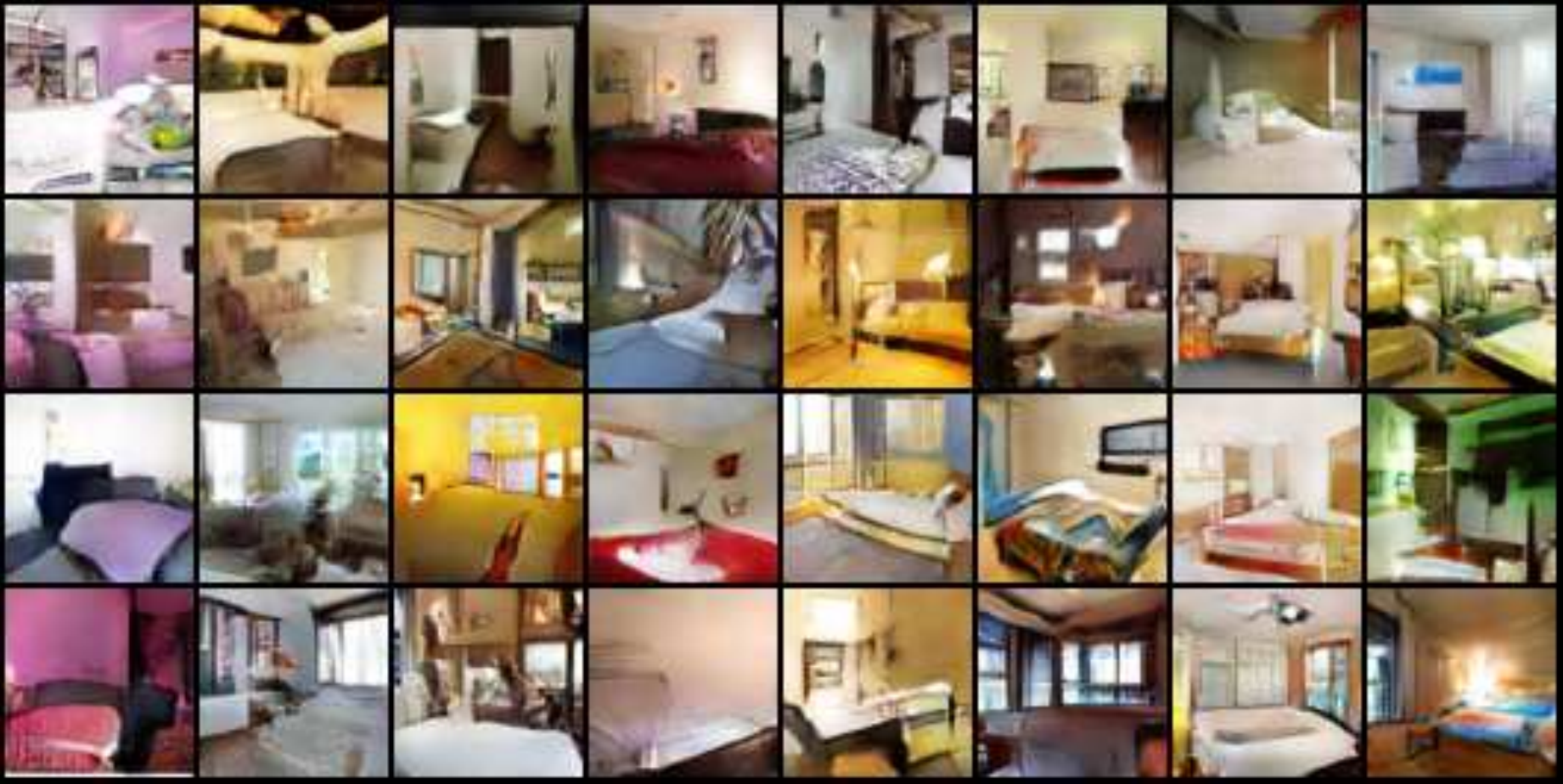}
    \\
      SW $n$=100 &  DSW $n$=100  &SW $n$=10000 &  DSW $n$=10000 
     \\
     \widgraph{0.21\textwidth}{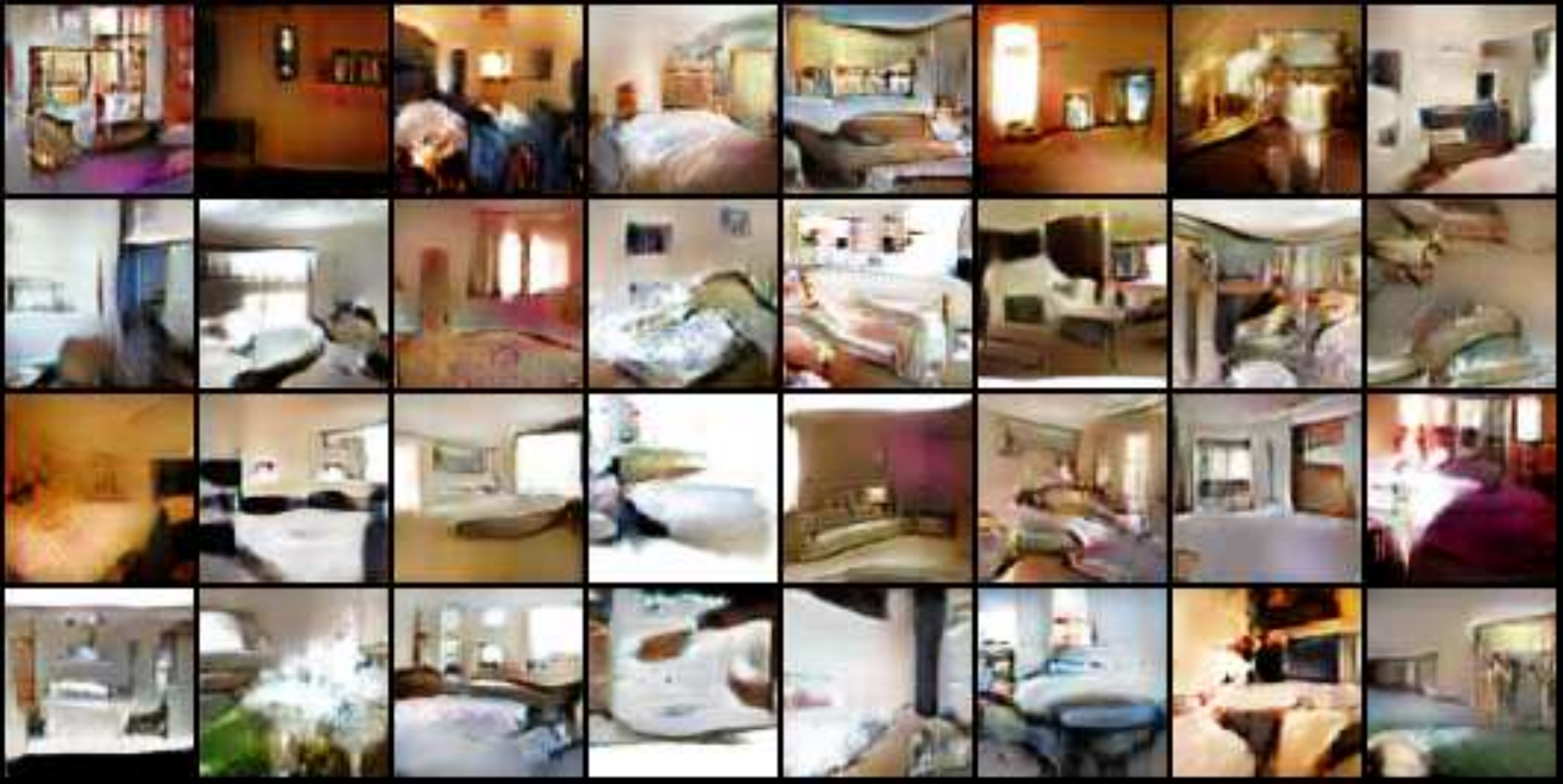}
    &
\widgraph{0.21\textwidth}{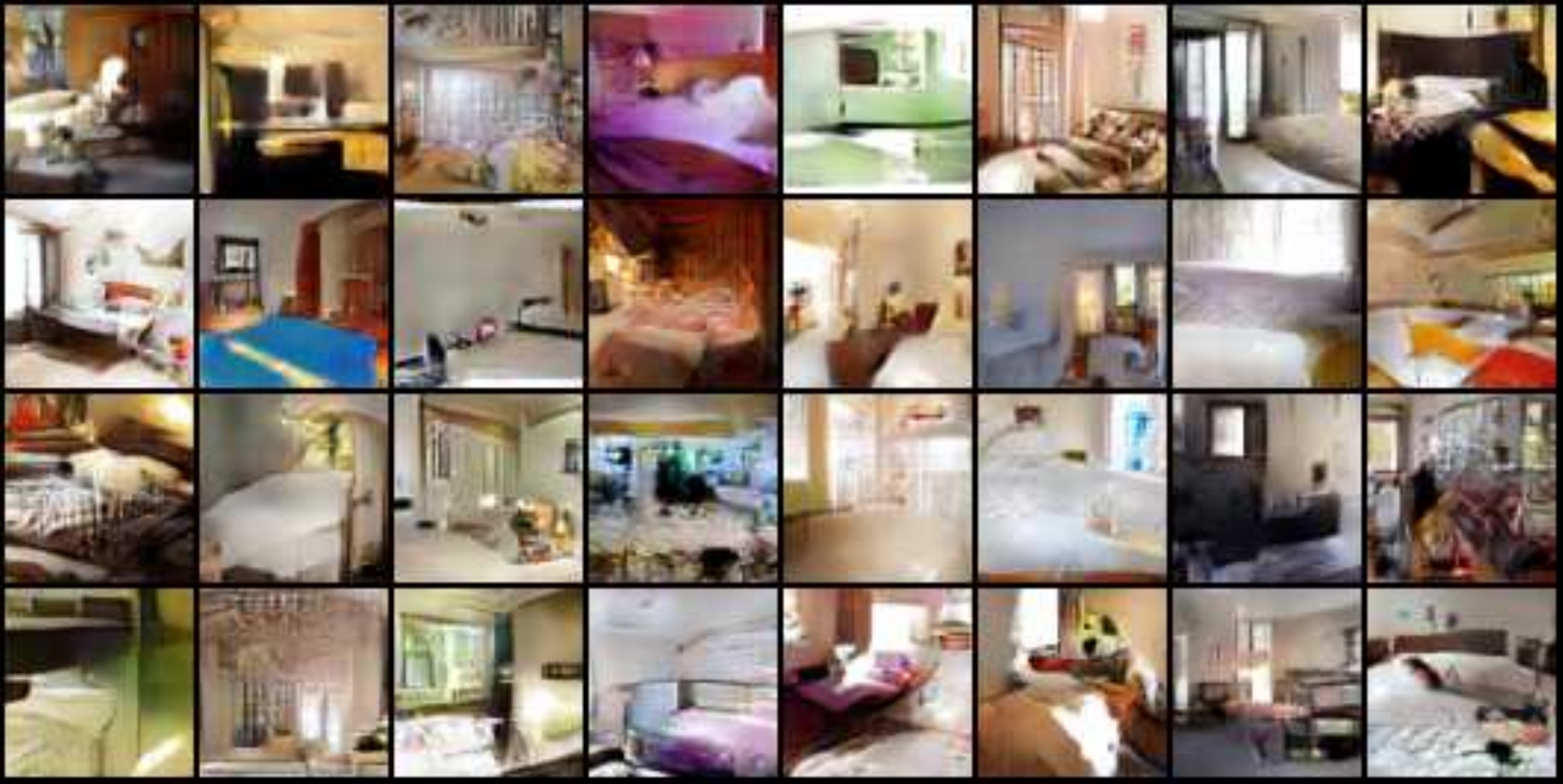}
    
    &
      \widgraph{0.21\textwidth}{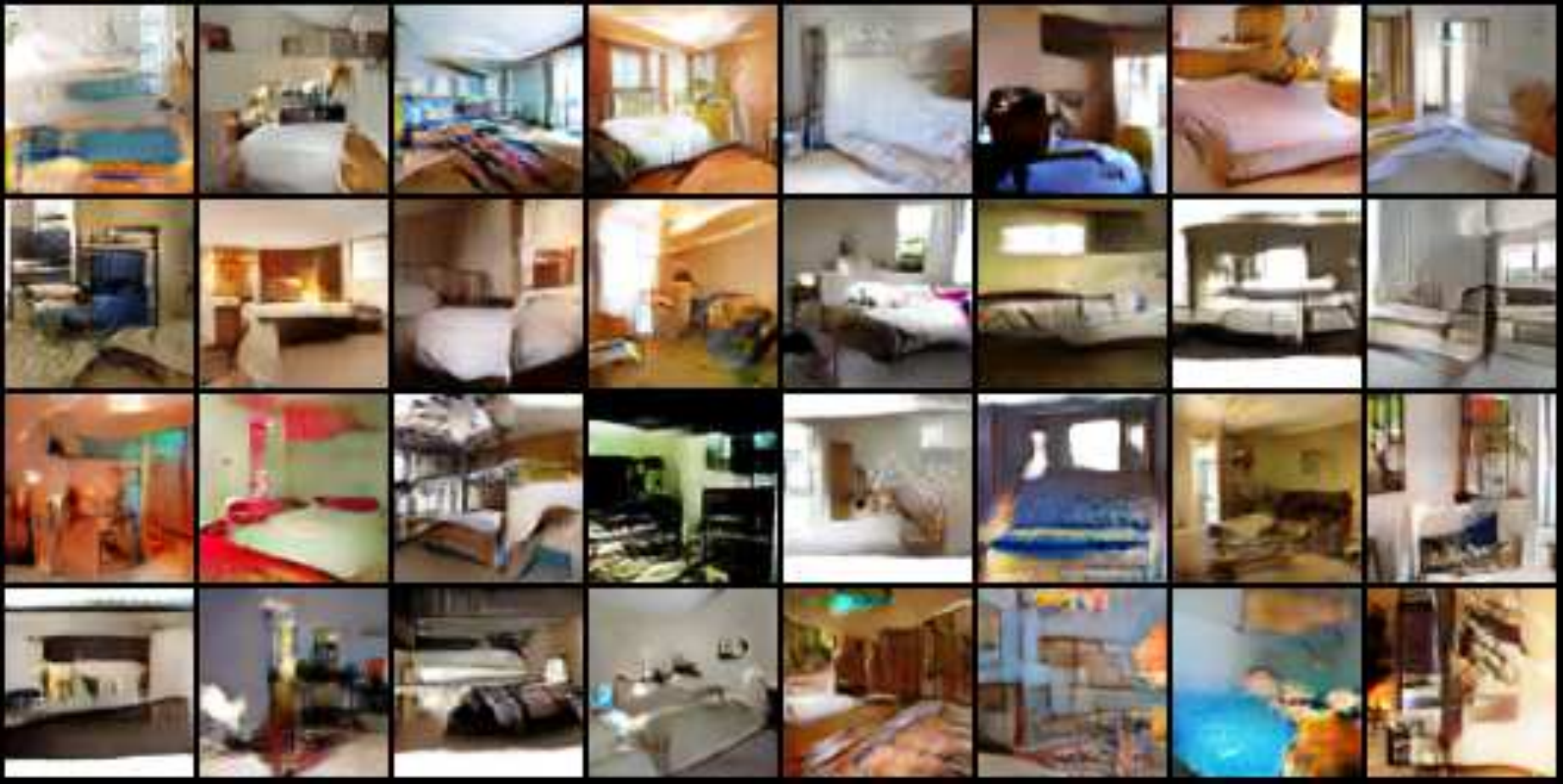}
    &
     \widgraph{0.21\textwidth}{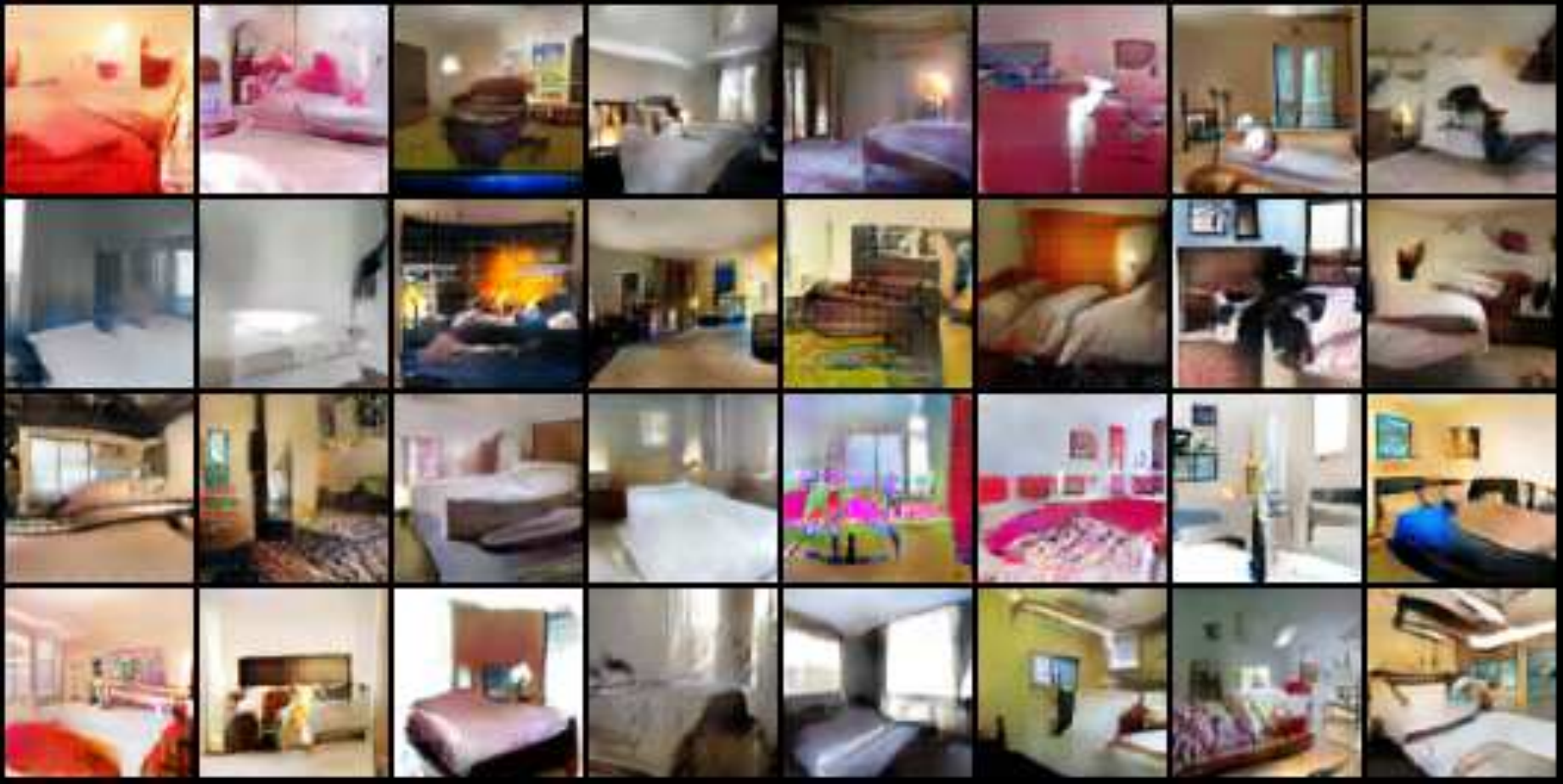}
    \\
      Max-SW&     Max-GSW-NN & GSW n=10000 & DGSW n=10000

  \end{tabular}
   \end{center}
  \caption{ \footnotesize{LSUN  generated images from different generators where $n$ is the number of projections.
    }}
  \label{fig:LSUNgenimages}
\end{figure}

\begin{figure}[!ht]
\begin{center}

  \begin{tabular}{cccc}
    \widgraph{0.21\textwidth}{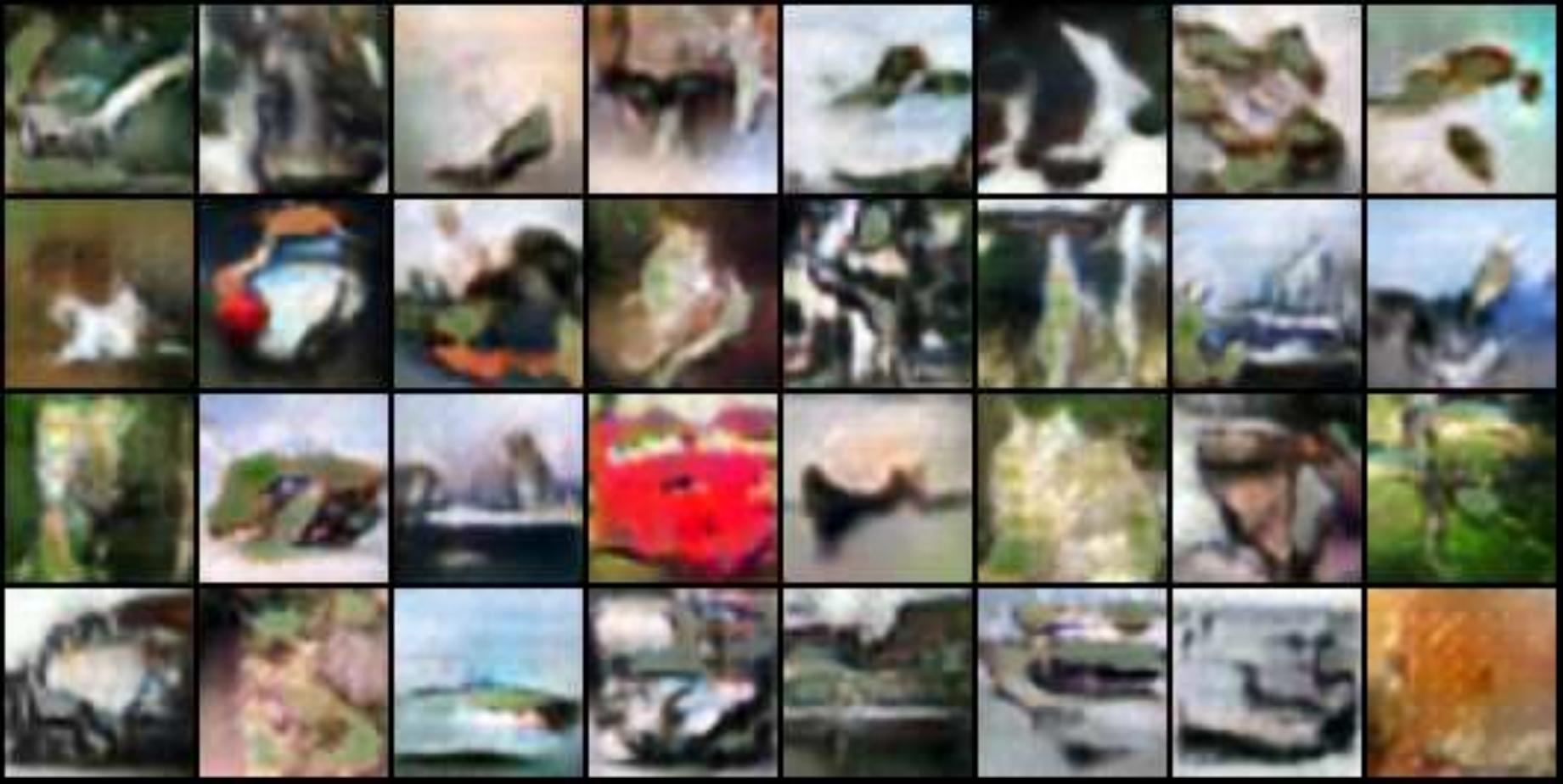}
    &
    \widgraph{0.21\textwidth}{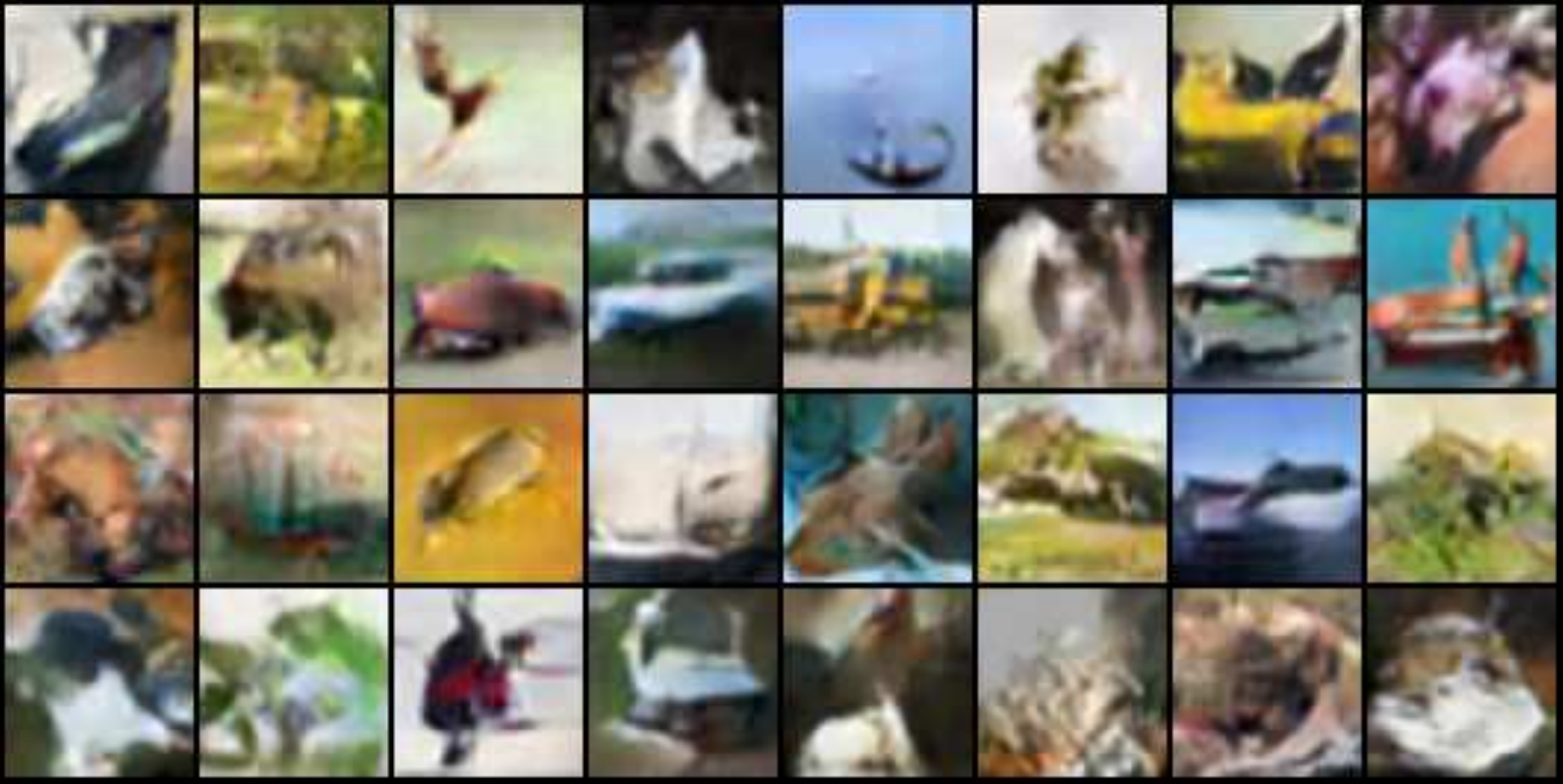}
    
    &
     \widgraph{0.21\textwidth}{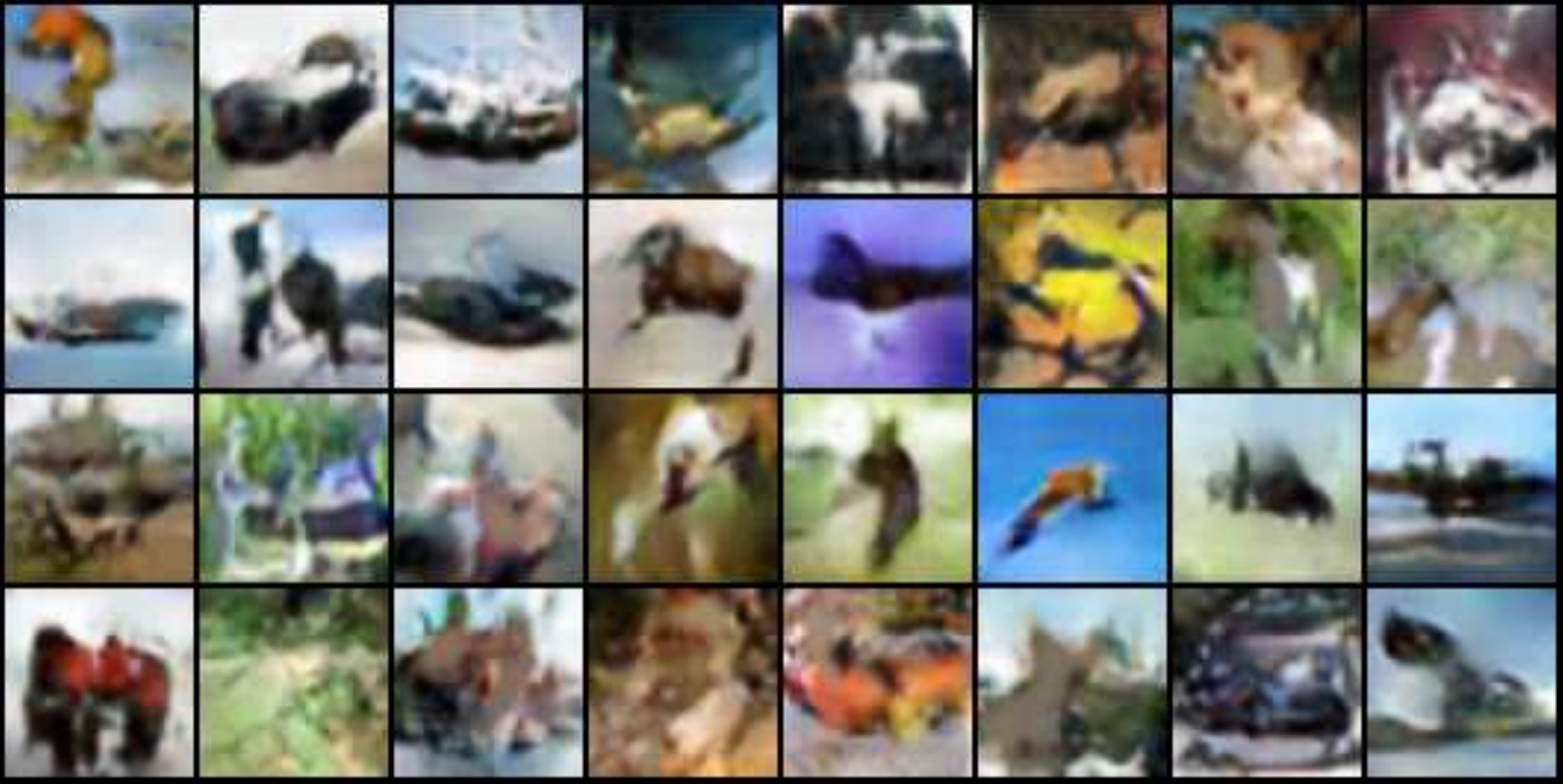}
    &
    \widgraph{0.21\textwidth}{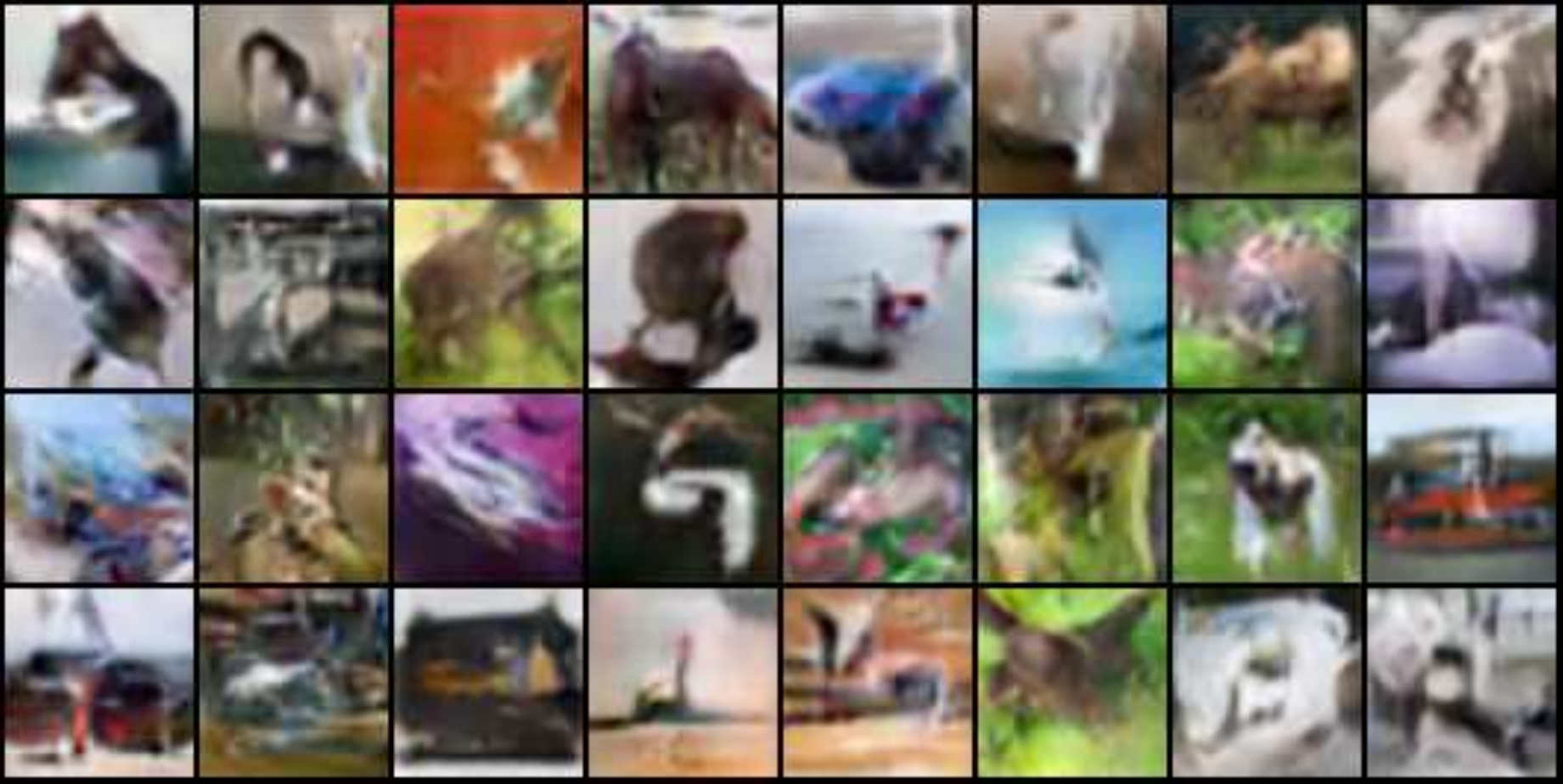}
    \\
      SW $n$=100 &  DSW $n$=100  &SW $n$=10000 &  DSW $n$=10000 
\\
     \widgraph{0.21\textwidth}{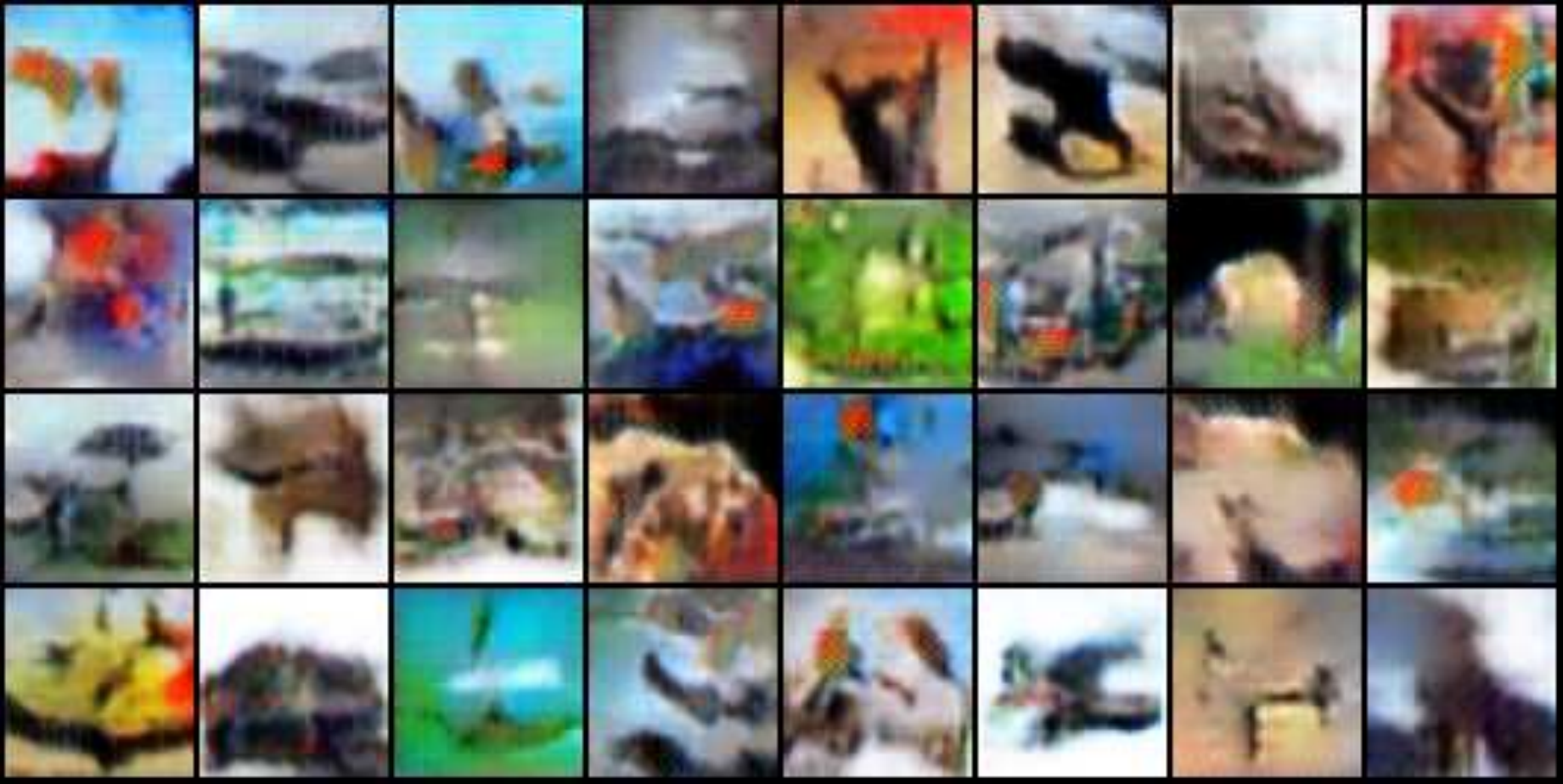}
    &
 \widgraph{0.21\textwidth}{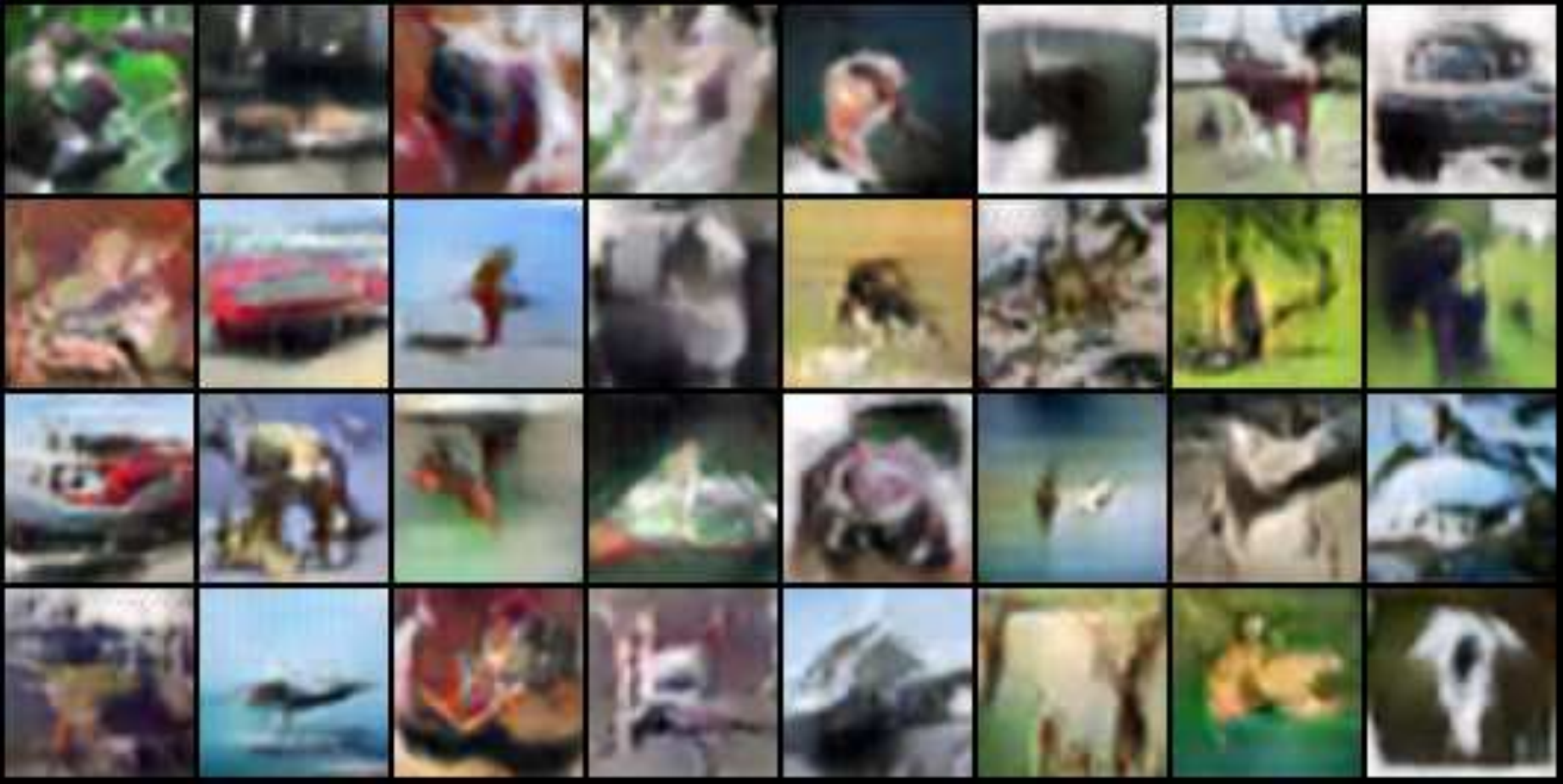}
    
    &
     \widgraph{0.21\textwidth}{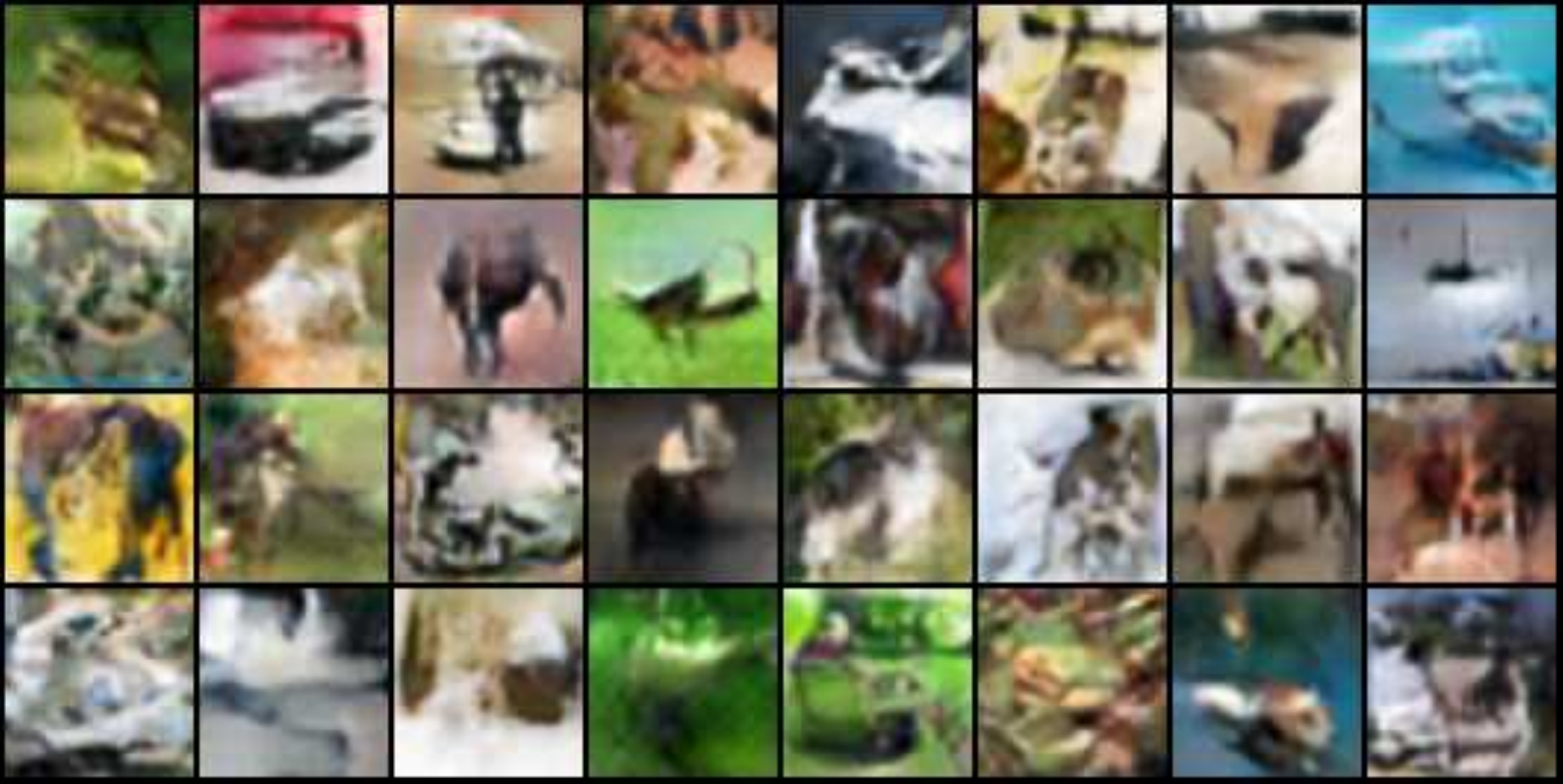}
    &
     \widgraph{0.21\textwidth}{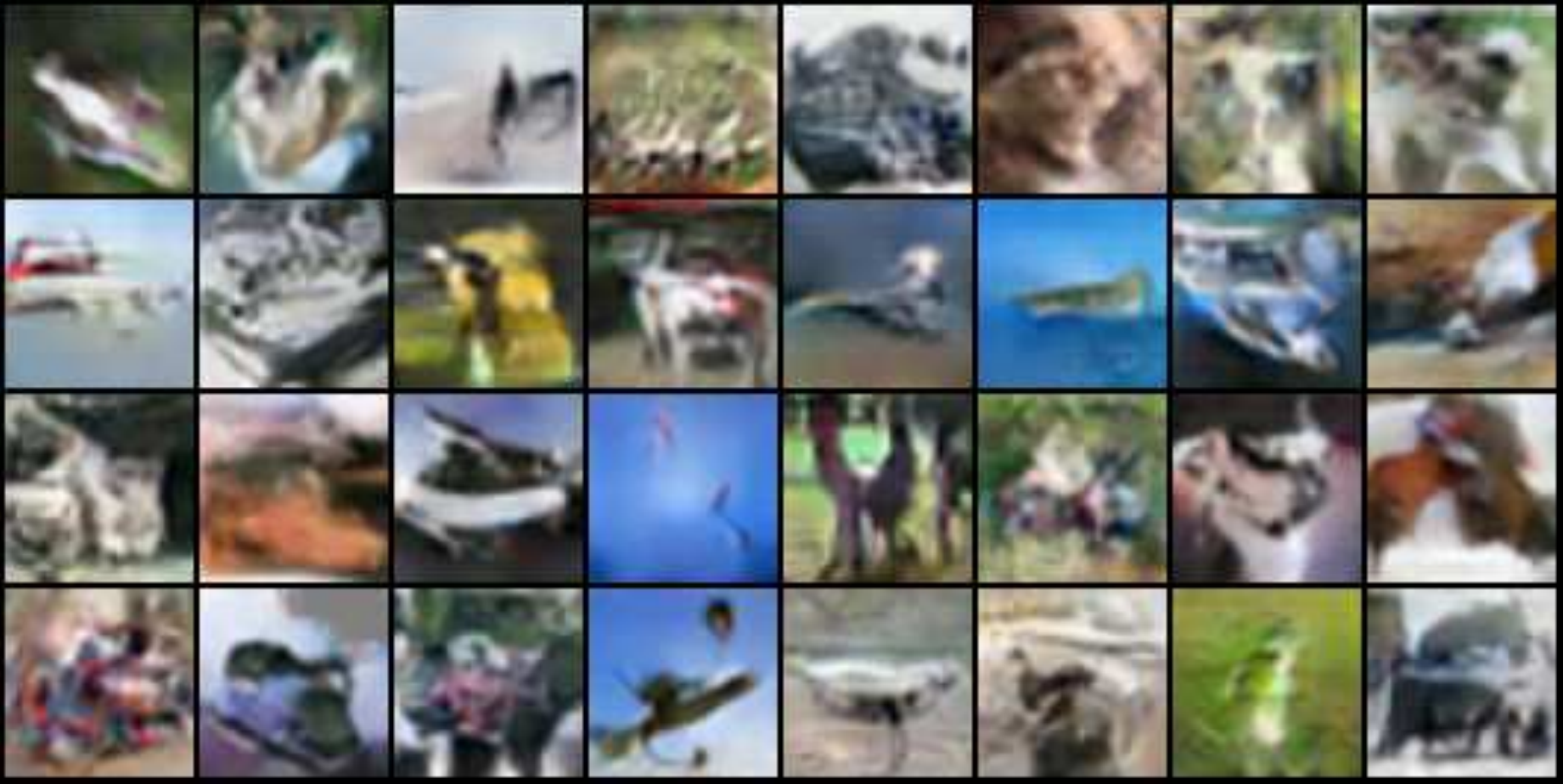}
    \\
      Max-SW&   Max-GSW-NN & GSW $n$=10000 &  DGSW $n$=10000 
 
  \end{tabular}
      
\end{center}
  \caption{ CIFAR10  generated images from different generators, $n$ is the number of projections.
    }
  \label{fig:CIFARgenimages}
\end{figure}

\begin{figure}[!ht]
\begin{center}

  \begin{tabular}{cccc}
    \widgraph{0.21\textwidth}{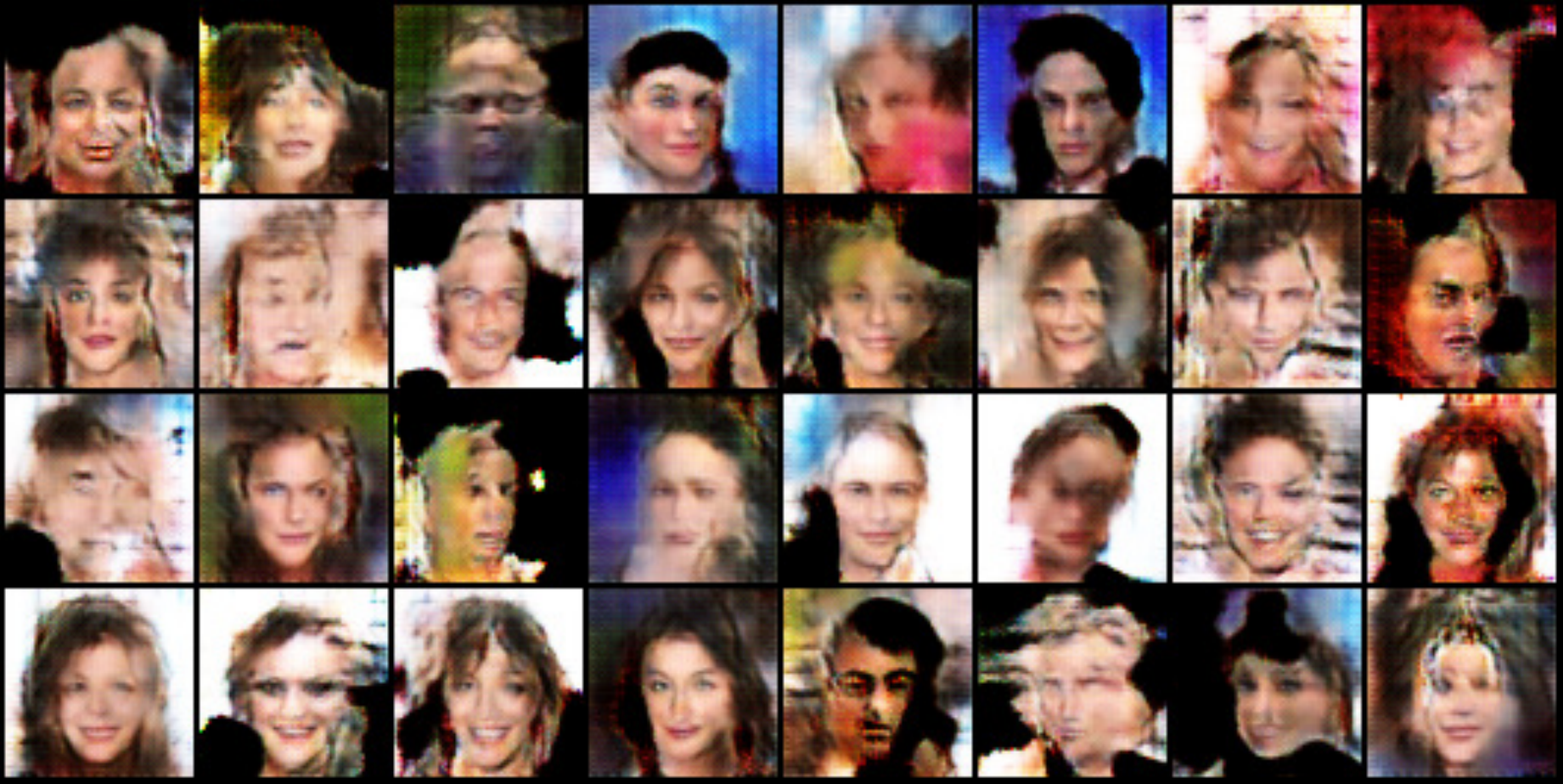}
    &
    \widgraph{0.21\textwidth}{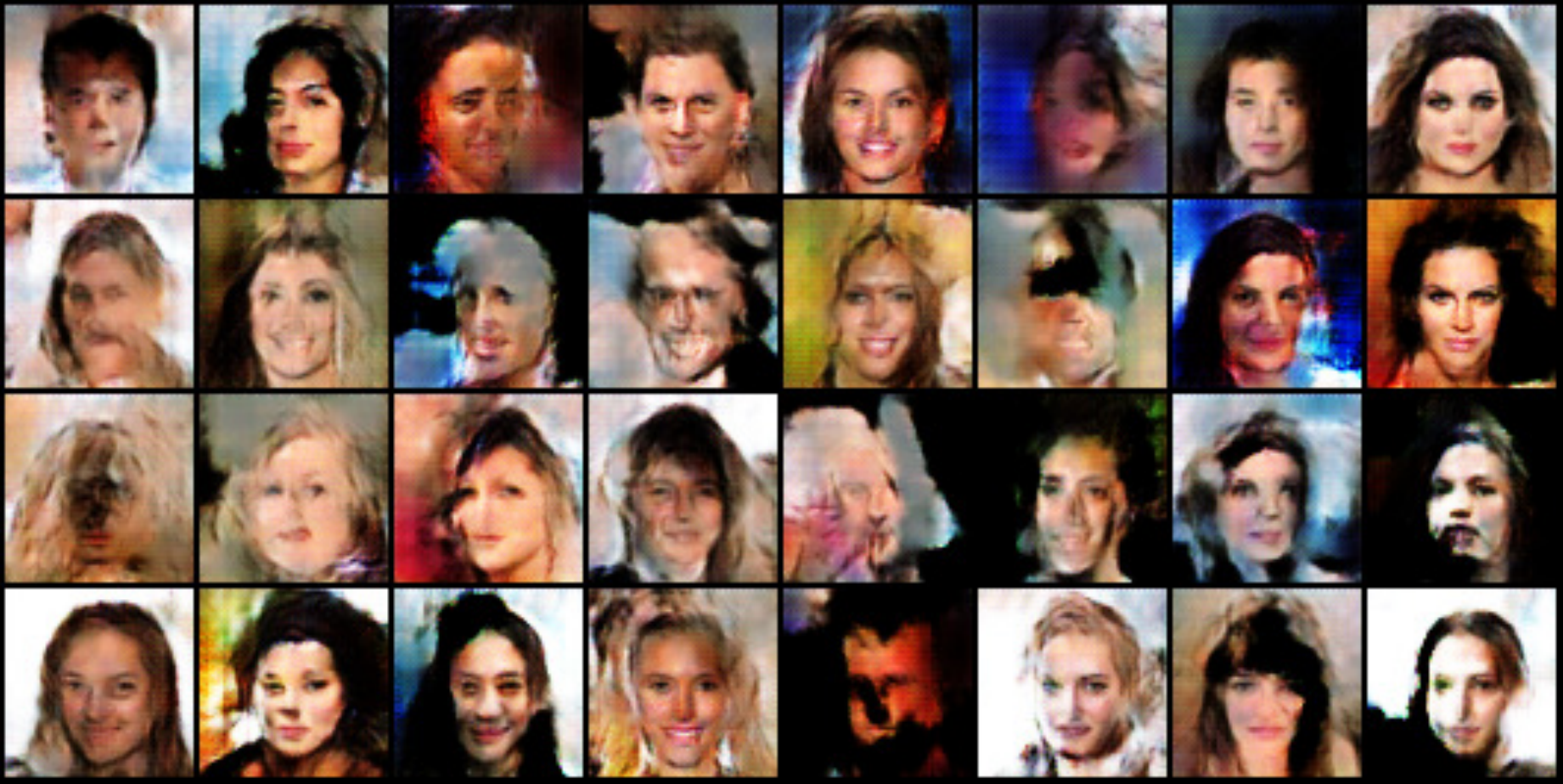}
    
    &
     \widgraph{0.21\textwidth}{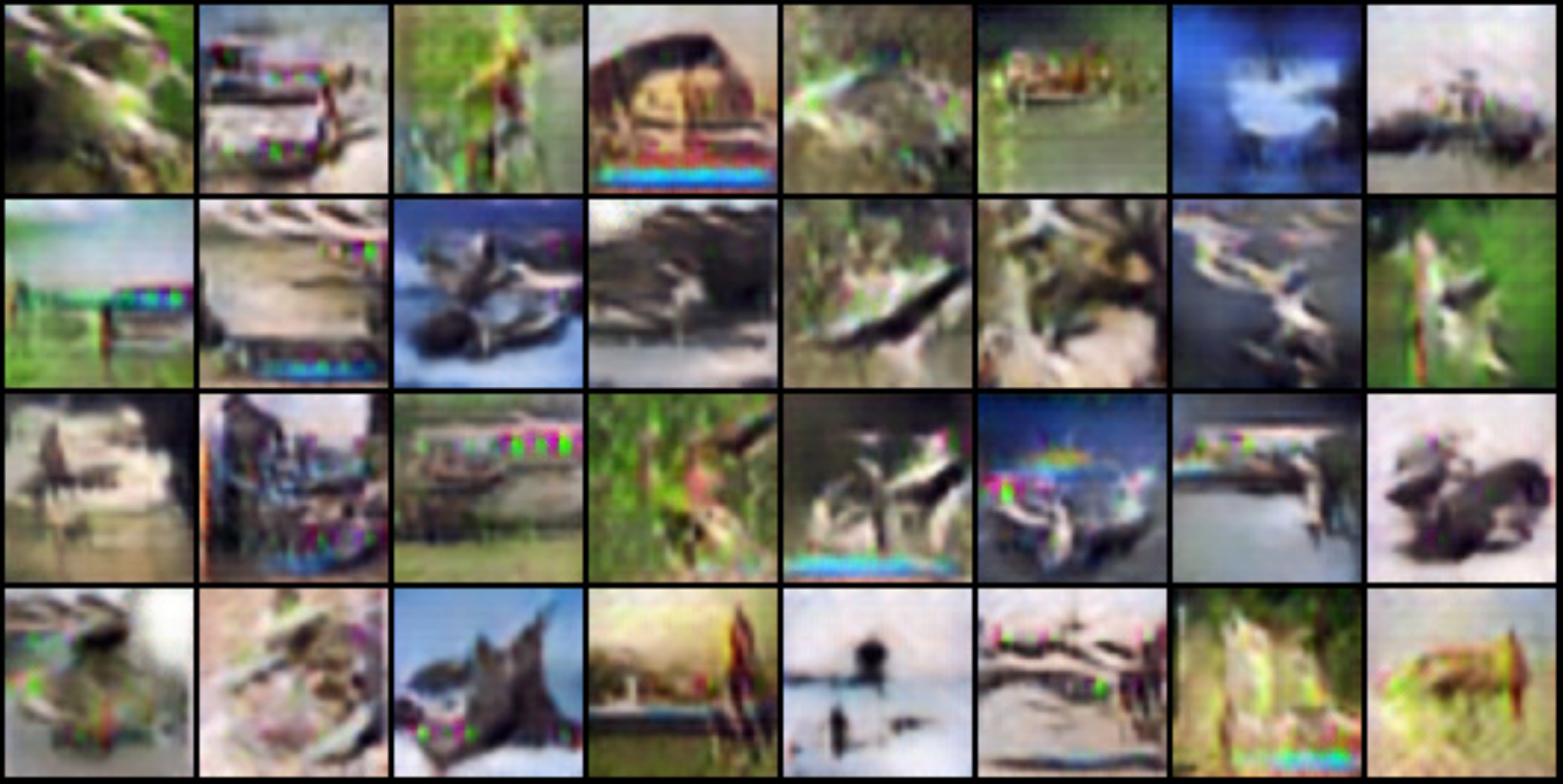}
    &
    \widgraph{0.21\textwidth}{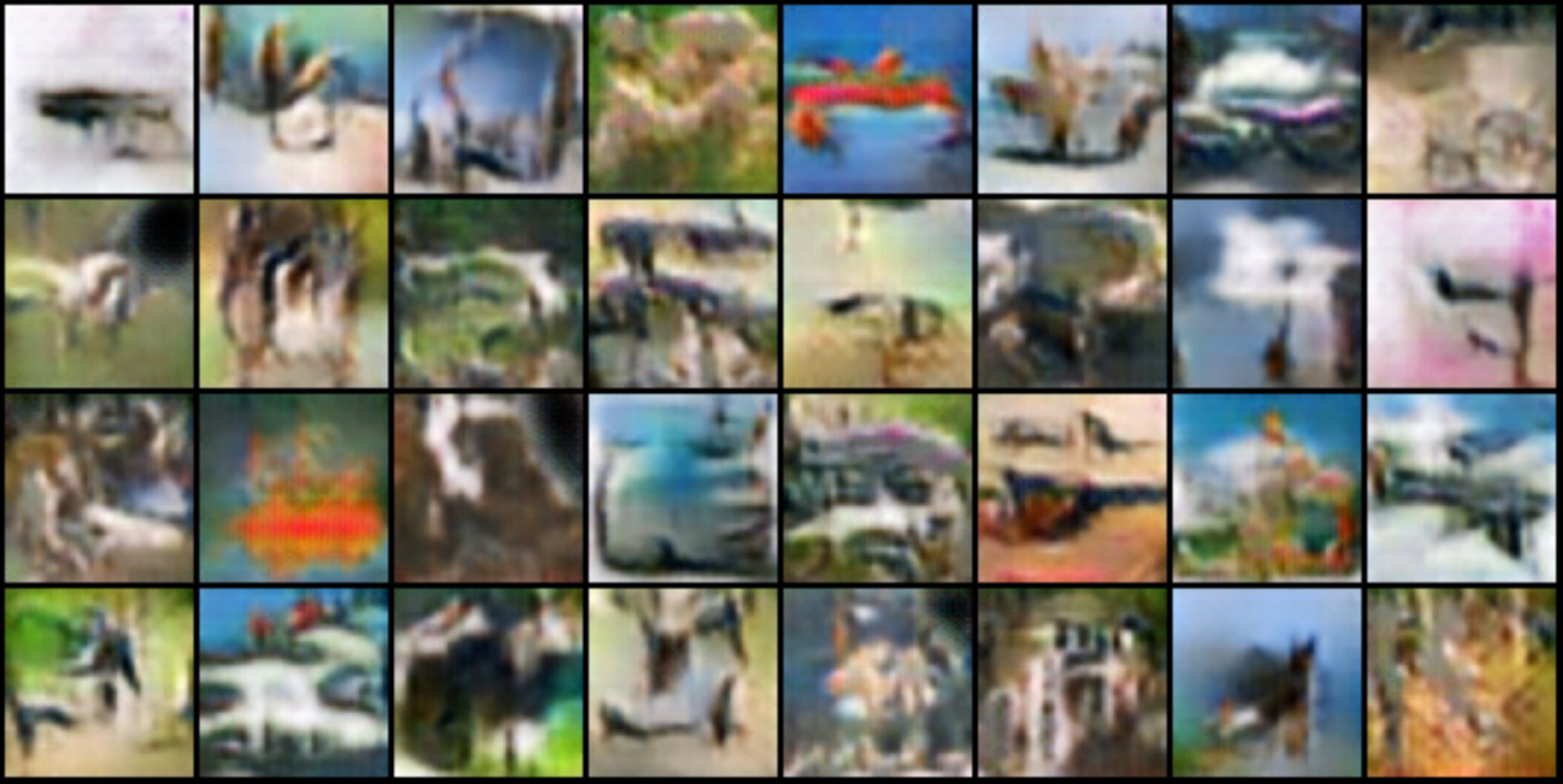}
    \\
      ASW $n$=100  &  ASW $n$=10000  &ASW $n$=100 &  ASW $n$=10000 
 
  \end{tabular}
      
\end{center}
  \caption{ ASW  generated images on CelebA and CIFAR10.
    }
  \label{fig:ASWgenimages}
\end{figure}

\paragraph{Qualitative results:} We show random generated images from trained generators on MNIST, CelebA, CIFAR10 and LSUN datasets in  Figures \ref{fig:MNISTgenimages}-\ref{fig:ASWgenimages}. Overall, we can see that the distributional approaches, i.e., DSW and DGSW distances, help to improve the quality of synthetic images in both linear and non-linear projection cases. 

\textbf{Comparison with the special case of Max-GSW-NN: }
In Max-GSW-NN~\cite{kolouri2019generalized}, one possible choice of neural network defining function is $g(x,\theta) = \langle x,f(\theta) \rangle$ where $f: \mathbb{S}^{d-1} \rightarrow \mathbb{S}^{d-1}$. That function $f$ induces a probability measure on $\sphere^{d-1}$. Hence, optimizing $f$ is equivalent to optimize over the set of probability measures without any constraints, which gives us an effect that is similar to max-SW. In contrast, the function $f$ in our DSW is to find a push-forward probability measure that distributes high probability to informative directions, and this probability measure is regularized to avoid collapsing to a Dirac measure. To support our previous claim, we also do extra experiments on MNIST in Table~\ref{tab:mgswnn_special} to clarify the role of the function $f$ of DSW which makes DSW different from the given special case of Max-GSW-NN.  The result shows that this version of Max-GSW-NN is similar to DSW when $\lambda_C=0$ and both of them have the same performance as Max-SW.
\begin{table}[!h]

\caption{Comparison with the special case of Max-GSW-NN, denoting Max-GSW-NN(*) in the table, that uses the defining function $g(x,\theta) = \langle x,f(\theta) \rangle$ where $f: \mathbb{S}^{d-1} \rightarrow \mathbb{S}^{d-1}$.}
\vskip 0.15in
\begin{center}
\begin{tabular}{llc}
\toprule
Model & $\lambda_C$ &Wasserstein-2\\
\midrule
Max-SW & -& 48.64 \\
Max-GSW-NN (*)&-& 49.21  \\
DSW-10 & 0 & 49.81  \\
DSW-10 & 1 & 38.41  \\
DSW-10 & 10 & \textbf{33.40}  \\
DSW-10 & 100 & 40.08  \\
DSW-10 & 1000 & 46.07 \\
\bottomrule

\end{tabular}
\label{tab:mgswnn_special}

\end{center}
\vskip -0.15in
\end{table}

\subsection{Comparison with Projected Robust Subspace Wasserstein}
\label{subsec:prw}
As shown in \cite{paty2019subspace,lin2020projection}, the main idea of projected robust subspace Wasserstein (PRW) is to find  the  optimal subspace (dimension $\geq$ 2) such that the Wasserstein-2 distance between two projected measures is maximal.

We first recall the definition of PRW in~\cite{paty2019subspace}. 
\begin{definition}
Let $\mathbb{V}_k (\mathbb{R}^d)= \{U \in \mathbb{R}^{d \times k}: U^\top U =I_k\}$ and $\mu, \nu \in \mathcal{P}(\mathbb{R}^d)$. Then, the projection robust 2-Wasserstein (PRW) distance between $\mu, \nu$ is given by: 
\begin{align}
\text{PRW}_k(\mu,\nu) = \max_{U \in \mathbb{V}_k (\mathbb{R}^d)} W_2 (U^\top \sharp \mu,U^\top \sharp \nu).
\end{align}
\end{definition}
Since the projected dimension is bigger than 1, PRW does not have close-form solution on the projected space.



 

\begin{figure}[!ht]
\begin{center}
  \begin{tabular}{cccc}
   
    \widgraph{0.21\textwidth}{MNISTfolder_DSWD_10_epoch199-eps-converted-to.pdf}
    &
    \widgraph{0.21\textwidth}{MNISTfolder_DSWD_1000_epoch199-eps-converted-to.pdf}
   &
     \widgraph{0.21\textwidth}{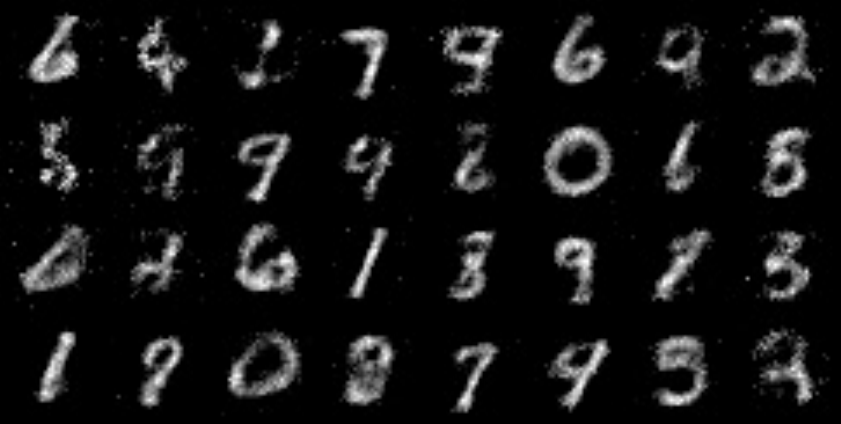}
&
     \widgraph{0.21\textwidth}{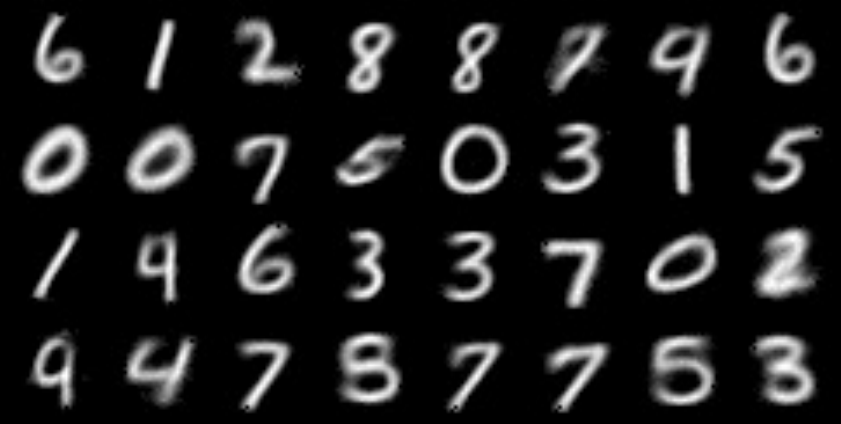}\\
     DSW $n$=10 &  DSW $n$=1000  &PRW $k$=5 &  PRW $k$=50 

  \end{tabular}
      
\end{center}
  \caption{ MNIST  generated images from generators of DSW and PRW. Here, $n$ is the number of projections of DSW and $k$ is the projected dimension of PRW.
    }
  \label{fig:PRWimages}
\end{figure}

\textbf{Experiments on generative model: } We continue to use the MEDE framework on the same settings as previous experiments to compare DSW and PRW. To solve the optimization on Stiefel manifold in PRW, we use the "geoopt" library \cite{geoopt2020kochurov}. We use one gradient step to solve the optimization problem of both DSW and PRW per one generator update. The experiments are carried out with both DSW and PRW on MNIST dataset. The number of projections of DSW takes value 10 and 1000 and the dimension of the subspace of PRW belongs to the set  $\{2,5,10,50 \}$. We report the Wasserstein-2 results and the computational time in Table~\ref{table:PRW_W2} and the generated images in Figure~\ref{fig:PRWimages}. 
\begin{table}[!h]

\caption{Empirical Wasserstein-2 score and computation speed per minibatch on MNIST dataset. }
\vskip 0.15in
\begin{center}
\begin{tabular}{llll}
\toprule
Model & $k$-dimension & Wasserstein-2 & Second/Minibatch\\
\midrule
DSW-10&-  &34.4& \textbf{0.003}\\
DSW-1000&-  &33.11& 0.018 \\
PRW &2&65.39 & 0.086 \\
PRW &5& 35.99& 0.092\\
PRW &10& 26.57 & 0.11\\
PRW &50& \textbf{24.38} & 0.12\\
\bottomrule

\end{tabular}
\label{table:PRW_W2}

\end{center}
\vskip -0.15in
\end{table}

According to Table \ref{table:PRW_W2}, DSW with 10 projections obtains a better Wasserstein-2 score than the PRW with 5-dimensional subspace, while its corresponding computational time is 30 times faster that of PRW. When PRW searches for the 50-dimensional subspace, the  Wasserstein-2 score only improves $32.25\%$ meanwhile the computational time increases by 10 times. 


Next, we show some generated images from both DSW and PRW. We observe that these images are consistent with Wasserstein-2 score in the previous experiments.
      
\subsection{Joint contrastive inference}
\label{subsec:exp_joint_contrastive}
\begin{figure}[!t]
  \begin{tabular}{cccc}

    \widgraph{0.22\textwidth}{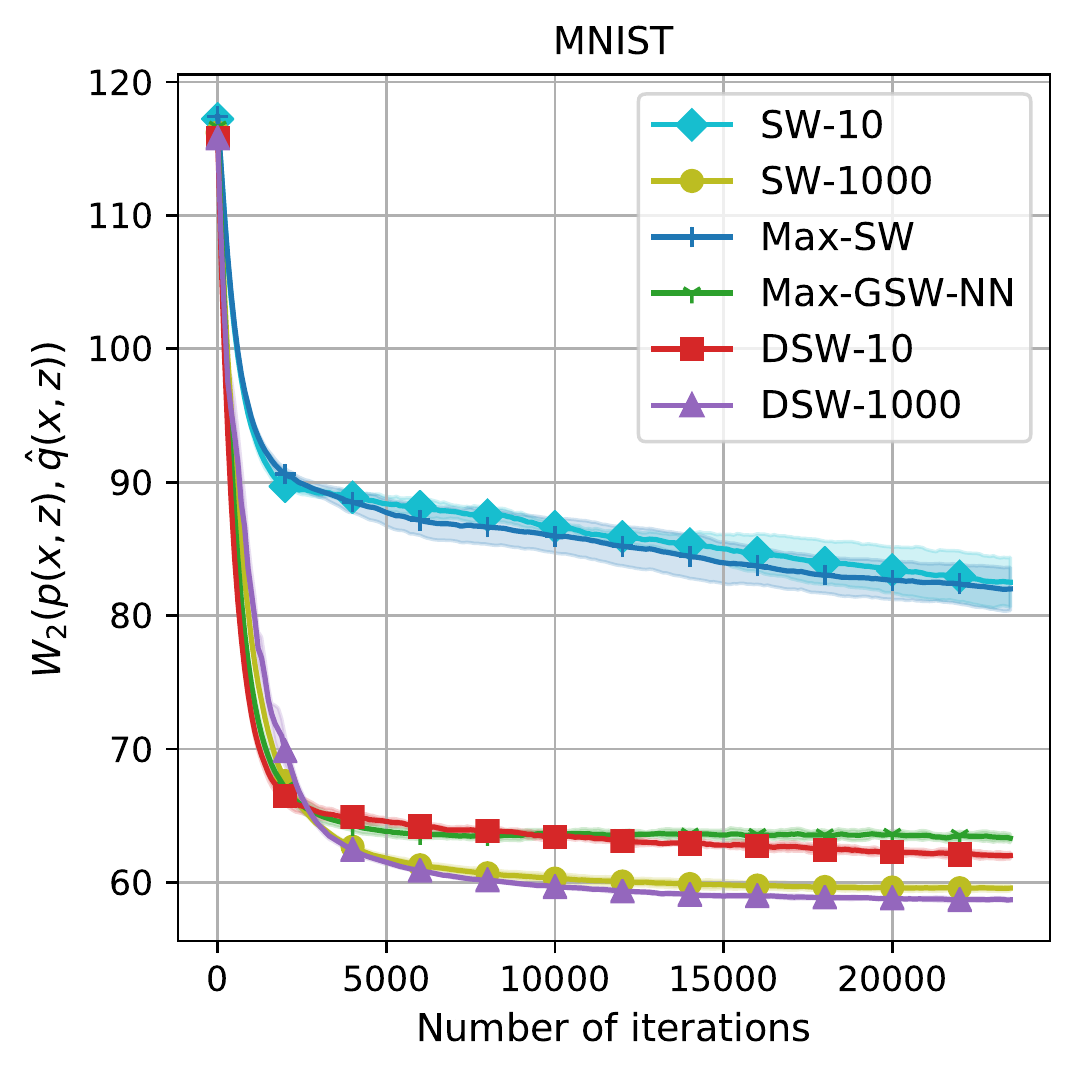} & \widgraph{0.22\textwidth}{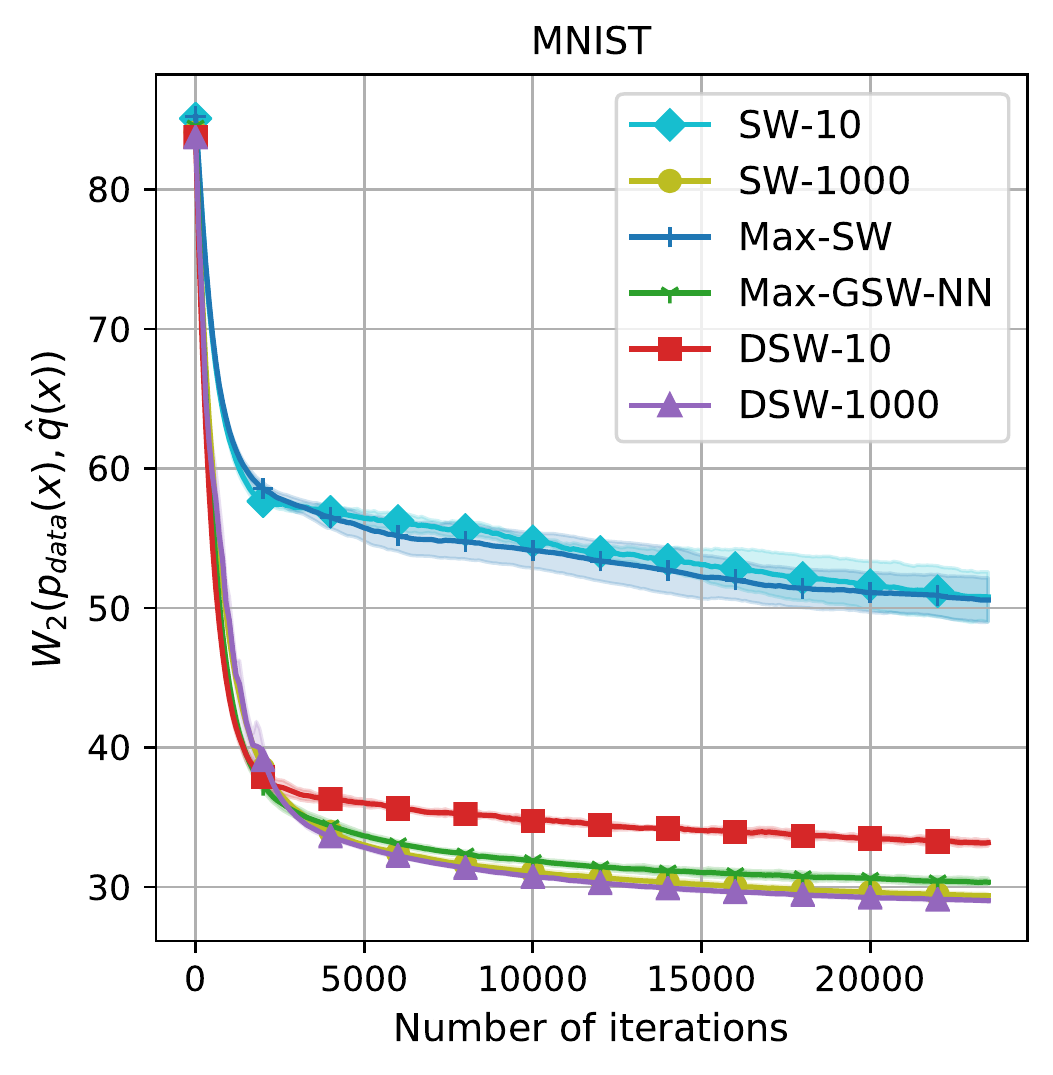}&
    \widgraph{0.22\textwidth}{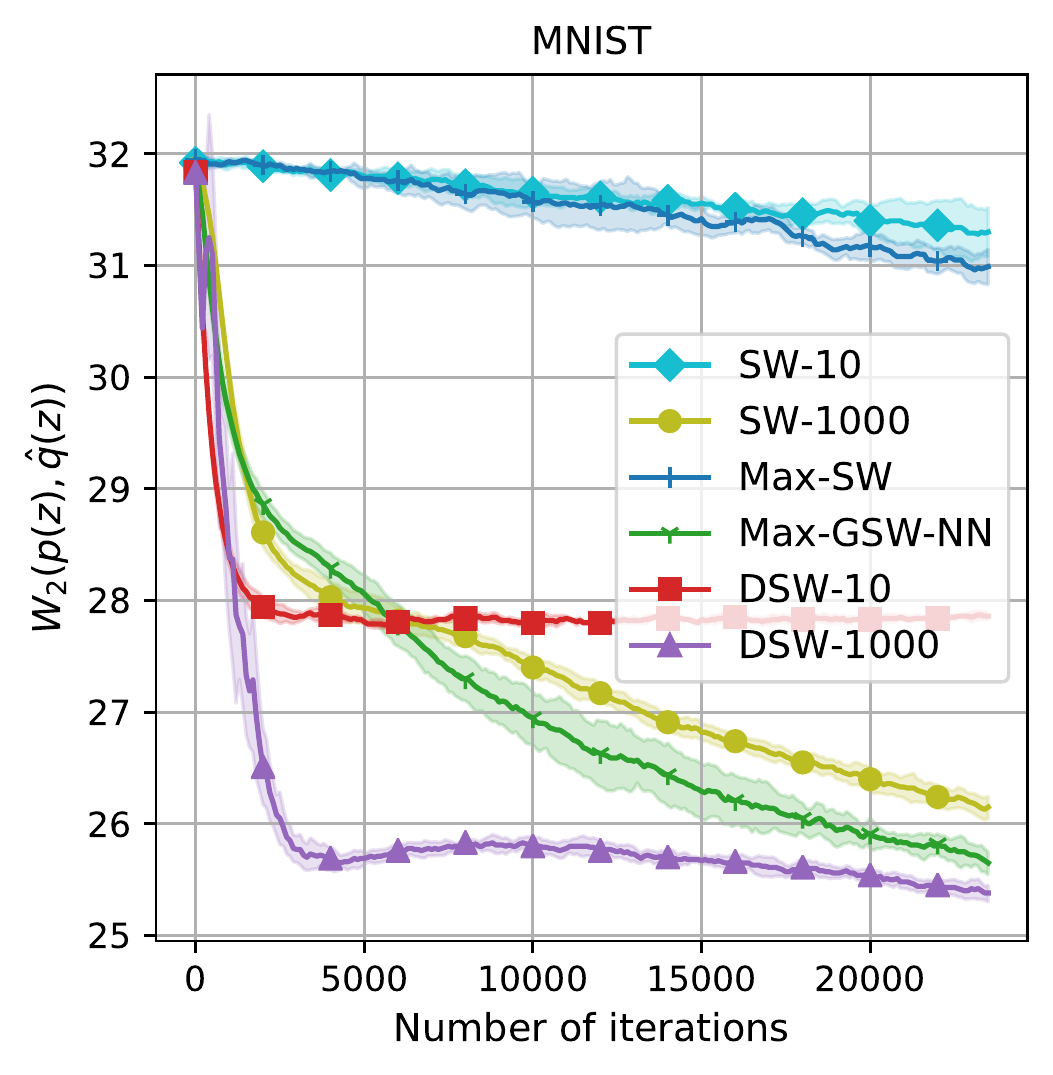} & \widgraph{0.22\textwidth}{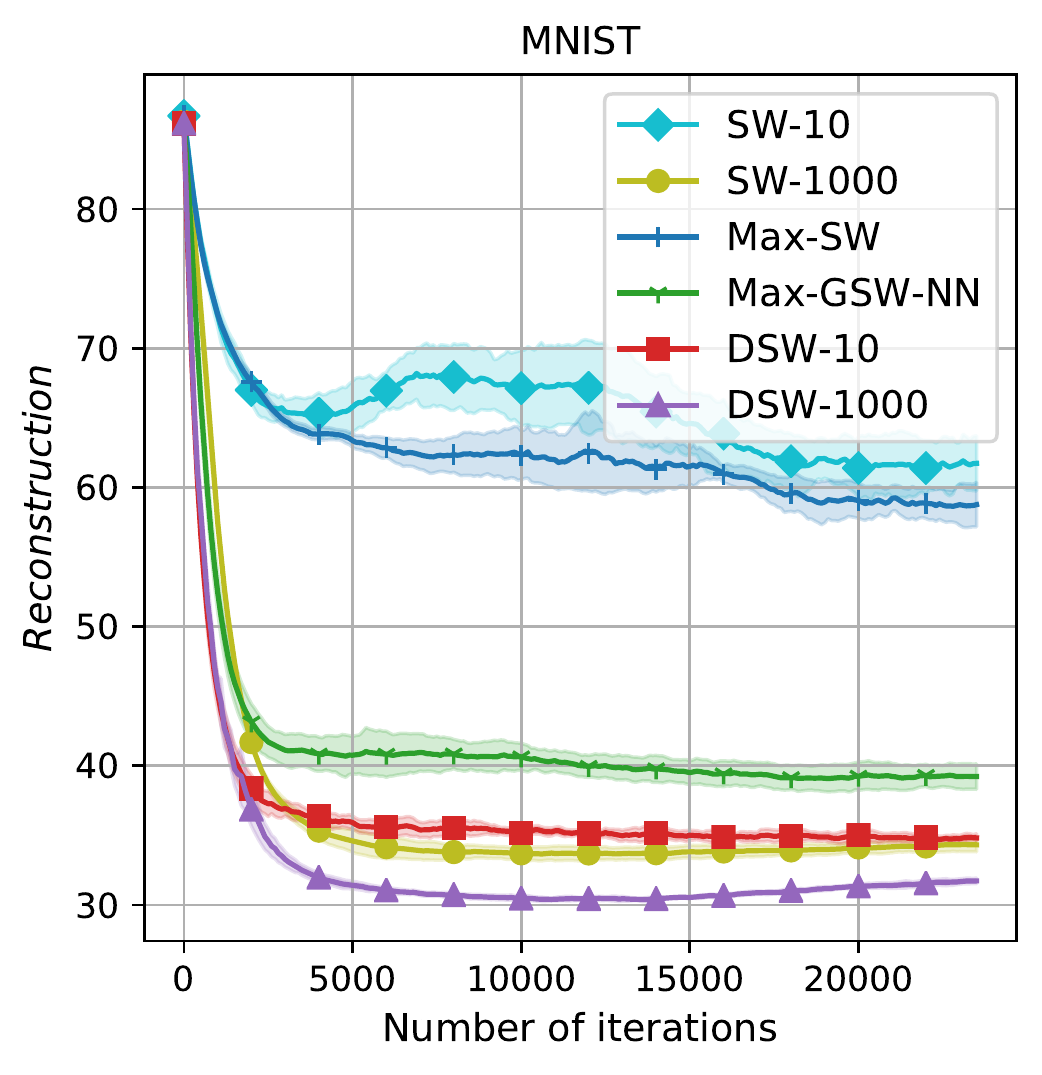}  \\
    (a)&(b)&(c)&(d)
  \end{tabular}
  \caption{
  \footnotesize{ Joint inference model comparisons among DSW, SW, Max-SW, and Max-GSW-NN.}
    }
  \label{fig:MNISTjoint}
  \vspace{-0.5 em}
\end{figure}

\begin{figure}[!ht]
  \begin{tabular}{cccc}
    \widgraph{0.22\textwidth}{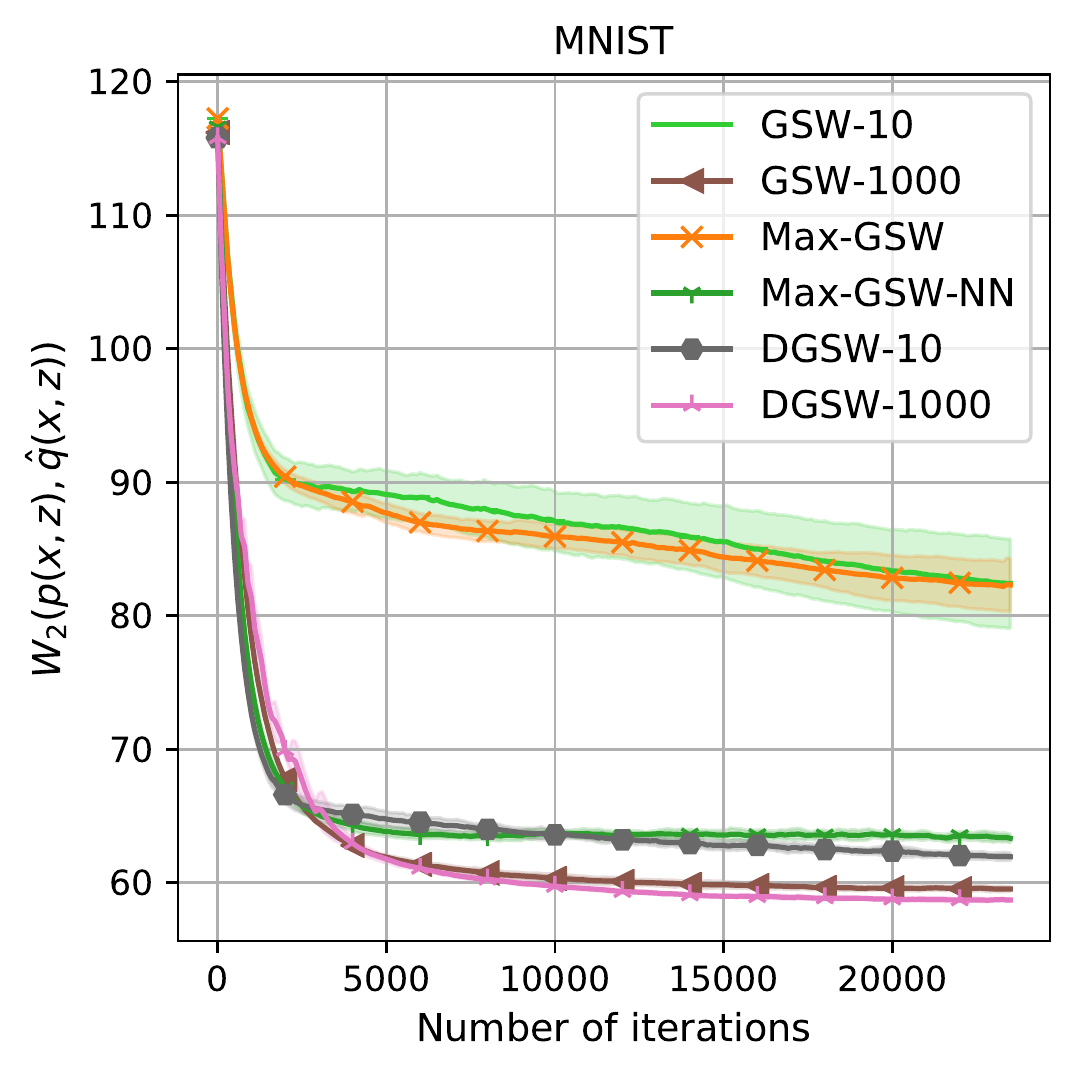} &\widgraph{0.22\textwidth}{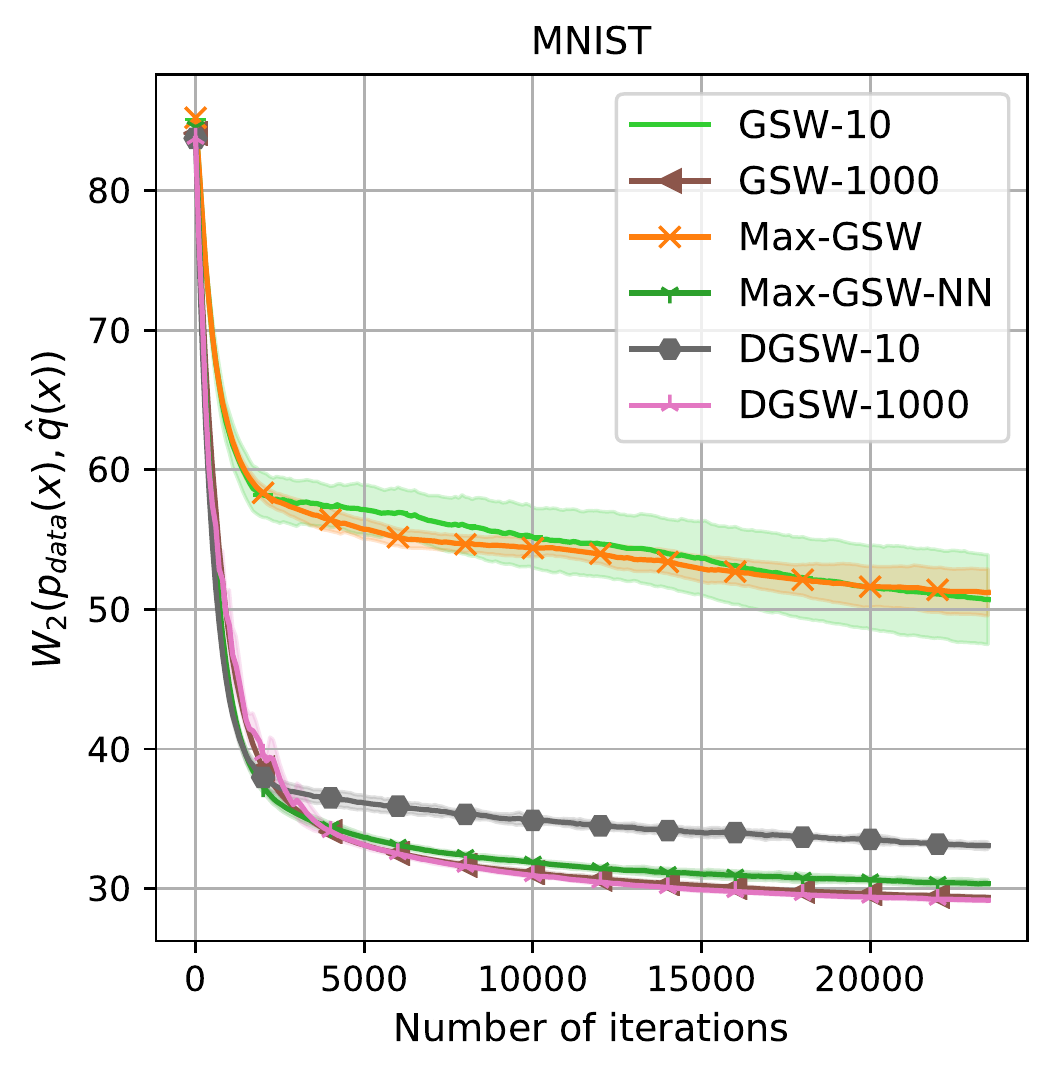}&\widgraph{0.22\textwidth}{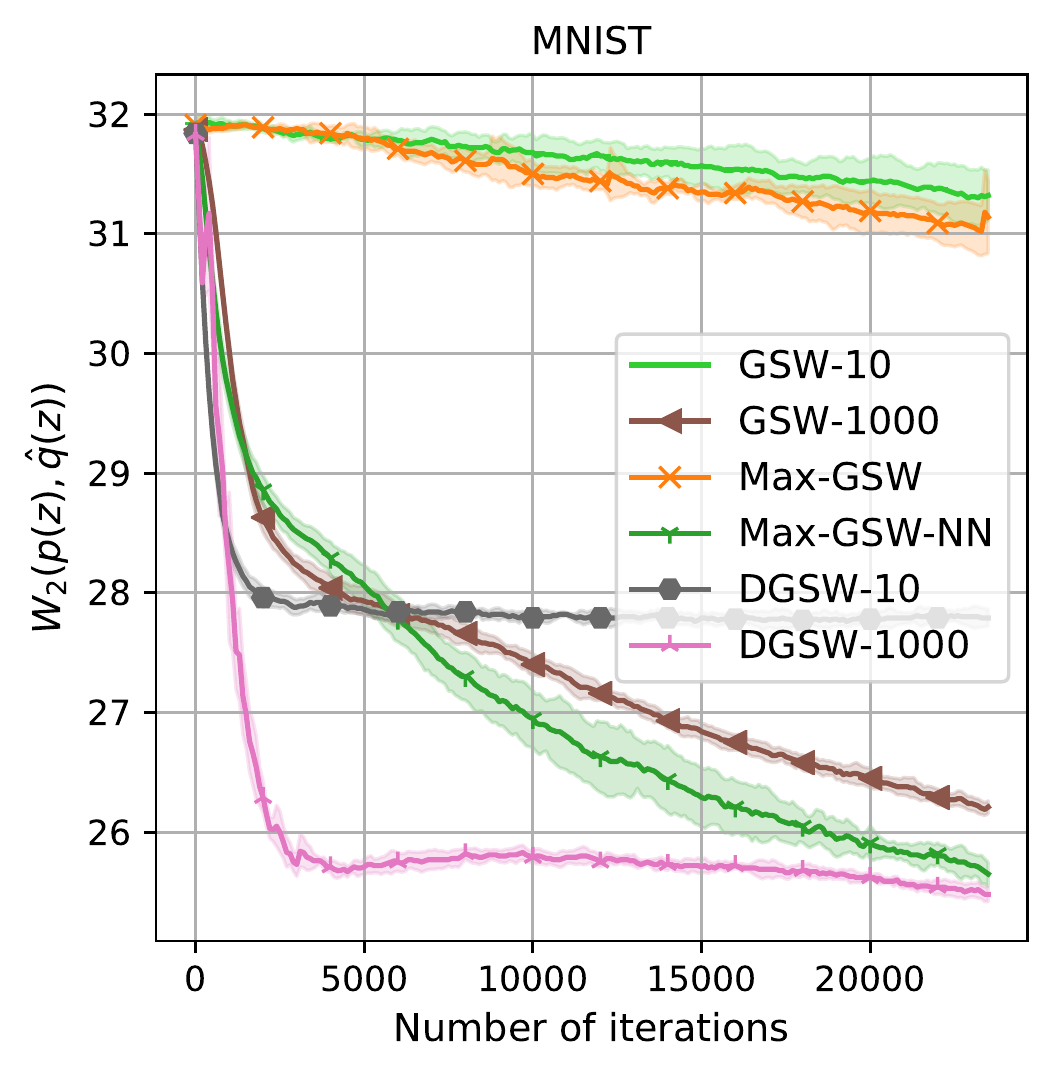} &\widgraph{0.22\textwidth}{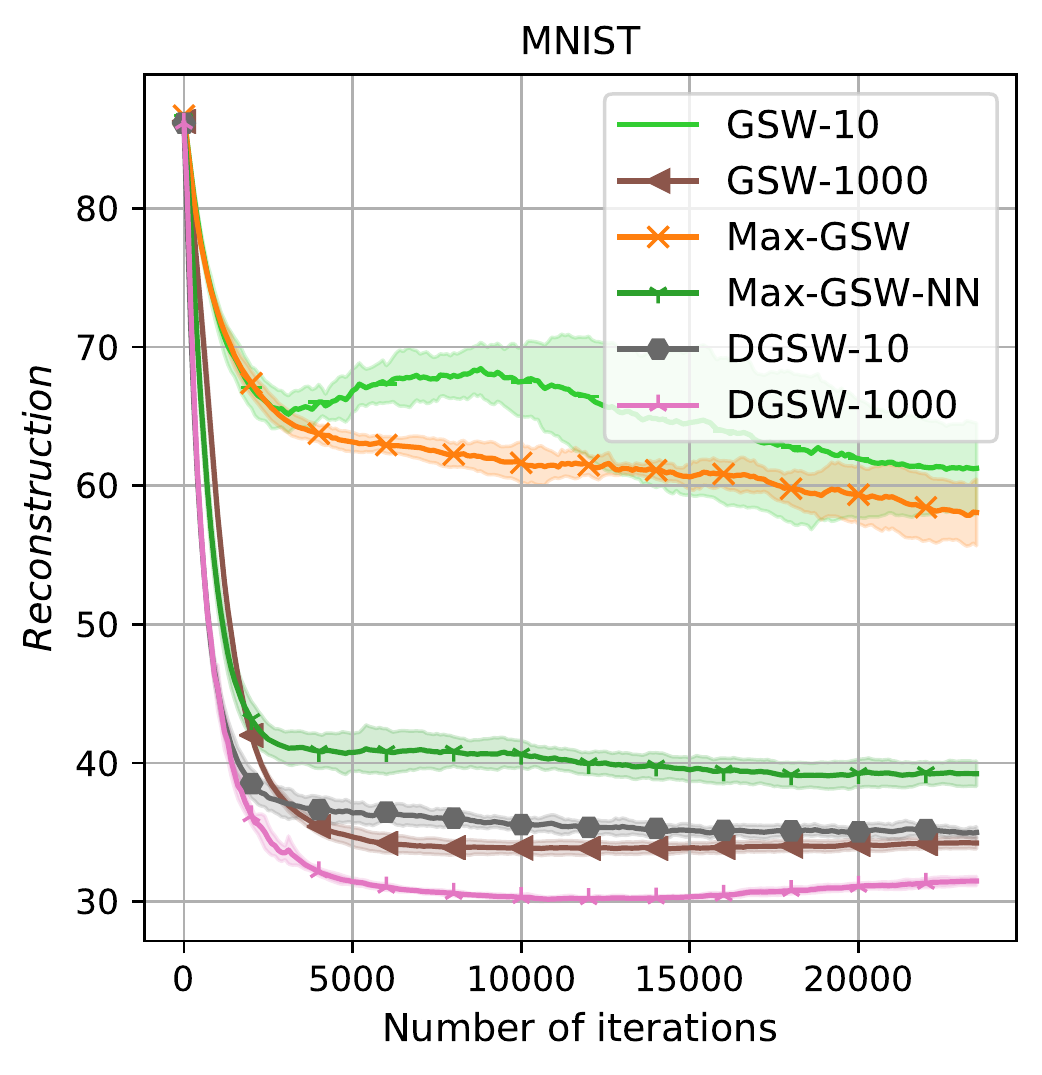} 
  \end{tabular}
  \caption{
  Joint inference model comparisons between non-linear sliced distances. 
    }
  \label{fig:addMNISTjoint}
\end{figure}

We test the performance of our distances in training encoder-generator models on MNIST using joint contrastive inference (JCI). In JCI, the joint generative latent-observed distribution $p_\theta(z,x)=p(z)p_\theta(x|z)$ is matched with the empirical joint latent-observed distribution $\hat{q}_\phi(z,x)=p_{\text{data}}(x)q_\phi(z|x)$ by minimizing a chosen distance (see Appendix~\ref{sec:application} for a description of these models). 
We evaluate how close the two joint latent-observed distributions $p_\theta(z,x)$ and $\hat{q}_\phi(z,x)$ are, how close their corresponding marginals are (in Wasserstein-2 distance) and the ability of the encoder-generator in reconstructing images. These metrics are shown in Figures~\ref{fig:MNISTjoint}(a)-(d). The results show that DSW achieves better performance than SW using the same number of projections, with DSW-1000 achieves the best performance among all the other baselines in all metrics.
We give experiments to compare DGSW with other non-linear sliced-Wasserstein distances in the joint contrastive inference task in Figure \ref{fig:addMNISTjoint}. We observed the same behavior as the linear case, the distributional version of GSW using circular function achieves better performance than the other non-linear sliced-based distances.

In order to illustrate the ability to reconstruct images of joint inference models, we show reconstructed images from the MNIST dataset. With 10 projections, SW and GSW were not able to recreate the digits; however, DSW and DGSW can recreate the digits quite correctly. Furthermore, Max-GSW-NN performs well in this task and is much better than Max-SW and Max-GSW. When having enough number of projections (for example 1000), it is very hard to compare SW, GSW, DSW, and DGSW by eyes. Nevertheless, according to reconstruction error plots in Figures \ref{fig:MNISTjoint} and \ref{fig:addMNISTjoint},  DSW, and DGSW distances are still better than the other sliced-based distances.
\begin{figure}[!ht]
  \begin{tabular}{l c}
     Data &\widgraph{0.8\textwidth}{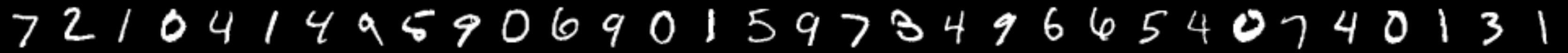}
     \\
     Max-SW  &\widgraph{0.8\textwidth}{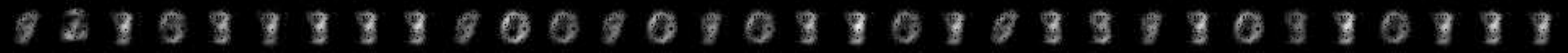}
     \\
     Max-GSW  &\widgraph{0.8\textwidth}{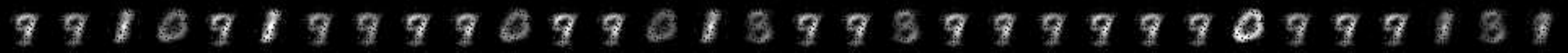}
     \\
    Max-GSW-NN &\widgraph{0.8\textwidth}{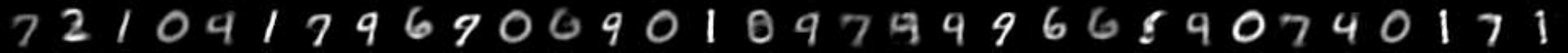}
     \\
     SW $n=10$ &\widgraph{0.8\textwidth}{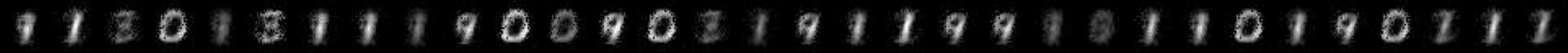}
     \\
     DSW $n=10$ &\widgraph{0.8\textwidth}{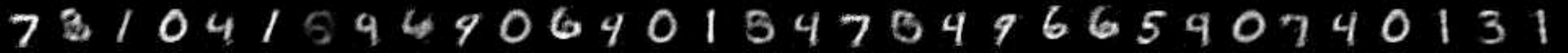}
     \\
     GSW $n=10$ &\widgraph{0.8\textwidth}{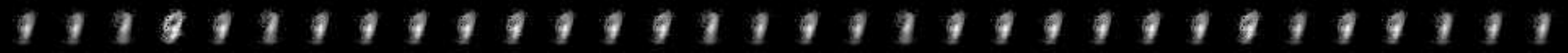}
     \\
     DGSW $n=10$ &\widgraph{0.8\textwidth}{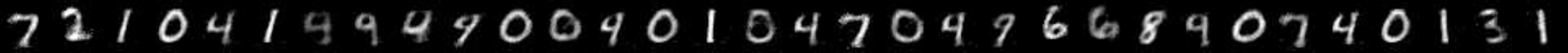}
     \\
     SW $n=1000$ &\widgraph{0.8\textwidth}{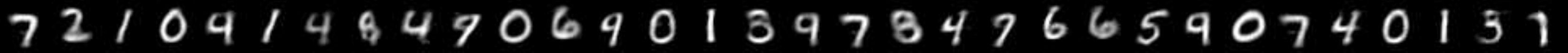}
     \\
     DSW $n=1000$ &\widgraph{0.8\textwidth}{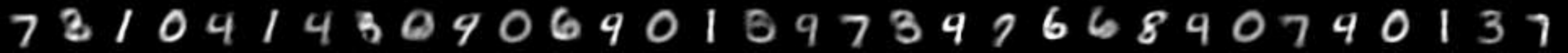}
     \\
     GSW $n=1000$ &\widgraph{0.8\textwidth}{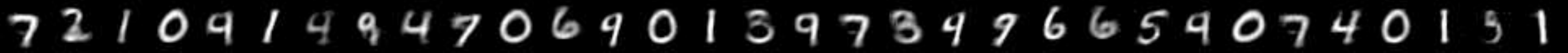}
     \\
     DGSW $n=1000$ &\widgraph{0.8\textwidth}{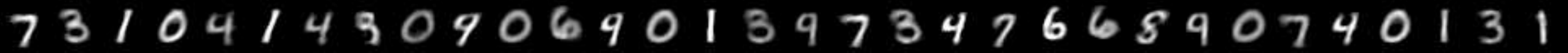}

  \end{tabular}
  \caption{MNIST dataset reconstruction images ($n$ is the number of projections).
    }
  \label{fig:MNISTrecontruction}
\end{figure}

\section{Experiment Settings}
\label{sec:setting}
We use a single multi layer perceptron (MLP) layer with normalized output as the $f$ function in the dual empirical forms of DSW and DGSW for the dual empirical forms of these distances). In all experiments, we use norm 2 as the ground metric for the Wasserstein distance. For GSW and DGSW, we use $r=1000$ for circular function. We use the code at \url{https://github.com/kimiandj/gsw} for Max-SW and Max-GSW-NN (use 3 MLP layers with Leaky ReLU activation as defining function). In this implementation, Max-SW and Max-GSW-NN uses 50 gradient-update times per minibatch to find the optimal direction.

We train models on MNIST, CelebA, CIFAR10 with batch size = 512. On LSUN we use batch size = 4096. We use Adam optimizer for all models with learning rate=0.0005 and betas=(0.5, 0.999), p =2. The range for hidden layer size of the MLP defining function of Max-GSW-NN is (32,100,784,1000). We tune $\lambda_C$ of DSW and DGSW by grid searching in $(1,10,100,1000)$ in every experiment. The number of epochs for MNIST is 200, CelebA is 50, CIFAR10 is 100, and LSUN is 20.

In evaluation, we use empirical distribution with 10000 samples from two target distribution to compute discrete Wasserstein distance via linear programming..

Generator architecture was used for MNIST  dataset:\\
$z \in \mathbb{R}^{32} \rightarrow FC_{100} \rightarrow ReLU \rightarrow FC_{200} \rightarrow ReLU \rightarrow FC_{400} \rightarrow ReLU \rightarrow FC_{784} \rightarrow ReLU $
\\
\\
Generator architecture was used for CelebA, CIFAR10 and LSUN dataset
$z \in \mathbb{R}^{100} \rightarrow TransposeConv_{512} \rightarrow BatchNorm \rightarrow ReLU  \rightarrow TransposeConv_{256} \rightarrow BatchNorm \rightarrow ReLU \rightarrow TransposeConv_{128} \rightarrow BatchNorm \rightarrow ReLU \rightarrow TransposeConv_{64} \rightarrow BatchNorm \rightarrow ReLU \rightarrow TransposeConv_{1}  \rightarrow Tanh$\\
Discriminator architecture was used for CelebA, CIFAR10 and LSUN dataset:\\\\
First part: $x \in \mathbb{R}^{64\times 64 \times 3} \rightarrow Conv_{64}  \rightarrow LeakyReLU_{0.2} \rightarrow Conv_{128} \rightarrow BatchNorm \rightarrow LearkyReLU_{0.2} \rightarrow Conv_{256} \rightarrow BatchNorm\rightarrow  LearkyReLU_{0.2}  \rightarrow Conv_{512} \rightarrow BatchNorm\rightarrow Tanh$ \\
Second part:  $ Conv_{1}  \rightarrow Sigmoid $ 
\\\\Joint Contrastive inference encoder architecture on MNIST: \\
$x \in \mathbb{R}^{28 \times 28}  \rightarrow  FC_{400}\rightarrow LeakyReLU_{0.2} \rightarrow  FC_{200} \rightarrow  LeakyReLU_{0.2} \rightarrow  FC_{100} \rightarrow  LeakyReLU_{0.2} \rightarrow FC_{32}$
\\\\
Joint Contrastive inference deocder architecture on MNIST:\\
$z \in \mathbb{R}^{32} \rightarrow FC_{100} \rightarrow ReLU \rightarrow FC_{200} \rightarrow ReLU \rightarrow FC_{400} \rightarrow ReLU \rightarrow FC_{784} \rightarrow ReLU $

\end{document}